\documentclass[11pt]{article}
\usepackage[usenames,dvipsnames,svgnames,table]{xcolor}

\usepackage{natbib}

\usepackage[ruled]{algorithm2e}

\usepackage{algorithmic}

\newtheorem{Theo}{Theorem}
\newtheorem{Def}{Definition}
\newtheorem{Lemma}{Lemma}
\newtheorem{Prop}{Proposition}

\usepackage{amssymb,amsmath}
\usepackage{stmaryrd}
\usepackage{graphicx}
\usepackage{natbib}
\usepackage{mathrsfs}
\usepackage{tikz}
\usepackage{pgfplots}
\usepackage[caption=false,font=footnotesize]{subfig}

\usepackage{enumerate}
\usepackage{latexsym}

\newcommand{\cC}{\mathscr{C}}

\newcommand{\cP}{\mathcal{P}}

\newcommand{\cS}{\mathcal{S}}
\newcommand{\cL}{\mathcal{L}}
\newcommand{\cT}{\mathcal{T}}
\newcommand{\defeq}{\stackrel{\mathrm{def}}{=}}

\newcommand{\beg}{\mathrm{beg}}
\newcommand{\fin}{\mathrm{end}}

\newcommand{\ud}{\mathrm{d}}

\newcommand{\taumax}{\tau^{\max}}

\newcommand{\bfa}{\mathbf{a}}
\newcommand{\bfb}{\mathbf{b}}
\newcommand{\bfc}{\mathbf{c}}

\newcommand{\bfu}{\mathbf{u}}
\newcommand{\bfv}{\mathbf{v}}

\newcommand{\bfx}{\mathbf{x}}

\newcommand{\bfq}{\mathbf{q}}
\newcommand{\bff}{\mathbf{f}}

\newcommand{\bfA}{\mathbf{A}}
\newcommand{\bfB}{\mathbf{B}}
\newcommand{\bfC}{\mathbf{C}}

\newcommand{\LC}{\mathrm{LC}}
\newcommand{\CLC}{\mathrm{CLC}}
\newcommand{\MVC}{\mathrm{MVC}}
\newcommand{\start}{\mathrm{start}}
\newcommand{\init}{\mathrm{init}}
\newcommand{\goal}{\mathrm{goal}}
\newcommand{\rand}{\mathrm{rand}}
\newcommand{\near}{\mathrm{near}}
\newcommand{\test}{\mathrm{test}}
\newcommand{\new}{\mathrm{new}}

\def\cf{cf.~}
\def\eg{e.g.~}

\def\False{\texttt{False}}
\def\True{\texttt{True}}
\def\wq{\widetilde{\bfq}}
\def\wqd{\dot{\wq}}
\def\wqdd{\ddot{\wq}}
\def\INTERPOLATE{\textrm{INTERPOLATE}}
\def\wtau{\widetilde{\tau}}

\usepackage{authblk}

\title{Admissible Velocity Propagation\,: Beyond Quasi-Static Path
  Planning for High-Dimensional Robots}

\author[1]{Quang-Cuong Pham}
\author[2]{St\'ephane Caron}
\author[1]{\\Puttichai Lertkultanon}
\author[3]{Yoshihiko Nakamura}
\affil[1]{School of Mechanical and Aerospace
  Engineering, NTU, Singapore}
\affil[2]{Laboratoire d'Informatique, de Robotique et
    de Micro\'{e}lectronique de Montpellier (LIRMM), CNRS / Universit\'{e} de
    Montpellier, France}
\affil[3]{Department of Mechano-Informatics,  University of Tokyo, Japan}

\usepackage{hyperref}

\begin{document}

\maketitle

\begin{abstract}
  Path-velocity decomposition is an intuitive yet powerful approach to
  address the complexity of kinodynamic motion planning. The difficult
  trajectory planning problem is solved in two separate and simpler
  steps\,: first, find a path in the configuration space that
  satisfies the geometric constraints (path planning), and second,
  find a time-parameterization of that path satisfying the kinodynamic
  constraints. A fundamental requirement is that the path found in the
  first step should be time-parameterizable. Most existing works
  fulfill this requirement by enforcing quasi-static constraints in
  the path planning step, resulting in an important loss in
  completeness. We propose a method that enables path-velocity
  decomposition to discover truly dynamic motions, i.e. motions that
  are not quasi-statically executable. At the heart of the proposed
  method is a new algorithm -- Admissible Velocity Propagation --
  which, given a path and an interval of reachable velocities at the
  beginning of that path, computes exactly and efficiently the
  interval of all the velocities the system can reach after traversing
  the path while respecting the system kinodynamic
  constraints. Combining this algorithm with usual sampling-based
  planners then gives rise to a family of new trajectory planners that
  can appropriately handle kinodynamic constraints while retaining the
  advantages associated with path-velocity decomposition. We
  demonstrate the efficiency of the proposed method on some difficult
  kinodynamic planning problems, where, in particular, quasi-static
  methods are guaranteed to fail\,\footnote{This paper is a
    substantially revised and expanded version of \citet{PhaX13rss},
    which was presented at the conference \emph{Robotics: Science and
      Systems}, 2013.}.
\end{abstract}

\section{Introduction}

Planning motions for robots with many degrees of freedom and subject
to kinodynamic constraints (i.e. constraints that involve higher-order
time-derivatives of the robot
configuration~\citep{DonX93acm,LK01ijrr}) is one of the most important
and challenging problems in robotics.  Path-velocity decomposition is
an intuitive yet powerful approach to address the complexity of
kinodynamic motion planning\,: first, find a \emph{path} in the
configuration space that satisfies the geometric constraints, such as
obstacle avoidance, joint limits, kinematic closure, etc. (path
planning), and second, find a \emph{time-parameterization} of that
path satisfying the kinodynamic constraints, such as torque limits for
manipulators, dynamic balance for legged robots, etc.

\paragraph{Advantages of path-velocity decomposition}

This approach was suggested as early as \citeyear{KZ86ijrr}~--~only a
few years after the birth of motion planning itself as a research
field~--~by \citeauthor{KZ86ijrr}, in the context of motion planning
amongst movable obstacles. Since then, it has become an important tool
to address many kinodynamic planning problems, from manipulators
subject to torque limits~\citep{BobX85ijrr,SM86tac,Bob88jra}, to
coordination of teams of mobile robots~\citep{SimX02tra,PA05ijrr}, to
legged robots subject to balance
constraints~\citep{KufX02ar,SulX10tr,HauX08ijrr,EscX13ras,Hau14ijrr,PS15tmech},
etc. In fact, to our knowledge, path-velocity decomposition [either
explicitly or implicitly, as e.g. when only the geometric motion is
planned and the time-parameterization is left to the execution phase]
is the only \emph{motion planning} approach that has been shown to
work on \emph{actual high-DOF robots} such as humanoids.

Path-velocity decomposition is appealing in that it \emph{exploits the
  natural decomposition} of the constraints, in most systems, into two
categories\,: those depending uniquely on the robot configuration, and
those depending in particular on the velocity, which in turn is
related to the energy of the system. Consider for instance a humanoid
robot in a multi-contact task. Such a robot must (1) avoid collision
with the environment, (2) avoid self-collisions, (3) respect kinematic
closure for the parts in contact with the environment (e.g. the stance
foot must be fixed with respect to the ground), (4) maintain
balance. It can be noted that constraints (1 -- 3) are exclusively
related to the configuration of the robot, while constraint (4), once
a path is given, depends mostly on the path velocity.

From a practical viewpoint, the two sub-problems -- geometric path
planning and kinodynamic time-parameterization -- have received so
much attention from the robotics community in the past three decades
that a large body of theory and good practices exist and can be
readily combined to yield efficient \emph{trajectory}
planners. Briefly, high-dimensional and cluttered geometric path
planning problems can now be solved in seconds thanks to
sampling-based planning algorithms such as PRM~\citep{KavX96tra} or
RRT~\citep{KL00icra} and to the dozens of heuristics that have been
developed for these algorithms. Regarding kinodynamic
time-parameterization, two important discoveries about the structure
of the problem have led to particularly efficient algorithmic
solutions. First, the bang-bang nature of the optimal velocity profile
was identified by \citep{BobX85ijrr,SM86tac}, leading to fast
\emph{numerical integration} methods~\citep[see][for extensive
  historical references]{Pha14tro}. Second, this problem was shown to
be reducible to a \emph{convex optimization} problem, leading to
robust and versatile convex-optimization-based solutions~\citep[see
  e.g.][]{VerX09tac,Hau14ijrr}.

\paragraph{Problems with state-space planning and trajectory
  optimization approaches}

Alternative approaches to path-velocity decomposition include planning
directly in the state space and trajectory optimization. The first
approach deploys traditional path planners such as
RRT~\citep{LK01ijrr} or PRM~\citep{HsuX02ijrr} directly into the
\emph{state space}, that is, the configuration space augmented with
velocity coordinates. Three main difficulties are associated with this
approach. First, the dimension of the state space is twice that of the
configuration space, resulting in higher algorithmic
complexity. Second, while connecting two adjacent configurations under
geometric constraints is trivial (using e.g. linear segments),
connecting two adjacent states under kinodynamic constraints is
considerably more challenging and time-consuming, requiring e.g. to
solve a two-point boundary value problem~\citep{LK01ijrr} or to sample
in the control space and to integrate forward the sampled
control~\citep{HsuX02ijrr,SK12tro,PapX14arxiv,LiX15wafr}. Third,
especially for state-space RRTs, designing a reasonable \emph{metric}
is particularly difficult\,:~\citet{ShkX09icra} showed that, even for
the 1-DOF pendulum subject to torque constraints, a state-space RRT
with a simple Euclidean metric is doomed to failure. The authors then
proposed to construct an efficient metric by solving local optimal
control problems. In a similar fashion, kinodynamic planners based on
locally linearized system dynamics were proposed, such as LQR-Tree
\citep{Ted09rss} or LQR-RRT$^*$ \citep{PerX12icra}.  While such methods
can be applied to low-DOF systems, the necessity to solve an optimal
control problem of the dimension of the system at each tree extension
makes it unlikely to scale to higher dimensions. For these reasons, in
spite of appealing completeness guarantees~\citep[under some precise
  conditions, see e.g.][]{CarX14icra,PapX14arxiv,KS15wafr}, there
exist, to our knowledge, few examples of successful application of
state-space planning to high-DOF systems with complex nonlinear
dynamics and constraints in challenging environments~\citep[see
  e.g.][]{SK12tro}.

The second approach, trajectory optimization, starts with an initial
trajectory, which may not be valid (for example the trajectory may not
reach the goal configuration, the robot may collide with the
environment or may lose balance at some time instants, etc.) One then
iteratively modifies the trajectory so as to decrease a cost -- which
encodes in particular how much the constraints are violated -- until
it falls below a certain threshold, implying in turn that the
trajectory reaches the goal and all constraints are satisfied. Many
interesting variations exist\,: the iterative modification step may be
deterministic~\citep{RatX09icra} or stochastic~\citep{KalX11icra}, the
optimization may be done \emph{through}
contact~\citep{MorX12acm,PT13wafr}, etc. However, for long
time-horizon and high-DOF systems, this approach requires
solving a large nonlinear optimization problem, which is
computationally challenging because of the huge problem size and the
existence of many local minima~\citep[see][for an extensive discussion
of the advantages and limitations of trajectory optimization and
comparison with path-velocity decomposition]{Hau14ijrr}.

\paragraph{The quasi-static condition and its limitations}

Coming back to path-velocity decomposition, a fundamental requirement
here is that the path found in the first step must be
time-parameterizable. A commonly-used method to fulfill this
requirement is to consider, in that step, the \emph{quasi-static}
constraints that are derived from the original kinodynamic constraints
by assuming that the motion is executed at zero velocity. Indeed, the
so-derived quasi-static constraints can be expressed using only
configuration-space variables, in such a way that planning with
quasi-static constraints is purely a geometric path planning
problem. In the context of legged robots for example, the balance of
the robot at zero velocity is guaranteed when the projection of the
center of gravity lies in the support area -- a purely geometric
condition. This quasi-static condition is assumed in most works
dedicated to the planning of complex humanoid motions~\citep[see
e.g.][]{KufX02ar}.

This workaround suffers however from a major limitation\,: the
quasi-static condition may be too restrictive and one thus may
overlook many possible solutions, i.e. incurring an important
\emph{loss in completeness}. For instance, legged robots walking with
ZMP-based control~\citep{VukX01humanoids} are dynamically balanced but
almost never satisfy the aforementioned quasi-static condition on the
center of gravity. Another example is provided by an actuated pendulum
subject to severe torque limits, but which can still be put into the
upright position by swinging back and forth several times. It is clear
that such solutions make an essential use of the system dynamics and
can in no way be discovered by quasi-static methods, nor by any method
that considers only configuration-space coordinates.

\paragraph{Planning truly dynamic motions}

Here we propose a method to overcome this limitation. At the heart of
the proposed method is a new algorithm -- Admissible Velocity
Propagation (AVP) -- which is based in turn on the classical
Time-Optimal Path Parameterization (TOPP) algorithm first introduced
by \citet{BobX85ijrr,SM86tac} and later perfected by many
others~\citep[see][and references therein]{Pha14tro}. In contrast with
TOPP, which determines \emph{one} optimal velocity profile along a
given path, AVP addresses \emph{all} valid velocity profiles along
that path, requiring only slightly more computation time than TOPP
itself. Combining AVP with usual sampling-based path planners, such as
RRT, gives rise to a family of new trajectory planners that can
appropriately handle kinodynamic constraints while retaining the
advantages associated with path-velocity decomposition.

The remainder of this article is organized as follows. In
Section~\ref{sec:avp}, we briefly recall the fundamentals of TOPP
before presenting AVP. In Section~\ref{sec:planning}, we show how to
combine AVP with usual sampling-based path planners such as RRT. In
Section~\ref{sec:applications}, we demonstrate the efficiency of the
new AVP-based planners on some challenging kinodynamic planning
problems -- in particular, those where the quasi-static approach is
\emph{guaranteed} to fail. In one of the applications, the planned
motion is executed on an actual 6-DOF robot. Finally, in
Section~\ref{sec:discussion}, we discuss the advantages and
limitations of the proposed approach (one particular limitation is
that the approach does not \emph{a priori} apply to under-actuated
systems) and sketch some future research directions.

\section{Propagating admissible velocities along a path}
\label{sec:avp}

\subsection{Background\,: Time-Optimal Path Parameterization (TOPP)}
\label{sec:topp}

As mentioned in the Introduction, there are two main approaches to
TOPP\,: ``numerical integration'' and ``convex optimization''. We
briefly recall the numerical integration
approach~\citep{BobX85ijrr,SM86tac}, on which AVP is based. For more
details about this approach, the reader is referred
to~\citet{Pha14tro}.

Let $\bfq$ be an $n$-dimensional vector representing the configuration
of a robot system. Consider second-order inequality constraints of the
form~\citep{Pha14tro}
%% \begin{equation}
%%   \label{eq:gengen}
%%   \bfA(\bfq)\ddot\bfq + \bfh(\dot\bfq,\bfq) \leq 0, 
%% \end{equation}
\begin{equation}
  \label{eq:gen}
  \bfA(\bfq)\ddot\bfq + \dot\bfq^\top \bfB(\bfq) \dot\bfq + \bff(\bfq) \leq 0, 
\end{equation}
where $\bfA(\bfq)$, $\bfB(\bfq)$ and $\bff(\bfq)$ are respectively an
$M\times n$ matrix, an $n\times M \times n$ tensor and an
$M$-dimensional vector. Inequality~(\ref{eq:gen}) is general and may
represent a large variety of second-order systems and constraints,
such as fully-actuated manipulators\,\footnote{When dry Coulomb
  friction or viscous damping are not negligible, one may consider
  adding an extra term $\bfC(\bfq)\dot\bfq$. Such a term would simply
  change the computation of the fields $\alpha$ and $\beta$ (see
  infra), but all the rest of the development would remain the
  same~\citep{SY89tra}.} subject to velocity, acceleration or torque
limits~\citep[see e.g.][]{BobX85ijrr,SM86tac}, wheeled vehicles
subject to sliding and tip-over constraints~\citep{SG91tra},
etc. \emph{Redundantly-actuated} systems, such as closed-chain
manipulators subject to torque limits or legged robots in
multi-contact subject to stability constraints, can also be
represented by inequality~(\ref{eq:gen})~\citep{PS15tmech}. However,
\emph{under-actuated} systems cannot be in general taken into account
by the framework, see Section~\ref{sec:discussion} for a more detailed
discussion.

Note that ``direct'' velocity bounds of the form
\begin{equation}
  \label{eq:velo}
\dot\bfq^\top \bfB_v(\bfq) \dot\bfq + \bff_v(\bfq) \leq 0,
\end{equation}
can also be taken into account~\citep{Zla96icra}. For clarity, we
shall not include such ``direct'' velocity bounds in the following
development. Rather, we shall discuss separately how to deal with such
bounds in Section~\ref{sec:avpremarks}.

Consider now a path $\cP$ in the configuration space, represented as the
underlying path of a trajectory $\bfq(s)_{s\in[0,s_\fin]}$.  Assume that
$\bfq(s)_{s\in[0,s_\fin]}$ is $C^1$- and piecewise $C^2$-continuous. 

\begin{Def}
  \label{def:valid}
  A \emph{time-parameterization} of $\cP$ -- or
  time-\emph{re}parameterization of $\bfq(s)_{s\in[0,s_\fin]}$ -- is
  an increasing \emph{scalar function} $s : [0,T']\rightarrow
  [0,s_\fin]$. A time-parameterization can be seen alternatively as a
  \emph{velocity profile}, which is the curve $\dot s(s)_{s \in
    [0,s_\fin]}$ in the $s$--$\dot s$ plane. We say that a
  time-parameterization or, equivalently, a velocity profile, is
  \emph{valid} if $s(t)_{t\in[0,T']}$ is continuous, $\dot s$ is
  always strictly positive, and the \emph{retimed} trajectory
  $\bfq(s(t))_{t\in[0,T']}$ satisfies the constraints of the system.
\end{Def}

To check whether the retimed trajectory satisfies the system
constraints, one may differentiate $\bfq(s(t))$ with respect to $t$\,:
\begin{equation}
  \label{eq:dotq}
  \dot\bfq= \bfq_s\dot s, \quad \ddot\bfq= \bfq_s  \ddot s + \bfq_{ss}
  \dot s^2 ,
\end{equation}
where dots denote differentiations with respect to the time parameter
$t$ and $\bfq_s=\frac{\ud \bfq}{\ud s}$ and $\bfq_{ss}=\frac{\ud^2
  \bfq}{\ud s^2}$.
Substituting (\ref{eq:dotq}) into (\ref{eq:gen}) then leads to
\[
\ddot s \bfA(\bfq)\bfq_s + \dot s^2 \bfA(\bfq)\bfq_{ss}+\dot s^2
\bfq_s^\top\bfB(\bfq)\bfq_s + \bff(\bfq) \leq 0,
\]
which can be rewritten as
\begin{equation}
  \label{eq:gen2}
  \ddot s \bfa(s) + \dot s^2 \bfb(s) + \bfc(s) \leq 0,\quad\textrm{where}  
\end{equation}
\begin{eqnarray}
  \label{eq:toto}
  \bfa(s)&\defeq&\bfA(\bfq(s))\bfq_s(s),\nonumber\\
  \bfb(s)&\defeq&\bfA(\bfq(s))\bfq_{ss}(s) + \bfq_s(s)^\top\bfB(\bfq(s))\bfq_s(s),\\
  \bfc(s)&\defeq&\bff(\bfq(s)).\nonumber 
\end{eqnarray}

Each row $i$ of equation~(\ref{eq:gen2}) is of the form
\[
a_i(s)\ddot s + b_i(s)\dot s^2 + c_i(s) \leq 0.
\]
Next,
\begin{itemize}
\item if $a_i(s)>0$, then one has $\ddot s \leq
  \frac{-c_i(s)-b_i(s)\dot s^2}{a_i(s)}$.  Define the acceleration
  \emph{upper bound} $\beta_i(s, \dot s) \defeq\frac{-c_i(s)-b_i(s)\dot
    s^2}{a_i(s)}$;
\item if $a_i(s)<0$, then one has $\ddot s \geq
  \frac{-c_i(s)-b_i(s)\dot s^2}{a_i(s)}$. Define the acceleration
  \emph{lower bound} $\alpha_i(s, \dot s) \defeq\frac{-c_i(s)-b_i(s)\dot
    s^2}{a_i(s)}$.
\end{itemize}

One can then define for each $(s,\dot s)$
\begin{equation}
  \label{eq:tata}
  \alpha(s,\dot s)\defeq\max_i \alpha_i(s,\dot s),\nonumber \quad
  \beta(s,\dot s)\defeq\min_i \beta_i(s,\dot s).\nonumber 
\end{equation}

From the above transformations, one can conclude that
$\bfq(s(t))_{t\in[0,T']}$ satisfies the constraints~(\ref{eq:gen}) if
and only if
\begin{equation}
\label{eq:bounds}
\forall t\in[0,T']\quad \alpha(s(t),\dot s(t)) \leq \ddot s(t) \leq
\beta(s(t),\dot s(t)).  
\end{equation}

Note that $(s,\dot s)\mapsto(\dot s,\alpha(s,\dot s))$ and $(s,\dot
s)\mapsto(\dot s,\beta(s,\dot s))$ can be viewed as two vector fields
in the $s$--$\dot s$ plane. One can integrate velocity profiles
following the field $(\dot s,\alpha(s,\dot s))$ (from now on, $\alpha$
in short) to obtain \emph{minimum acceleration} profiles (or
$\alpha$-profiles), or following the field $\beta$ to obtain
\emph{maximum acceleration} profiles (or $\beta$-profiles).

Next, observe that if $\alpha(s,\dot s)>\beta(s,\dot s)$ then, from
(\ref{eq:bounds}), there is no possible value for $\ddot s$. Thus, to
be valid, every velocity profile must stay below the maximum velocity
curve ($\MVC$ in short) defined by\,\footnote{Setting $\MVC(s)=0$
  whenever $\alpha(s,0) > \beta(s,0)$ as in~(\ref{eq:mvc}) precludes
  multiple-valued MVCs \citep[cf.][]{SD85icra}. We made this choice
  throughout the paper for clarity of exposition. However, in the
  implementation, we did consider multiple-valued MVCs.}
\begin{equation}
  \label{eq:mvc}
\MVC(s)\defeq\left\{
  \begin{array}{cl}
    \min \{\dot s\geq 0: \alpha(s,\dot s)= \beta(s,\dot
    s)\}&\mathrm{if}\ \alpha(s,0) \leq \beta(s,0),\\
    0&\mathrm{if}\ \alpha(s,0) > \beta(s,0).
  \end{array}
\right.  
\end{equation}

\begin{figure}[htp]
    \centering
    \includegraphics[width=10cm]{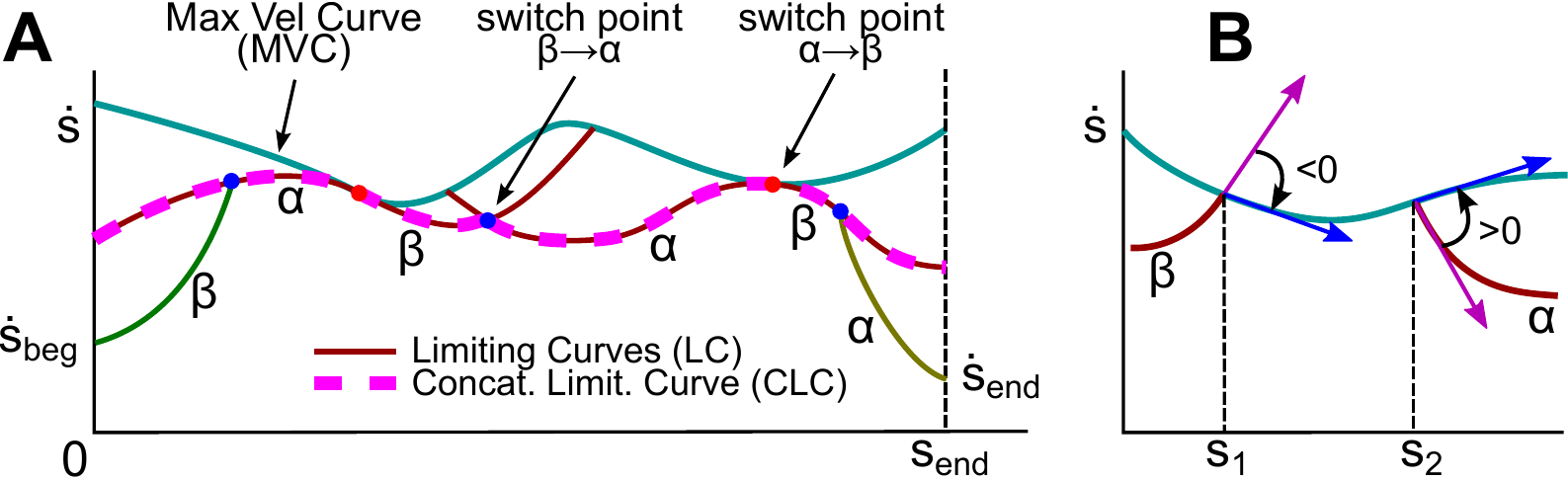}
    \caption{\textbf{A}\,: Illustration for Maximum Velocity Curve
      (MVC) and Concatenated Limiting Curve (CLC). The optimal
      velocity profile follows the green $\beta$-profile, then a
      portion of the CLC, and finally the yellow
      $\alpha$-profile. \textbf{B}\,: Illustration for the Switch
      Point Lemma.}
  \label{fig:bobrow}
\end{figure}

It was shown \citep[see e.g.][]{SL92jdsmc} that the time-minimal
velocity profile is obtained by a \emph{bang-bang}-type control, i.e.,
whereby the optimal profile follows alternatively the $\beta$ and
$\alpha$ fields while always staying below the $\MVC$. A method to
find the optimal profile then consists in (see Fig.~\ref{fig:bobrow}A
for illustration)\,:
\begin{itemize}
\item find all the possible $\alpha\rightarrow\beta$ switch
  points. There are three types of such switch points\,:
  ``discontinuous'', ``singular'' or ``tangent'' and they must all be
  on the $\MVC$. The procedure to find these switch points is detailed
  in~\citet{Pha14tro};
\item from each of these switch points, integrate backward following
  $\alpha$ and forward following $\beta$ to obtain the Limiting Curves
  ($\LC$)~\citep{SY89tra};
\item construct the Concatenated Limiting Curve (CLC) by considering,
  for each $s$, the value of the lowest $\LC$ at~$s$;
\item integrate forward from $(0,\dot s_\beg)$ following $\beta$ and
  backward from $(s_\fin,\dot s_\fin)$ following $\alpha$, and
  consider the intersection of these profiles with each other or with
  the CLC. Note that the path velocities $\dot s_\beg$ and $\dot
  s_\fin$ are computed from the desired initial and final velocities
  $v_\beg$ and $v_\fin$ by
  \begin{equation}
    \label{eq:convert}
    \dot s_\beg\defeq v_\beg/\|\bfq_s(0)\|, \quad \dot s_\fin\defeq v_\fin/\|\bfq_s(s_\fin)\|.
  \end{equation}
\end{itemize}

We now prove two lemmata that will be important later on.

\begin{Lemma}[Switch Point Lemma]
  \label{lemma:sp}
  Assume that a forward $\beta$-profile hits the $\MVC$ at $s=s_1$ and
  a backward $\alpha$-profile hits the $\MVC$ at $s=s_2$, with
  $s_1<s_2$, then there exists at least one $\alpha\rightarrow\beta$
  switch point on the $\MVC$ at some position $s_3\in[s_1,s_2]$.
\end{Lemma}

\noindent\textbf{Proof} At $(s_1,\MVC(s_1))$, the angle from the vector
$\beta$ to the tangent to the $\MVC$ is negative (see
Fig.~\ref{fig:bobrow}B). In addition, since we are on the $\MVC$, we
have $\alpha=\beta$, thus the angle from $\alpha$ to the tangent is
negative too. Next, at $(s_2,\MVC(s_2))$, the angle of $\alpha$ to the
tangent to the $\MVC$ is positive (see Fig.~\ref{fig:bobrow}B). Thus,
since the vector field $\alpha$ is continuous, there exists, between
$s_1$ and $s_2$
\begin{enumerate}[(i)]
\item either a point where the angle between $\alpha$ and the tangent to the
  $\MVC$ is 0 -- in which case we have a \emph{tangent} switch point;
\item or a point where the $\MVC$ is discontinuous -- in which case we
  have a \emph{discontinuous} switch point;
\item or a point where the $\MVC$ is continuous but non differentiable
  -- in which case we have a \emph{singular} switch point.
\end{enumerate}
For more details, the reader is referred to~\citet{Pha14tro}. $\Box$

\begin{Lemma}[Continuity of the CLC]
  \label{lemma:clc}
  Either one of the $\LC$'s reaches $\dot s=0$, or the $\CLC$ is
  \emph{continuous}.
\end{Lemma}

\noindent\textbf{Proof} Assume by contradiction that no $\LC$ reaches $\dot
s=0$ and that there exists a ``hole'' in the $\CLC$. The left border
$s_1$ of the hole must then be defined by the intersection of the
$\MVC$ with a forward $\beta$-$\LC$ (coming from the previous
$\alpha\rightarrow\beta$ switch point), and the right border $s_2$ of
the hole must be defined by the intersection of the $\MVC$ with a
backward $\alpha$-$\LC$ (coming from the following
$\alpha\rightarrow\beta$ switch point). By Lemma~\ref{lemma:sp} above,
there must then exist a switch point between $s_1$ and $s_2$, which
contradicts the definition of the hole. $\Box$

\subsection{Admissible Velocity Propagation (AVP)}
\label{sec:avp2}

This section presents the Admissible Velocity Propagation algorithm
(AVP), which constitutes the heart of our approach. This algorithm
takes as inputs\,:
\begin{itemize}
\item a path $\cP$ in the configuration space, and
\item an interval $[\dot s_\beg^{\min},\dot s_\beg^{\max}]$ of initial
  path velocities;
\end{itemize}
and returns the \emph{interval} (cf. Theorem~\ref{theo:interval})
$[\dot s_\fin^{\min}, \dot s_\fin^{\max}]$ of \emph{all} path
velocities that the system can reach \emph{at the end} of $\cP$ after
traversing $\cP$ while respecting the system
constraints\,\footnote{\citet{JH12icra} also introduced a velocity
  interval propagation algorithm along a path but for pure kinematic
  constraints and moving obstacles.}. The algorithm comprises the
following three steps\,:
\begin{description}
\item[A] Compute the limiting curves;
\item[B] Determine the \emph{maximum} final velocity $\dot s_\fin^{\max}$ by
  integrating \emph{forward} from $s=0$;
\item[C] Determine the \emph{minimum} final velocity $\dot s_\fin^{\min}$ by
  bisection search and by integrating \emph{backward} from $s=s_\fin$.
\end{description}

We now detail each of these steps.

\paragraph{A ~ Computing the limiting curves}

We first compute the Concatenated Limiting Curve (CLC) as shown in
Section~\ref{sec:topp}. From Lemma~\ref{lemma:clc}, either one of the
$\LC$'s reaches 0 or the $\CLC$ is continuous. The former case is
covered by A1 below, while the latter is covered by A2--5.

\begin{figure}[htp]
    \centering
    \hspace{-0.5cm}\textbf{A}\hspace{5cm}\textbf{B}\\    
    \includegraphics[width=10cm]{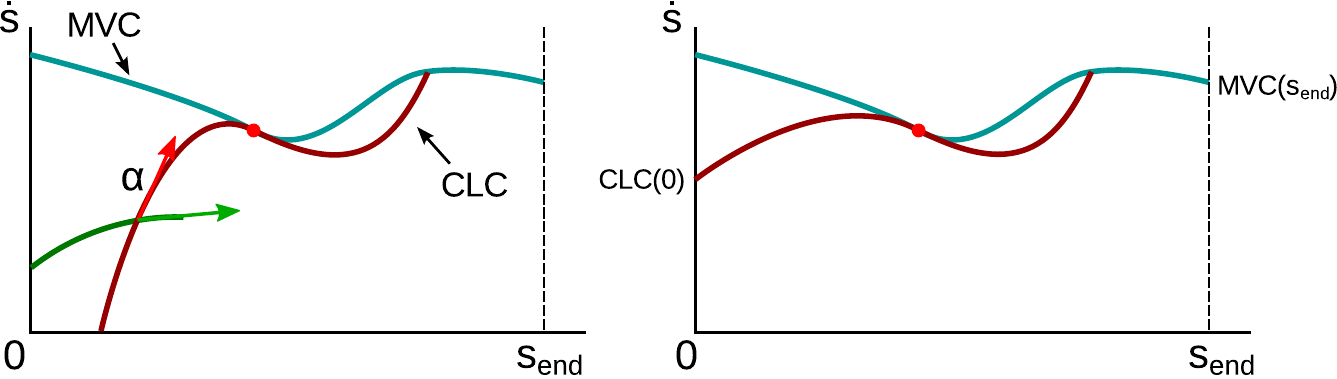}
    \caption{Illustration for step A (computation of the
      $\LC$'s). \textbf{A}\,: illustration for case A1. A profile that
      crosses an $\alpha$-$\CLC$ violates the $\alpha$
      bound. \textbf{B}\,: illustration for case A3.}
  \label{fig:A}
\end{figure}

\begin{description}
\item[A1] One of the $\LC$'s hits the line $\dot s=0$. In this case, the
  path cannot be traversed by the system without violating the
  kinodynamic constraints\,: AVP returns \texttt{Failure}.
  %\TODO{the following is not clear (because internal reference to the
  %LC algorithm), let's discuss it}
  Indeed, assume that a backward ($\alpha$) profile hits $\dot
  s=0$. Then any profile that goes from $s=0$ to $s=s_\fin$ must cross
  that profile somewhere and \emph{from above}, which violates the
  $\alpha$ bound (see Figure~\ref{fig:A}A). Similarly, if a
  forward ($\beta$) profile hits $\dot s=0$, then that profile must be
  crossed somewhere and \emph{from below}, which violates the $\beta$
  bound. Thus, no valid profile can go from $s=0$ to $s=s_\fin$;
\end{description}

The CLC is now assumed to be continuous and strictly positive. Since
it is bounded by $s=0$ from the left, $s=s_\fin$ from the right, $\dot
s=0$ from the bottom and the MVC from the top, there are only four
exclusive and exhaustive cases, listed below.

\begin{description}
\item[A2] The $\CLC$ hits the $\MVC$ while
  integrating backward and while integrating forward. In this case,
  let $\dot s_\beg^*\defeq\MVC(0)$ and go to \textbf{B}. The 
  situation where there is no switch point is assimilated to this case;
\item[A3] The $\CLC$ hits $s=0$ while integrating backward, and the
  $\MVC$ while integrating forward (see Figure~\ref{fig:A}B). In
  this case, let $\dot s_\beg^*\defeq\CLC(0)$ and go to \textbf{B};
\item[A4] The $\CLC$ hits the $\MVC$ while
  integrating backward, and $s=s_\fin$ while integrating forward. In
  this case, let $\dot s_\beg^*\defeq\MVC(0)$ and go to
  \textbf{B};
\item[A5] The $\CLC$ hits $s=0$ while integrating
  backward, and $s=s_\fin$ while integrating forward. In this case,
  let $\dot s_\beg^*\defeq\CLC(0)$ and go to \textbf{B}.
\end{description}

\paragraph{B ~ Determining the maximum final velocity}
\label{sec:B}

Note that, in any of the cases A2--5, $\dot s_\beg^*$ was defined so
that no valid profile can start above it. Thus, if $\dot s_\beg^{\min}
> \dot s_\beg^*$, the path is not traversable\,: AVP returns
\texttt{Failure}. Otherwise, the interval of \emph{valid} initial
velocities is $[\dot s_\beg^{\min}, \dot s_\beg^{{\max}*}]$ where
$\dot s_\beg^{{\max}*} \defeq \min(\dot s_\beg^{\max},\dot s_\beg^*)$.

\begin{Def}
  Under the nomenclature introduced in Definition~\ref{def:valid}, we
  say that a velocity $\dot s_\fin$ is a \emph{valid} final velocity
  if there exists a valid profile that starts at $(0,\dot s_0)$ for
  some $\dot s_0\in[\dot s_\beg^{\min},\dot s_\beg^{\max}]$ and ends
  at $(s_\fin,\dot s_\fin)$.
\end{Def}

\begin{figure}[htp]
    \centering
    \includegraphics[width=7cm]{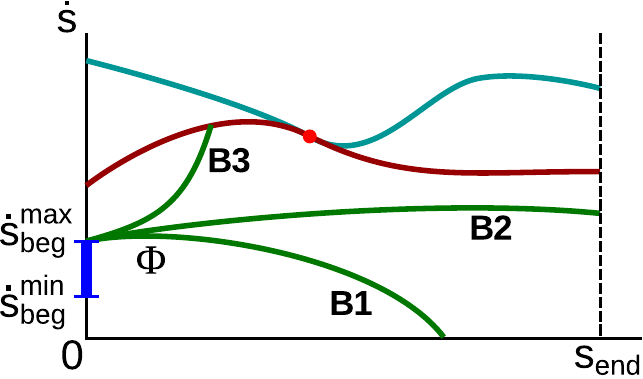}
    \caption{Illustration for step B\,: one can determine the maximum
      final velocity by integrating forward from $(0,\dot s_\beg^*)$.}
  \label{fig:B}
\end{figure}

We argue that the maximum valid final velocity can be obtained by
integrating forward from $\dot s_\beg^{{\max}*}$ following
$\beta$. Let's call $\Phi$ the velocity profile obtained by doing so.
Since $\Phi$ is continuous and bounded by $s=s_\fin$ from the right,
$\dot s=0$ from the bottom, and either the MVC or the CLC from the
top, there are four exclusive and exhaustive cases, listed below (see
Figure~\ref{fig:B} for illustration).

\begin{description}
\item[B1] $\Phi$ hits $\dot s=0$ (cf. profile B1 in
  Fig.~\ref{fig:B}). Here, as in the case A1, the path is not
  traversable\,: AVP returns \texttt{Failure}.
  %\TODO{same question here...} 
  Indeed, any profile that
  starts below $\dot s_\beg^{{\max}*}$ and tries to reach $s=s_\fin$
  must cross $\Phi$ somewhere and \emph{from below}, thus violating
  the $\beta$ bound;
\item[B2] $\Phi$ hits $s=s_\fin$ (cf. profile B1 in
  Fig.~\ref{fig:B}). Then $\Phi(s_\fin)$ corresponds to the $\dot
  s_\fin^{\max}$ we are looking for. Indeed, $\Phi(s_\fin)$ is
  reachable -- precisely by $\Phi$ --, and to reach any value above
  $\Phi(s_\fin)$, the corresponding profile would have to cross $\Phi$
  somewhere and from below;
  %\TODO{why?}
\item[B3]  $\Phi$ hits the $\CLC$. There are two sub-cases:
  \begin{enumerate}[{B3}a]
  \item If we proceed from cases A4 or A5 (in which the $\CLC$ reaches
    $s=s_\fin$, cf. profile B3 in Fig.~\ref{fig:B}), then
    $\CLC(s_\fin)$ corresponds to the $\dot s_\fin^{\max}$ we are
    looking for.  Indeed, $\CLC(s_\fin)$ is reachable -- precisely by
    the concatenation of $\Phi$ and the $\CLC$ --, and no value above
    $\CLC(s_\fin)$ can be valid by the definition of the $\CLC$;
  \item If we proceed from cases A2 or A3, then the $\CLC$ hits the
    $\MVC$ while integrating forward, say at $s=s_1$; we then proceed
    as in case B4 below;
  \end{enumerate}
\item[B4] $\Phi$ hits the $\MVC$, say at $s=s_1$. It is clear that
  $\MVC(s_\fin)$ is an upper bound of the valid final velocities, but
  we have to ascertain whether this value is reachable. For this, we
  use the predicate IS\_VALID defined in Box~\ref{algo:valid} of
  \textbf{C}\,:
  \begin{itemize}
  \item if IS\_VALID$(\MVC(s_\fin))$, then
    $\MVC(s_\fin)$ is the $\dot s_\fin^{\max}$ we are looking for;
  \item else, the path is not traversable\,: AVP returns
    \texttt{Failure}. Indeed, as we shall see, if for a certain $\dot
    s_\test$, the predicate IS\_VALID($\dot s_\test$) is
    \texttt{False}, then no velocity below $\dot s_\test$ can be valid
    either.
  \end{itemize}
\end{description}

%\TODO{do all these cases happen in practice?}

\paragraph{C ~ Determining the minimum final velocity}
\label{sec:C}

Assume that we proceed from the cases B2--4. Consider a final velocity
$\dot s_\test$ where
\begin{itemize}
\item $\dot{s}_\test <\Phi(s_\fin)$ if we proceed from B2;
\item $\dot s_\test < \CLC(s_\fin)$ if we proceed from B3a;
\item $\dot s_\test < \MVC(s_\fin)$ if we proceed from B3b or B4.
\end{itemize}

Let us integrate backward from $(s_\fin,\dot s_\mathrm{test})$
following $\alpha$ and call the resulting profile $\Psi$. We have the
following lemma.

\begin{Lemma}
  \label{lemma:psi}
  $\Psi$ cannot hit the $\MVC$ before hitting either $\Phi$ or the
  $\CLC$.
\end{Lemma}

\noindent\textbf{Proof} If we proceed from B2 or B3a, then it is clear that
$\Psi$ must first hit $\Phi$ (case B2) or the $\CLC$ (case B3a) before
hitting the $\MVC$. If we proceed from B3b or B4, assume by
contradiction that $\Psi$ hits the $\MVC$ first at a position
$s=s_2$. Then by Lemma~\ref{lemma:sp}, there must exist a switch point
between $s_2$ and the end of the $\CLC$ (in case B3b) or the end of
$\Phi$ (in case B4). In both cases, there is a contradiction with the
fact that the $\CLC$ is continuous. $\Box$

We can now detail in Box~\ref{algo:valid} the predicate IS\_VALID
which assesses whether a final velocity $\dot s_\test$ is valid.

%\TODO{I think we should rename $\Phi \to \Phi_\textrm{fwd}$ and $\Psi$ to $\Phi_{\textrm{bwd}}$}

\begin{algorithm}
\caption{IS\_VALID}
\textbf{Input:} candidate final velocity $\dot s_\mathrm{test}$

\textbf{Output:} \texttt{True} iff there exists a valid velocity profile
  with final velocity $\dot s_\mathrm{test}$

  Consider the profile $\Psi$ constructed above. Since it must hit
  $\Phi$ or the CLC before hitting the MVC, the following five cases
  are exclusive and exhaustive (see Fig.~\ref{fig:C} for
  illustrations)\,:

\begin{description}
\item[C1] $\Psi$ hits $\dot s=0$
  (Fig.~\ref{fig:C}, profile C1). Then, as in cases A1 or B1, no
  velocity profile can reach $s_\mathrm{test}$\,: return \False;
\item[C2] $\Psi$ hits $s=0$ for some $\dot
  s_0<\dot s_\beg^{\min}$ (see Figure~\ref{fig:C}, profile C2). Then
  any profile that ends at $\dot s_\mathrm{test}$ would have to hit
  $\Psi$ from above, which is impossible\,: return \False;
\item[C3] $\Psi$ hits $s=0$ at a point $\dot
  s_0 \in [\dot s_\beg^{\min},\dot s_\beg^{{\max}*}]$
  (Fig.~\ref{fig:C}, profile C3). Then $\dot s_\mathrm{test}$ can be
  reached following the valid velocity profile $\Psi$\,: return
  \True. (Note that, if $\dot s_0 > \dot s_\beg^{{\max}*}$ then $\Psi$
  must have crossed $\Phi$ somewhere before arriving at $s=0$, which
  is covered by case C4 below);
\item[C4] $\Psi$ hits $\Phi$ (Fig.~\ref{fig:C},
  profile C4). Then $\dot s_\mathrm{test}$ can be reached, precisely
  by the concatenation of a part of $\Phi$ and $\Psi$\,: return \True;
\item[C5] $\Psi$ hits the $\CLC$
  (Fig.~\ref{fig:C}, profile C5). Then $\dot s_\mathrm{test}$ can be
  reached, precisely by the concatenation of $\Phi$, a part of the
  $\CLC$ and $\Psi$\,: return \True.
\end{description}
\label{algo:valid}
\end{algorithm}

\begin{figure}[htp]
    \centering
    \includegraphics[width=7cm]{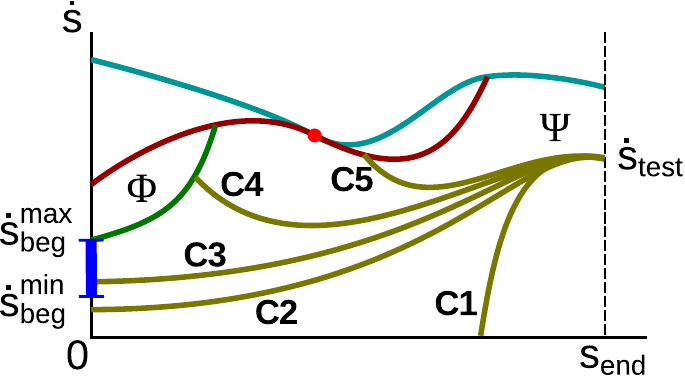}
    \caption{Illustration for the predicate IS\_VALID\,: one can
      assess whether a final velocity $\dot s_\mathrm{test}$ is valid
      by integrating backward from $(s_\fin,\dot s_\mathrm{test})$.}
  \label{fig:C}
\end{figure}

At this point, we have that, either the path is not traversable, or we
have determined $\dot s_\fin^{\max}$ in \textbf{B}. Remark from C3--5
that, if some $\dot s_0$ is a valid final velocity, then any $\dot s
\in [\dot s_0, \dot s_\fin^{\max}]$ is also valid.
%\TODO{this is not clear to me (a not-so-lambda reader)}. 
Similarly, from C1 and C2, if some $\dot s_0$ is \emph{not} a valid
final velocity, then \emph{no} $\dot s \leq s_0$ can be valid.  We
have thus established the following result\,:

\begin{Theo}
  \label{theo:interval}
  The set of valid final velocities is an interval.  
\end{Theo}

This interval property enables one to efficiently search for the
minimum final velocity as follows. First, test whether~$0$ is a valid
final velocity: if IS\_VALID$(0)$, then the sought-after $\dot
s_\fin^{\min}$ is~0. Else, run a standard bisection search with
initial bounds $(0,\dot s_\fin^{\max}]$ where 0 is not valid and $\dot
s_\fin^{\max}$ is valid. Thus, after executing $\log_2(1/\epsilon)$
times the routine IS\_VALID, one can determine $\dot s_\fin^{\min}$
with a precision~$\epsilon$.

\subsection{Remarks}
\label{sec:avpremarks}

\paragraph{Implementation and complexity of AVP} As clear from the
previous section, AVP can be readily adapted from the numerical
integration approach to TOPP. As a matter of fact, we implemented AVP
in about 100 lines of C++ code based on the TOPP library we developed
previously (see \url{https://github.com/quangounet/TOPP}).

In terms of complexity, the main difference between AVP and TOPP lies
in the bisection search of step C, which requires $\log(1/\epsilon)$
backward integrations. However, in practice, these integrations
terminate quickly, either by hitting the MVC or the line $\dot
s=0$. Thus, the actual running time of AVP is only slightly larger
than that of TOPP. As illustration, in the bottle experiment of
Section~\ref{sec:bottle}, we considered 100 random paths, discretized
with grid size $N=1000$. TOPP and AVP (with the bisection precision
$\epsilon=0.01$) under velocity, acceleration and balance constraints
took the same amount of computation time 0.033 $\pm$ 0.003\,s per
path.

\paragraph{``Direct'' velocity bounds} ``Direct'' velocity bounds in
the form of (\ref{eq:velo}) give rise to another maximum velocity
curve, say $\MVC_D$, in the $(s,\dot s)$ space. When
a forward profile intersects $\MVC_D$ (before reaching the ``Bobrow's
$\MVC$''), several cases can happen :
\begin{enumerate}
\item If ``sliding'' along the $\MVC_D$ does not violate the actuation
  bounds~(\ref{eq:gen}), then slide as far as possible along the
  $\MVC$. The ``slide'' terminates either (a) when the maximum
  acceleration vector $\beta$ points downward from the $\MVC_D$: in
  this case follow that vector out of $\MVC_D$ or (b) when the minimum
  acceleration vector $\alpha$ points upward from the $\MVC_D$: in
  this case, proceed as in~2;
\item If not, then search forward on the $\MVC_D$ until finding a
  point from which one can integrate backward. Such a point is
  guaranteed to exist and the backward profile will intersect the
  forward profile. 
\end{enumerate}
For more details, the reader is referred to~\citep{Zla96icra}. 

This reasoning can be extended to AVP: when integrating the forward or
the backward profiles (in steps A, B, C of the algorithm), each time a
profile intersects the $\MVC_D$, one simply applies the above steps.

\paragraph{AVP-backward} Consider the ``AVP-backward'' problem\,:
given an interval of final velocities $[\dot s_\fin^{\min}, \dot
s_\fin^{\max}]$, compute the interval $[\dot s_\beg^{\min},\dot
s_\beg^{\max}]$ of all possible initial velocities. As we shall see in
Section~\ref{sec:implementation}, AVP-backward is essential for the
\emph{bi-directional} version of AVP-RRT \citep[see also][]{NM91tra}.

It turns out that AVP-backward can be easily obtained by modifying AVP
as follows~\citep{LP14ijcai}\,:
\begin{itemize}
\item step A of AVP-backward is the same as in AVP;
\item in step B of AVP-backward, one integrates \emph{backward} from
  $\dot s_\fin^{{\min}*}$ instead of integrating \emph{forward} from
  $\dot s_\beg^{{\max}*}$;
\item in the bisection search of step C of AVP-backward, one
  integrates \emph{forward} from $(0,\dot s_\test)$ instead of
  integrating \emph{backward} from$(s_\fin,\dot s_\test)$.
\end{itemize}

\paragraph{Convex optimization approach} As mentioned in the
Introduction, ``convex optimization'' is another possible approach to
TOPP~\citep{VerX09tac,Hau14ijrr}. It is however unclear to us whether
one can modify that approach to yield a ``convex-optimization-based
AVP'' other than sampling a large number of $(\dot s_\start,\dot
s_\fin)$ pairs and running the ``convex-optimization-based TOPP''
between $(0,\dot s_\start)$ and $(s_\fin,\dot s_\fin)$, which would
arguably be very slow.

\section{Kinodynamic trajectory planning using AVP}
\label{sec:planning}

\subsection{Combining AVP with sampling-based planners}
\label{sec:avprrt}

\begin{figure}[htp]
    \centering
    \includegraphics[width=14cm]{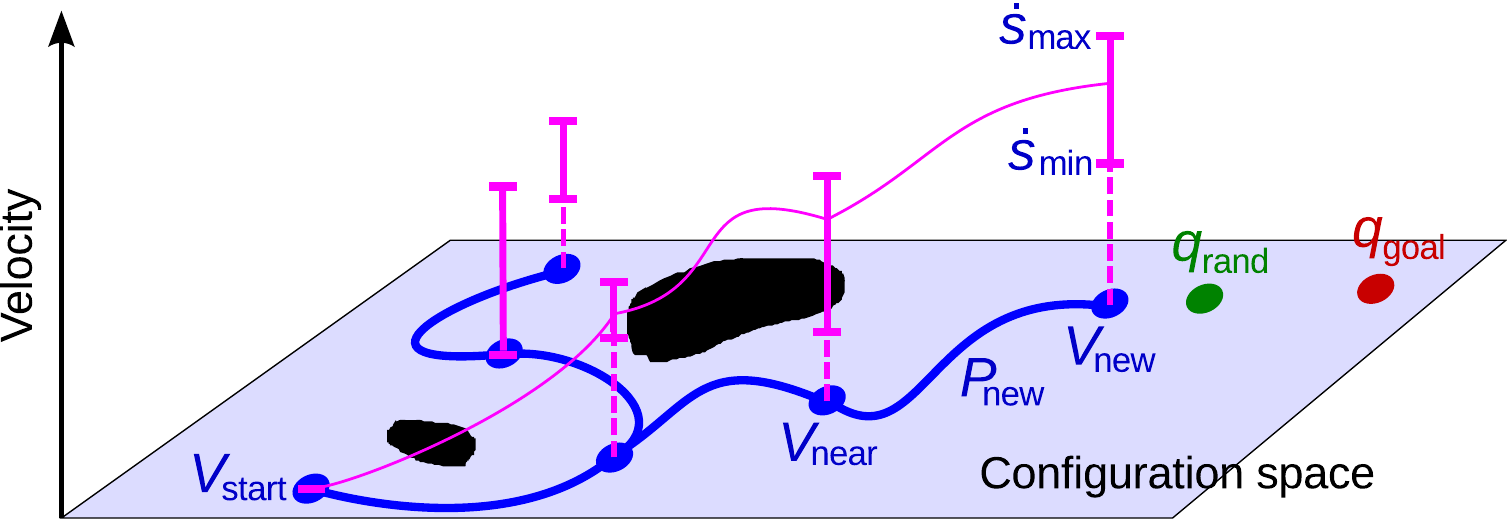}
    \caption{Illustration for AVP-RRT. The horizontal plane represents
      the configuration space while the vertical axis represents the
      path velocity space. Black areas represent configuration space
      obstacles. A vertex in the tree is composed of a configuration
      (blue disks), the incoming path from the parent (blue curve),
      and the interval of admissible velocities (bold magenta
      segments). At each tree extension step, one interpolates a
      smooth, collision-free path in the configuration space and
      propagates the interval of admissible velocities along that path
      using AVP. The fine magenta line shows one possible valid
      velocity profile (which is guaranteed to exist by AVP) ``above''
      the path connecting $\bfq_\start$ and $\bfq_\new$.}
  \label{fig:avp-rrt}
\end{figure}

The AVP algorithm presented in Section~\ref{sec:avp2} is general and
can be combined with various iterative path planners. As an example,
we detail in Box~\ref{algo:rrt} and illustrate in Figure~\ref{fig:avp-rrt}
a planner we call AVP-RRT, which results from the combination of AVP
with the standard RRT path planner~\citep{KL00icra}.

As in the standard RRT, AVP-RRT iteratively constructs a tree $\cT$ in
the configuration space. However, in contrast with the standard RRT, a
vertex $V$ here consists of a triplet ($V$.config, $V$.inpath,
$V$.interval) where $V$.config is an element of the configuration
space $\cC$, $V$.inpath is a path $\cP \subset \cC$ that connects the
configuration of $V$'s parent to $V$.config, and $V$.interval is the
interval of reachable velocities at $V$.config, that is, at the end of
$V$.inpath.

At each iteration, a random configuration $\bfq_\rand$ is
generated. The EXTEND routine (see Box~\ref{algo:extend}) then tries
to extend the tree $\cT$ \emph{towards} $\bfq_\rand$ from the closest
-- in a certain metric $d$ -- vertex in $\cT$. The algorithm
terminates when either
\begin{itemize}
\item A newly-found vertex can be connected to the goal configuration
  (line 10 of Box~\ref{algo:rrt}). In this case, AVP guarantees by
  recursion that there exists a path from $\bfq_\start$ to
  $\bfq_\goal$ and that this path is time-parameterizable;
\item After $N_\mathrm{maxrep}$ repetitions, no vertex could be
  connected to $\bfq_\goal$. In this case, the algorithm returns
  \texttt{Failure}.
\end{itemize}

\begin{algorithm}
\caption{AVP-RRT}
\label{algo:rrt}

\textbf{Input\,:} $\bfq_\start$, $\bfq_\goal$

\textbf{Output\,:} A valid trajectory connecting $\bfq_\start$
to $\bfq_\goal$ or \texttt{Failure}

\begin{algorithmic}[1]
\STATE $\cT\leftarrow$ NEW\_TREE()

\STATE $V_\start\leftarrow$ NEW\_VERTEX()

\STATE $V_\start.\mathrm{config}\leftarrow\bfq_\start$; $V_\start.\mathrm{inpath}\leftarrow\texttt{Null}$; $V_\start.\mathrm{interval}\leftarrow [0,0]$

\STATE INITIALIZE($\cT,V_\start$)

\FOR{$\mathrm{rep}=1$ to $N_\mathrm{maxrep}$}

\STATE $\bfq_\rand\leftarrow$ RANDOM\_CONFIG()

\STATE $V_\new\leftarrow$ EXTEND($\cT,\bfq_\rand$)

\IF{EXTEND succeeds}

\STATE ADD\_VERTEX($\cT,V_\new$)

\IF{CONNECT($V_\new,\bfq_\goal$) succeeds}

\RETURN COMPUTE\_TRAJECTORY($\cT,\bfq_\goal$)

\ENDIF
\ENDIF
\ENDFOR
\RETURN \texttt{Failure}
\end{algorithmic}

\end{algorithm}

\begin{algorithm}
\caption{EXTEND}
\label{algo:extend}

\textbf{Input\,:} $\cT$, $\bfq_\rand$

\textbf{Output\,:} A new vertex $V_\new$ or \texttt{Null}

\begin{algorithmic}[1]

\STATE $V_\near\leftarrow$ NEAREST\_NEIGHBOR($\cT,\bfq_\rand$)
\STATE $(\cP_\new,\bfq_\new) \leftarrow$ INTERPOLATE($V_\near,\bfq_\rand$)
\IF{$\cP$ is collision-free}
\STATE $[\dot s_{\min},\dot s_{\max}]\leftarrow$ AVP($\cP_\new,V_\near.\mathrm{interval}$)
\IF{AVP succeeds}
\STATE $V_\new\leftarrow$ NEW\_VERTEX()
\STATE $V_\new.\mathrm{config}\leftarrow\bfq_\new$; $V_\new.\mathrm{inpath}\leftarrow \cP_\new$;
 $V_\new.\mathrm{interval}\leftarrow [\dot s_{\min},\dot s_{\max}]$
\RETURN $V_\new$
\ENDIF
\ENDIF
\RETURN \texttt{Failure}
\end{algorithmic}
\end{algorithm}

The other routines are defined as follows\,:
\begin{itemize}
\item CONNECT($V, \bfq_\goal$) attempts at connecting directly $V$ to
  the goal configuration $\bfq_\goal$, using the same algorithm as in
  lines 2 to 10 of Box~\ref{algo:extend}, but with the further
  requirement that the goal velocity is included in the final velocity
  interval;
\item COMPUTE\_TRAJECTORY($\cT,\bfq_\goal$) reconstructs the entire
  path $\cP_\textrm{total}$ from $\bfq_\start$ to $\bfq_\goal$ by
  recursively concatenating the $V$.inpath. Next, $\cP_\textrm{total}$
  is time-parameterized by applying TOPP. The existence of a valid
  time-parameterization is guaranteed by recursion by AVP.
\item NEAREST\_NEIGHBOR($\cT, \bfq$) returns the vertex of $\cT$ whose
  configuration is closest to configuration $\bfq$ in the metric $d$,
  see Section~\ref{sec:implementation} for a more detailed discussion.
\item INTERPOLATE($V,\bfq$) returns a pair $(\cP_\new,\bfq_\new)$ where
  $\bfq_\new$ is defined as follows
  \begin{itemize}
  \item if $d(V$.config,$\bfq)\leq R$ where $R$ is a user-defined
    extension radius as in the standard RRT
    algorithm~\citep{KL00icra}, then $\bfq_\new\leftarrow\bfq$;
  \item otherwise, $\bfq_\new$ is a configuration ``in the direction
    of'' $\bfq$ but situated within a distance $R$ of $V$.config
    (contrary to the standard RRT, it might not be wise to choose a
    configuration laying exactly on the segment connecting $V$.config
    and $\bfq$ since here one has also to take care of
    $C^1$-continuity, see below).
  \end{itemize}
  The path $\cP_\new$ is a smooth path connecting $V$.config and
  $\bfq_\new$, and such that the concatenation of $V$.inpath and
  $\cP_\new$ is $C^1$ at $V$.config, see
  Section~\ref{sec:implementation} for a more detailed discussion.
\end{itemize}

% \begin{algorithm}
%   \caption{Explanation of the other routines in Boxes~\ref{algo:rrt}
%     and~\ref{algo:extend}}
% \label{algo:explanations}
% \end{algorithm}

\subsection{Implementation and variations}
\label{sec:implementation}

As in the standard RRT~\citep{KL00icra}, some implementation choices
influence substantially the performance of the algorithm.

\begin{description}
\item[Metric] In state-space RRTs, the most critical choice is that
  of the metric $d$, in particular, the \emph{relative weighting}
  between configuration-space coordinates and velocity coordinates. In
  our approach, since the whole interval of valid path velocities is
  considered, the relative weighting does not come into play. 
  In practice, a simple Euclidean metric on the configuration
  space is often sufficient. However, in some applications, one may
  also include the \emph{final orientation} of $V$.inpath in the
  metric.

\item[Interpolation] In geometric path planners, the interpolation
  between two configurations is usually done using a straight
  segment. Here, since one needs to propagate velocities, it is
  necessary to enforce $C^1$-continuity at the junction point. In the
  examples of Section~\ref{sec:applications}, we used third-degree
  polynomials to do so. Other interpolation methods are possible\,:
  higher-order polynomials, splines, etc. The choice of the
  appropriate method depends on the application and plays an important
  role in the performance of the algorithm.

\item[K-nearest-neighbors] Attempting connection from $K$ nearest
  neighbors, where $K>1$ is a judiciously chosen parameter, has been
  found to improve the performance of RRT. To implement this, it
  suffices to replace line~2 of Box~\ref{algo:extend} with a FOR loop
  that enumerates the $K$ nearest neighbors. Note that this procedure
  is geared towards reducing the search time, not at improving the
  trajectory quality as in RRT$^*$~\citep{KF11ijrr}, see also below.

\item[Post-processing] After finding a solution trajectory, one can
  improve its quality (e.g. trajectory duration, trajectory
  smoothness, etc.), by repeatedly applying the following shortcutting
  procedure~\citep{GO07ijrr,Pha12mms}:
  \begin{enumerate}
  \item select two random configurations on the trajectory;
  \item interpolate a smooth shortcut path between these two
    configurations;
  \item time-parameterize the shortcut using TOPP;
  \item if the time-parameterized shortcut has shorter duration than
    the initial segment, then replace the latter by the former.
  \end{enumerate}

  Instead of shortcutting, one may also give the trajectory found by
  AVP-RRT as initial guess to a trajectory optimization algorithm, or
  implement a re-wiring procedure as in RRT$^*$~\citep{KF11ijrr},
  which has been shown to be asymptotically optimal in the context of
  state-based planning (note however that such re-wiring is not
  straightforward and might require additional algorithmic
  developments).

\end{description}

Another significant benefit of AVP is that one can readily adapt
heuristics that have been developed for geometric path planners. We
discuss two such heuristics below.

\begin{description}
\item[Bi-directional RRT] \citet{KL00icra} remarked that growing
  simultaneously two trees, one rooted at the initial configuration
  and one rooted at the goal configuration yielded significant
  improvement over the classical uni-directional RRT. This idea
  \citep[see also][]{NM91tra} can be easily implemented in the context
  of AVP-RRT as follows~\citep{LP14ijcai}\,:
  \begin{itemize}
  \item The start tree is grown normally as in
    Section~\ref{sec:avprrt};
  \item The goal tree is grown similarly, but using AVP-backward (see
    Section~\ref{sec:avpremarks}) for the velocity propagation step;
  \item Assume that one finds a configuration where the two trees are
    \emph{geometrically} connected. If the forward velocity interval
    of the start tree and the backward velocity interval of the goal
    tree have a non-empty intersection at this configuration, then the
    two trees can be connected \emph{dynamically}.
  \end{itemize}
  
\item[Bridge test] If two nearby configurations are in the obstacle
  space but their midpoint $\bfq$ is in the free space, then most
  probably $\bfq$ is in a narrow passage. This idea enables one to
  find a large number of such configurations $\bfq$, which is
  essential in problems involving narrow
  passages~\citep{HsuX03icra}. This idea can be easily implemented in
  AVP-RRT by simply modifying RANDOM\_CONFIG in line 6 of
  Box~\ref{algo:rrt} to include the bridge test.

\end{description}

One can observe from the above discussion that powerful heuristics
developed for geometric path planning can be readily used in
AVP-RRT, precisely because the latter is built on the idea of
path-velocity decomposition. It is unclear how such heuristics can be
integrated in other approaches to kinodynamic motion planning such as
the trajectory optimization approach discussed in the Introduction.

\section{Examples of applications}
\label{sec:applications}

As AVP-RRT is based on the classical Time-Optimal Path
Parameterization (TOPP) algorithm, it can be applied to any type of
systems and constraints TOPP can handle, from double-integrators
subject to velocity and acceleration bounds, to manipulator subject to
torque limits~\citep{BobX85ijrr,SM86tac}, to wheeled vehicles subject
to balance constraints~\citep{SG91tra}, to humanoid robots in
multi-contact tasks~\citep{PS15tmech}, etc. Furthermore, the overhead
for addressing a new problem is minimal\,: it suffices to reduce the
system constraints to the form of inequality~(\ref{eq:gen}), and
\emph{le tour est jou\'e!}

In this section, we present two examples where AVP-RRT was used to
address planning problems in which \emph{no quasi-static solution
  exists}. In the first example, the task consisted in swinging a
double pendulum into the upright configuration under severe torque
bounds. While this example does not fully exploit the advantages
associated with path-velocity decomposition (no configuration-space
obstacle nor kinematic closure constraint was considered), we chose it
since it was simple enough to enable a careful comparison with the
usual state-space planning approach~\citep{LK01ijrr}. In the second
example, the task consisted in transporting a bottle placed on a tray
through a small opening using a commercially-available manipulator (6
DOFs). This example demonstrates the full power of path-velocity
decomposition\,: geometric constraints (going through the small
opening) and dynamics constraints (the bottle must remain on the tray)
could be addressed separately. To the best of our knowledge, this is
the first successful demonstration on a non custom-built robot that
kinodynamic planning can succeed where quasi-static planning is
guaranteed to fail.

\subsection{Double pendulum with severe torque bounds}
\label{sec:pendulum}

We first consider a fully-actuated double pendulum (see
Figure~\ref{fig:pendulum}B), subject to torque limits
\[
|\tau_1|\leq \tau_1^{\max}, \qquad |\tau_2|\leq \tau_2^{\max}. 
\]
Such a pendulum can be seen as a 2-link manipulator, so that the
reduction to the form of~(\ref{eq:gen}) is straightforward, see
\citet{Pha14tro}.

\subsubsection{Obstruction to quasi-static planning}

The task consisted in bringing the pendulum from its initial state
$(\theta_1, \theta_2, \dot\theta_1, \dot\theta_2) = (0, 0, 0, 0)$
towards the upright state $(\theta_1, \theta_2, \dot\theta_1,
\dot\theta_2) = (\pi, 0, 0, 0)$, while respecting the torque
bounds. For simplicity, we did not consider self-collision issues.

Any trajectory that achieves the task must pass through a
configuration where $\theta_1=\pi/2$. Note that the configuration with
$\theta_1=\pi/2$ that requires the smallest torque at the first joint
to stay still is $(\theta_1, \theta_2) = (\pi/2, \pi)$.  Let then
$\tau_1^\mathrm{qs}$ be this smallest torque. It is clear that, if
$\tau_1^{\max} < \tau_1^\mathrm{qs}$, then \emph{no quasi-static
  trajectory} can achieve the task.

In our simulations, we used the following lengths and masses for the
links: $l=0.2$\,m and $m=8$\,kg, yielding
$\tau_1^\mathrm{qs}=15.68$\,N$\cdot$m. For information, the smallest
torque at the second joint to keep the configuration
$(\theta_1,\theta_2)=(0,\pi/2)$ stationary was $7.84$\,N$\cdot$m.  We
carried experiments in the following scenarii: $(\tau_1^{\max},
\tau_2^{\max}) \in \{(11,7),(13,5),(11,5) \}$ (N$\cdot$m).

\subsubsection{Solution using AVP-RRT} 
\label{sec:pend-res}

For simplicity we used the uni-directional version of AVP-RRT as
described in Section~\ref{sec:planning}, without any
heuristics. Furthermore, for fair comparison with state-space RRT in
Python (see Section~\ref{sec:comparison}), we used a Python
implementation of AVP rather than the C++ implementation contained in
the TOPP library~\citep{Pha14tro}.

\begin{figure}[htp]
    \centering
    \textbf{A}\hspace{5cm}\textbf{B}\\
    \includegraphics[height=4.7cm]{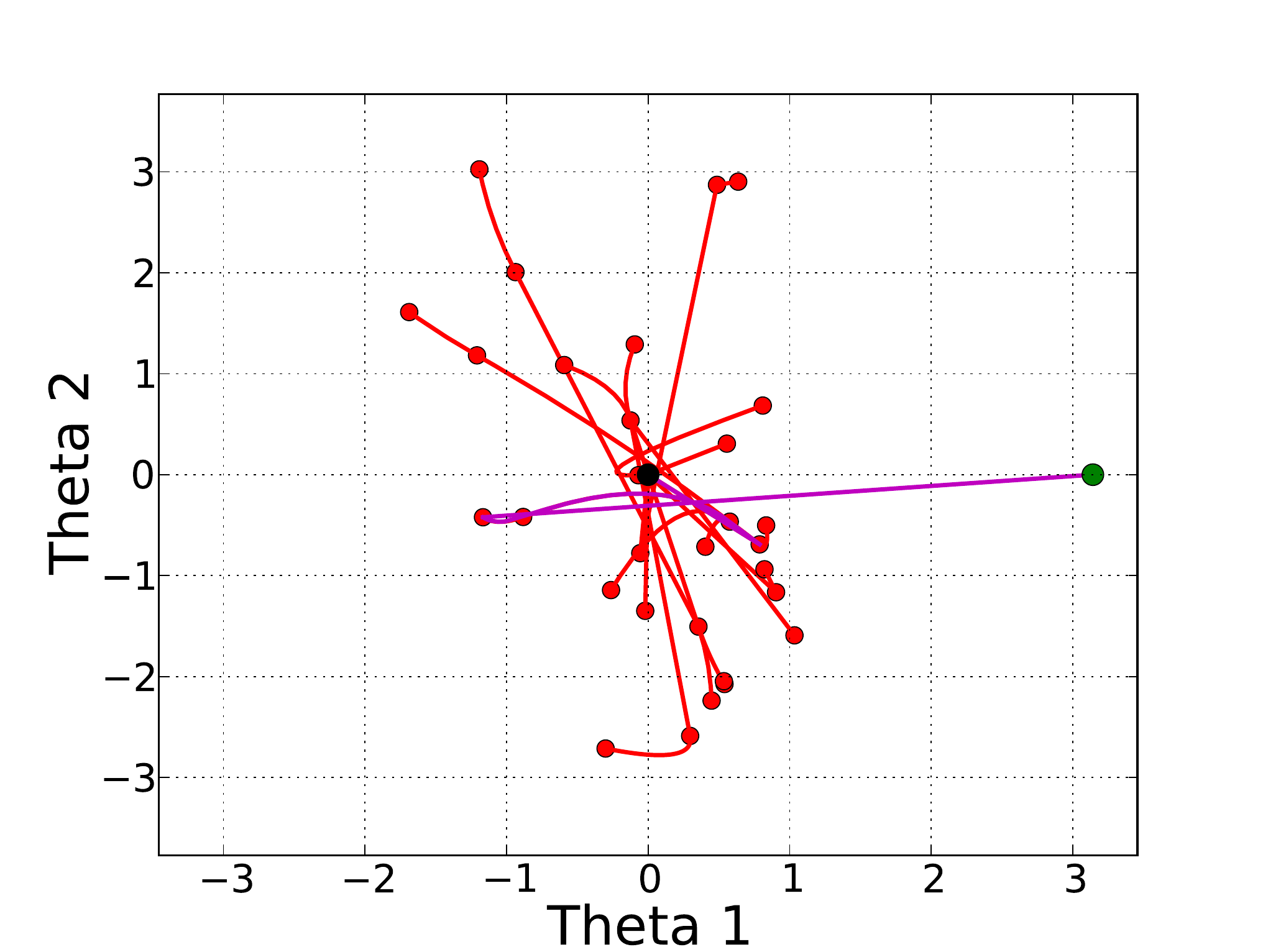}
    \includegraphics[height=4.7cm]{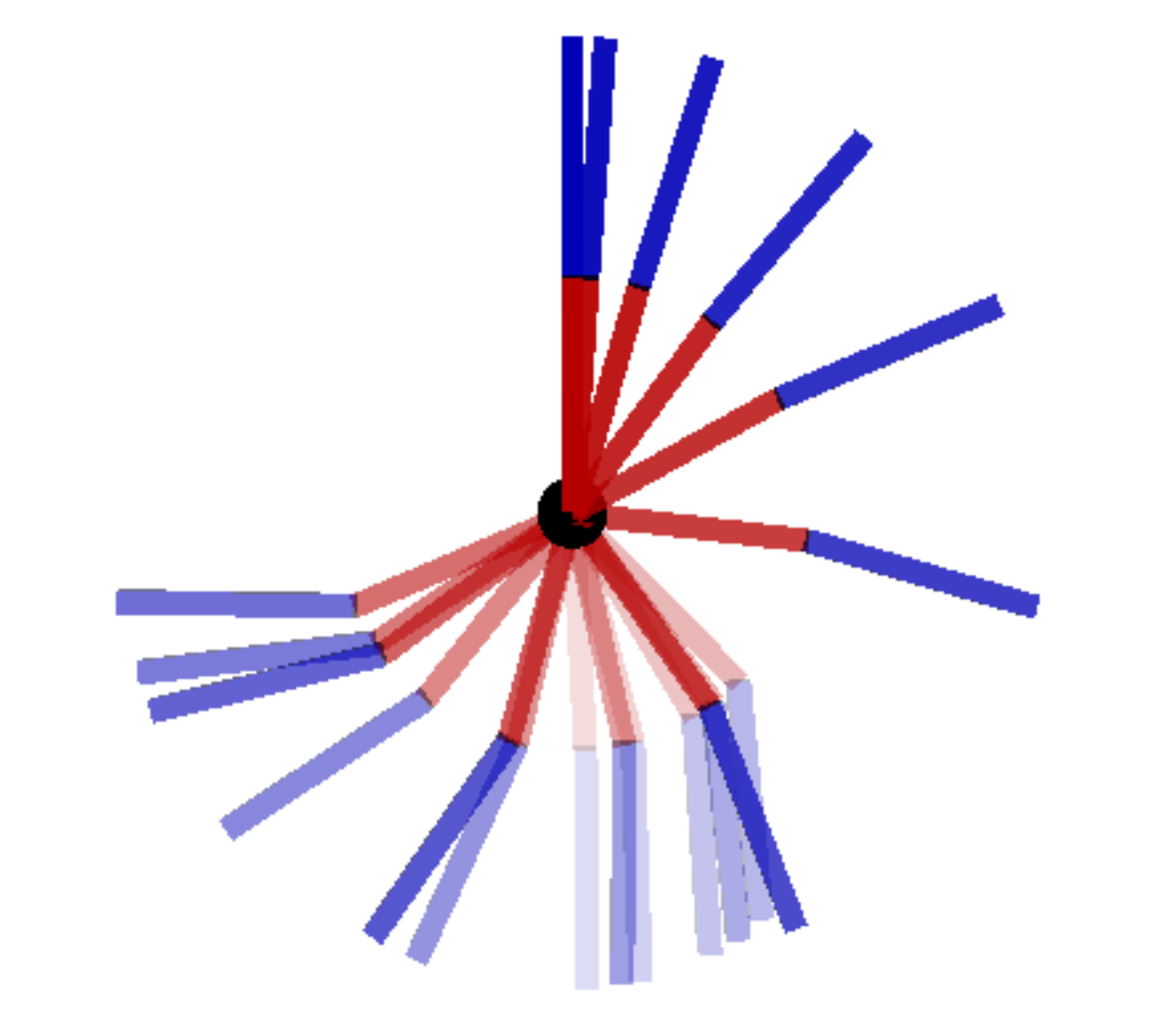}\\
    \textbf{C}\hspace{5cm}\textbf{D}\\
    \includegraphics[height=4.7cm]{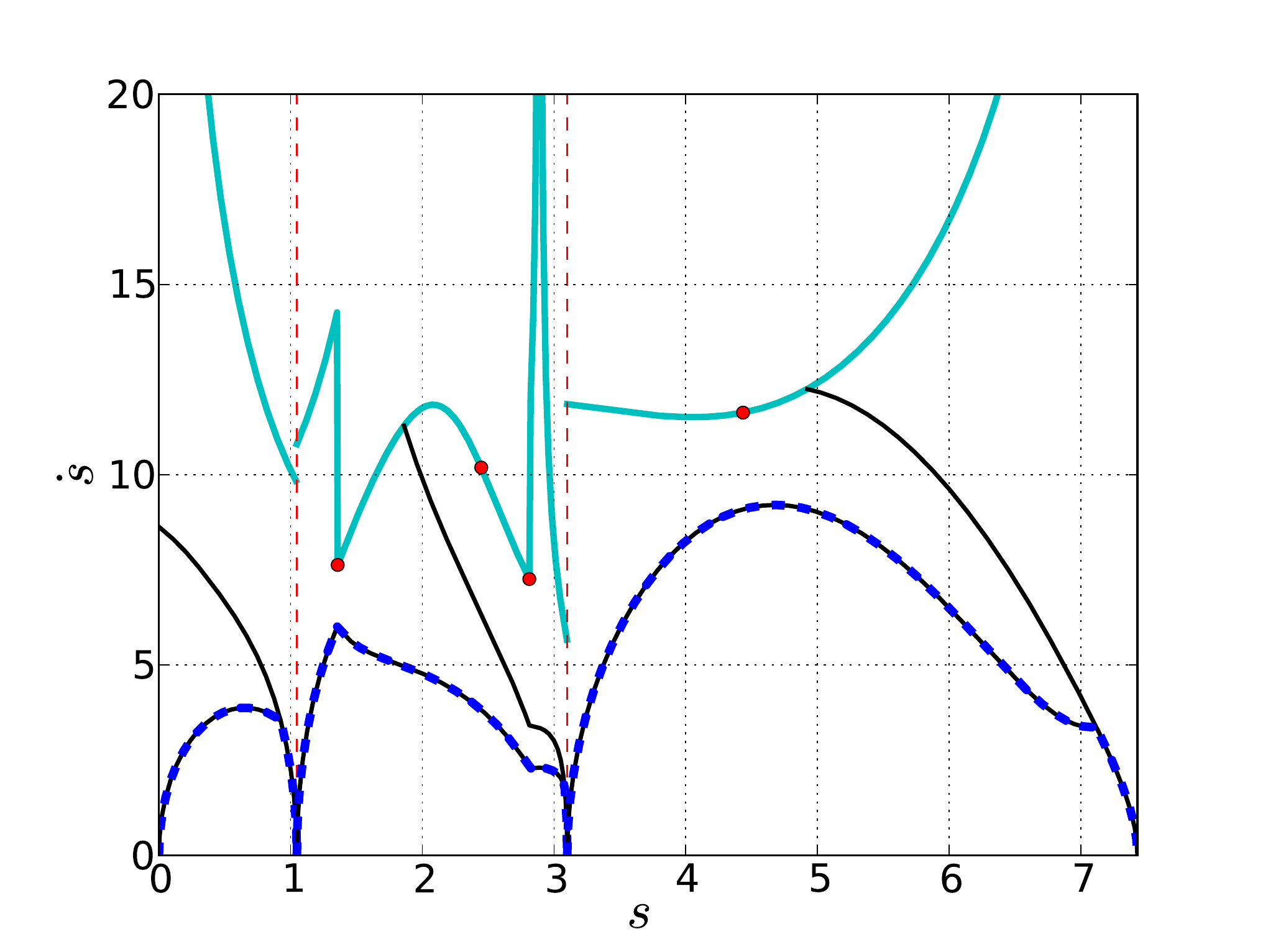}
    \includegraphics[height=4.7cm]{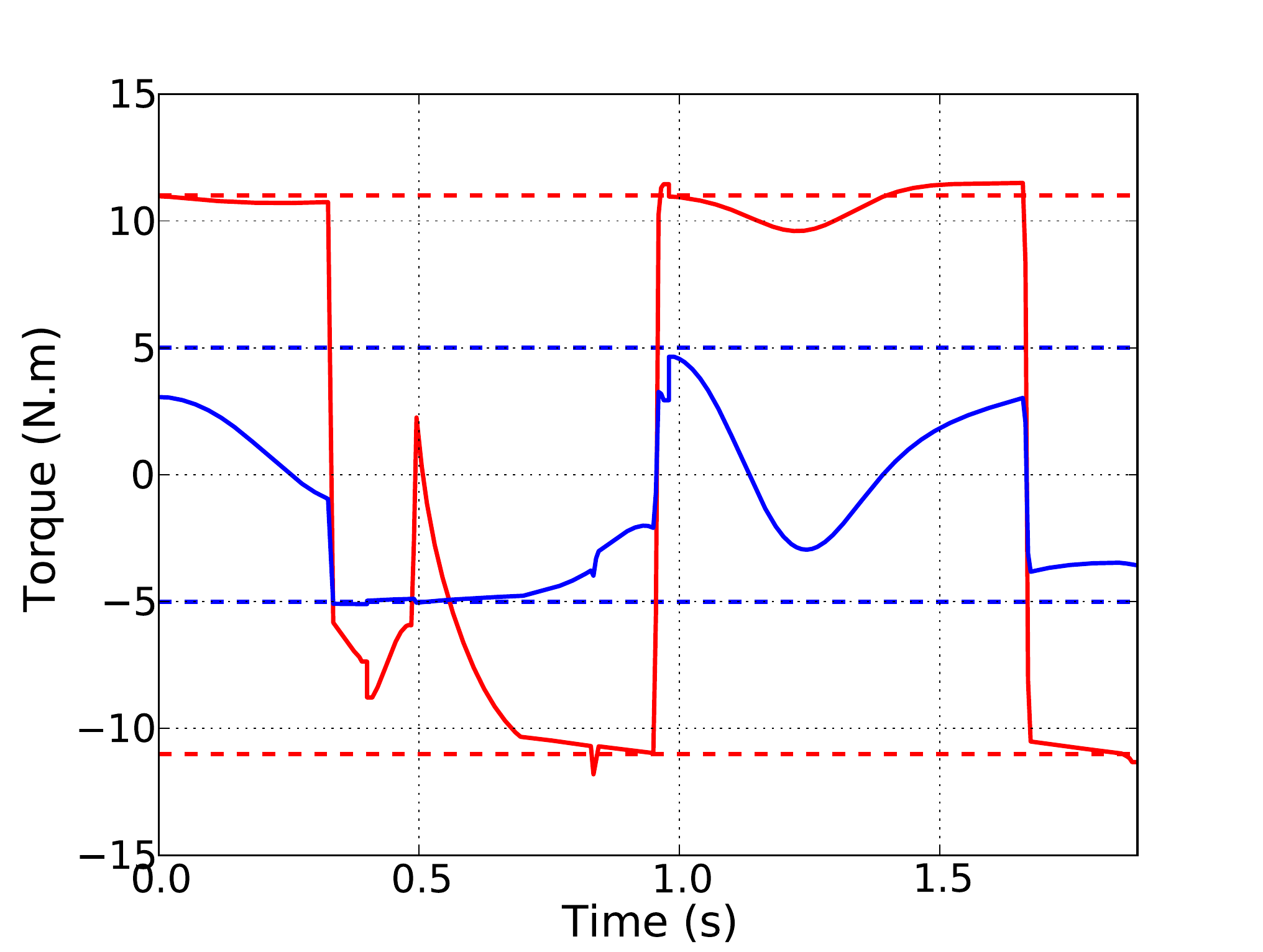}
    \caption{Swinging up a fully-actuated double pendulum. A typical
      solution for the case $(\tau_1^{\max},\tau_2^{\max})=(11,5)$
      N$\cdot$m, with trajectory duration 1.88\,s (see also the
      attached video). \textbf{A}: The tree in the
      $(\theta_1,\theta_2)$ space. The final path is highlighted in
      magenta. \textbf{B}: snapshots of the trajectory, taken every
      0.1\,s. Snapshots taken near the beginning of the trajectory are
      lighter. A video of the movement is available at
      \url{http://youtu.be/oFyPhI3JN00}. \textbf{C}: Velocity profiles
      in the $(s,\dot s)$ space. The MVC is in cyan. The various
      velocity profiles (CLC, $\Phi$, $\Psi$,
      cf. Section~\ref{sec:avp2}) are in black. The final, optimal,
      velocity profile is in dashed blue. The vertical dashed red
      lines correspond to vertices where 0 is a valid velocity, which
      allowed a discontinuity of the path tangent at that
      vertex. \textbf{D}: Torques profiles. The torques for joint 1
      and 2 are respectively in red and in blue. The torque limits are
      in dotted line. Note that, in agreement with time-optimal
      control theory, at each time instant, at least one torque limit
      was saturated (the small overshoots were caused by
      discretization errors).}
  \label{fig:pendulum}
\end{figure}

Regarding the number of nearest neighbors to consider, we chose
$K=10$. The maximum number of repetitions was set to
$N_\mathrm{maxrep}=2000$. Random configurations were sampled uniformly
in $[-\pi, \pi]^2$. A simple Euclidean metric in the configuration
space was used. Inverse Dynamics computations (required by the TOPP
algorithm) were performed using OpenRAVE~\citep{Dia10these}.  We ran
40 simulations for each value of $(\tau_1^{\max}, \tau_2^{\max})$ on a
2\,GHz Intel Core Duo computer with 2\,GB RAM. The results are given
in Table~\ref{tab:pend-results} and Figure~\ref{fig:pendulum}. A video
of some successful trajectories are shown at
\url{http://youtu.be/oFyPhI3JN00}.

\begin{table}[th]
  \caption{Results for the pendulum simulations}   
  \centering
{\footnotesize
    \begin{tabular}{|c|c|c|c|c|}
      \hline
      $\tau^{\max} $&Success&Configs&Vertices&Search time\\
      (N$\cdot$m)&rate&tested&added&(min)\\
      \hline
      (11,7)&100\%&64$\pm$44&31$\pm$23&4.2$\pm$2.7\\
      \hline
      (13,5)&100\%&92$\pm$106&29$\pm$30&5.9$\pm$6.3\\
      \hline
      (11,5)&92.5\%&212$\pm$327&56$\pm$81&12.1$\pm$15.0\\
      \hline
     \end{tabular}}
 \label{tab:pend-results}  
\end{table}

\subsubsection{Comparison with state-space RRT} 
\label{sec:comparison}

We compared our implementation of AVP-RRT with the standard
state-space RRT~\citep{LK01ijrr} including the $K$-nearest-neighbors
heuristic ($K$NN-RRT).  More complex kinodynamic planners have been
applied to low-DOF systems like the double pendulum, in particular
those based on locally linearized dynamics~\citep[such as
LQR-RRT$^*$][]{PerX12icra}. However, such planners require delicate
tunings and have not been shown to scale to systems with DOF $\geq$
4. 
% On the contrary, both RRT and AVP-RRT can be readily applied to any
% system for which one can compute inverse dynamics, and have been
% demonstrated on humanoids \citep{KufX02ar, PS15tmech}. 
The goal of the
present section is to compare the behavior of AVP-RRT to its RRT
counterpart on a low-DOF system. (In particular, we do not claim that
AVP-RRT is the best planner for a double pendulum.)

\begin{figure}[htp]
  \centering
  \textbf{A}\hspace{6cm}\textbf{B}\\
  \vspace{0.1cm}
  \includegraphics[width=6cm]{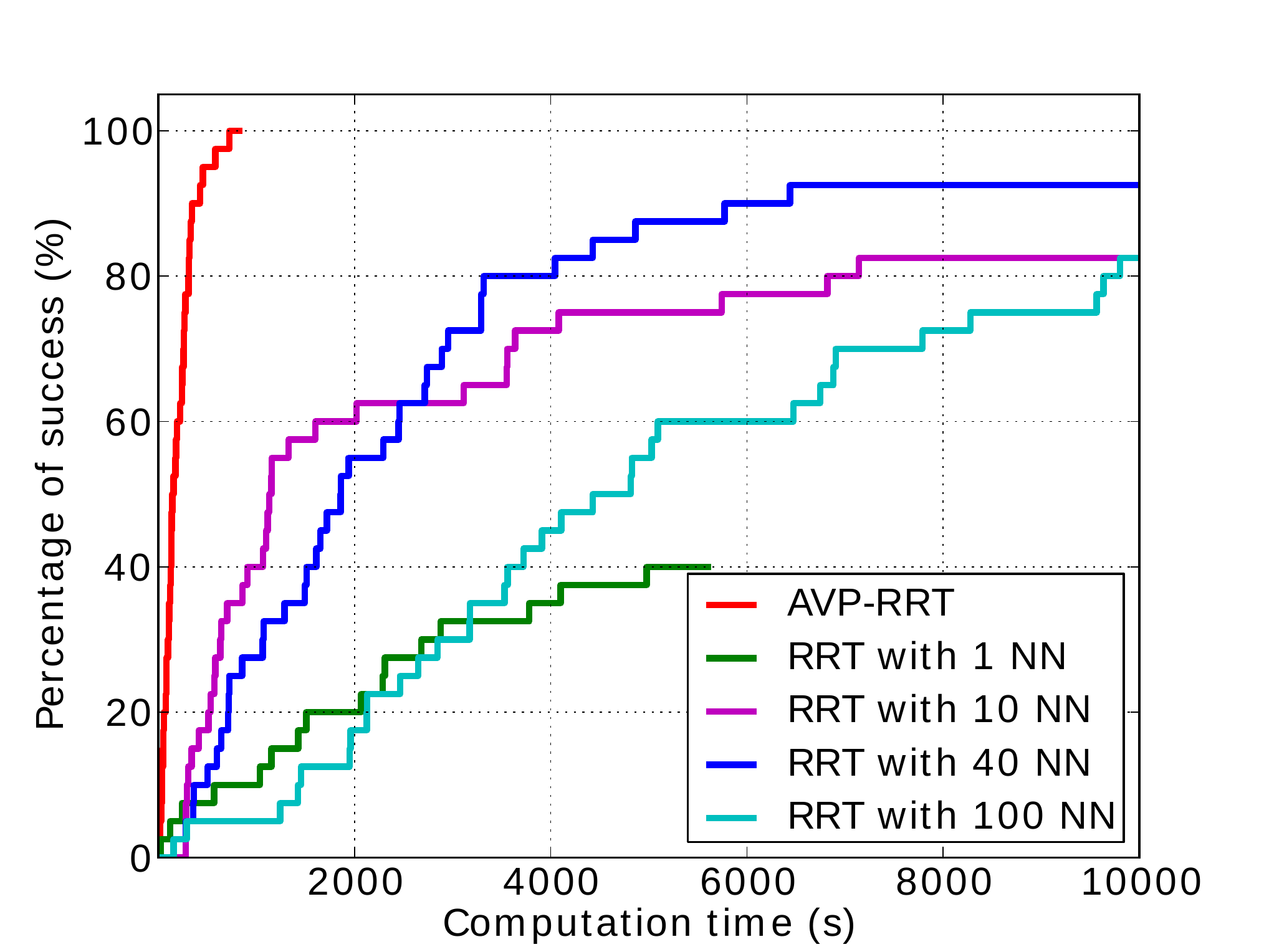}
  \includegraphics[width=6cm]{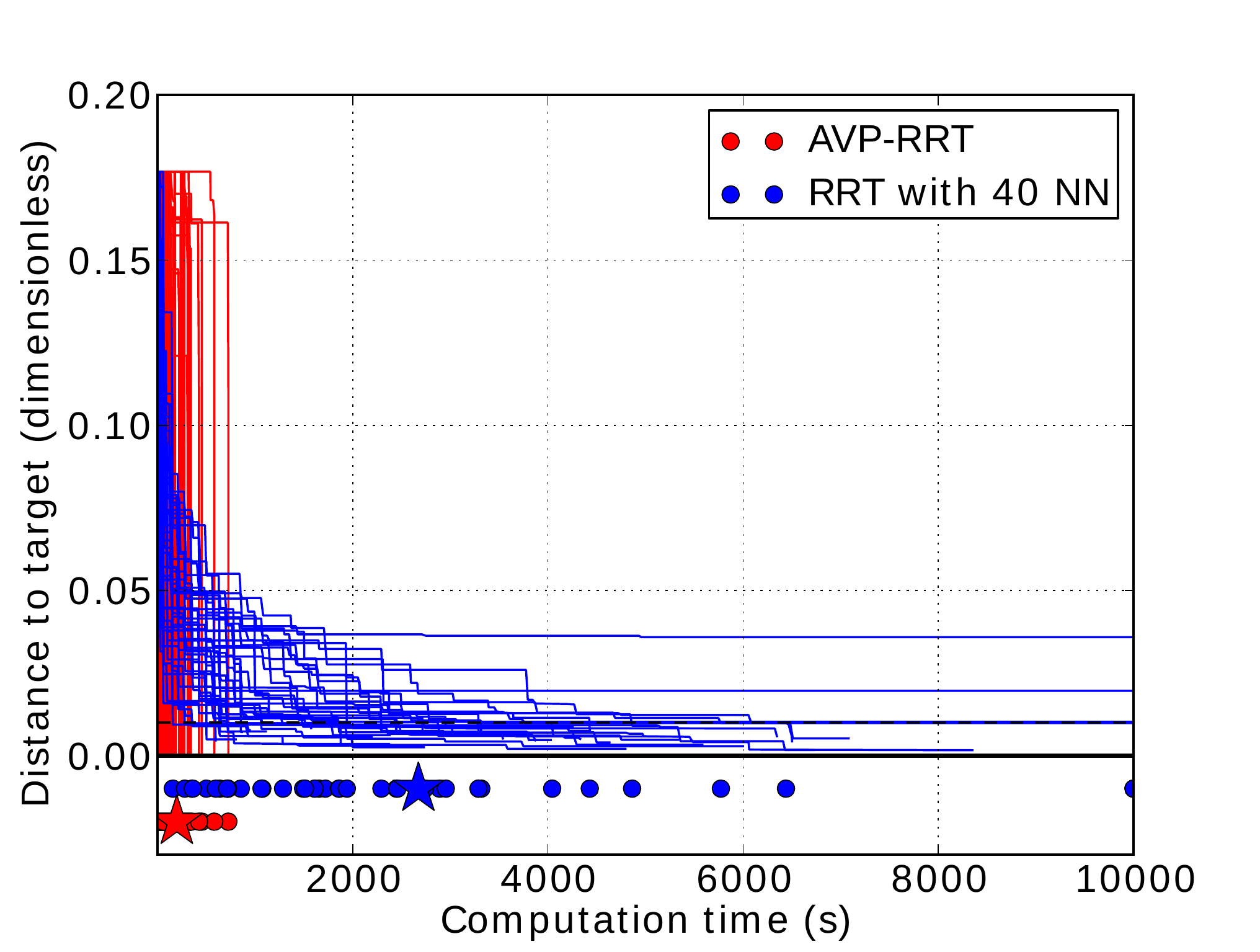}\\
  \vspace{0.2cm}
  \textbf{C}\hspace{6cm}\textbf{D}\\
  \vspace{0.1cm}
  \includegraphics[width=6cm]{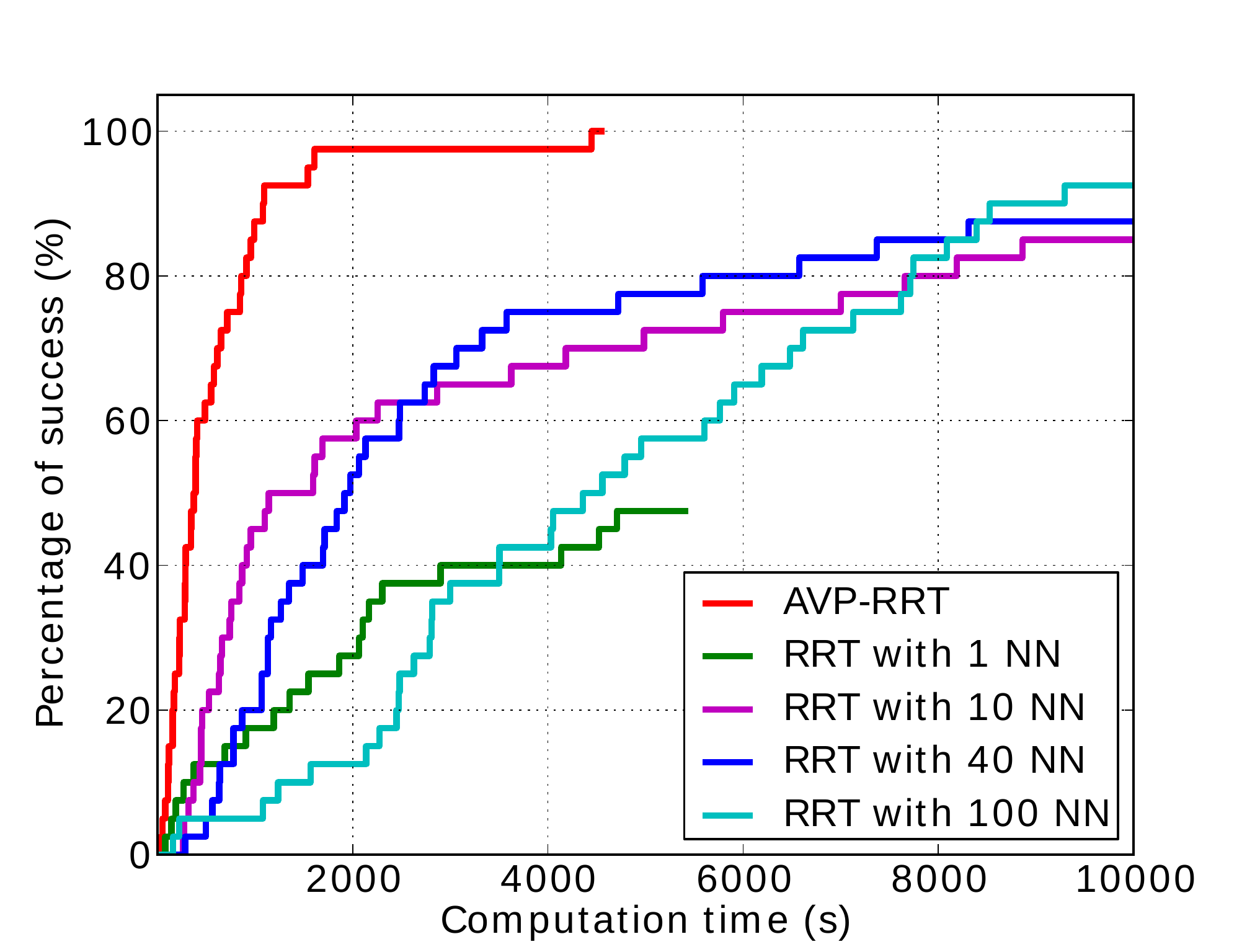} 
  \includegraphics[width=6cm]{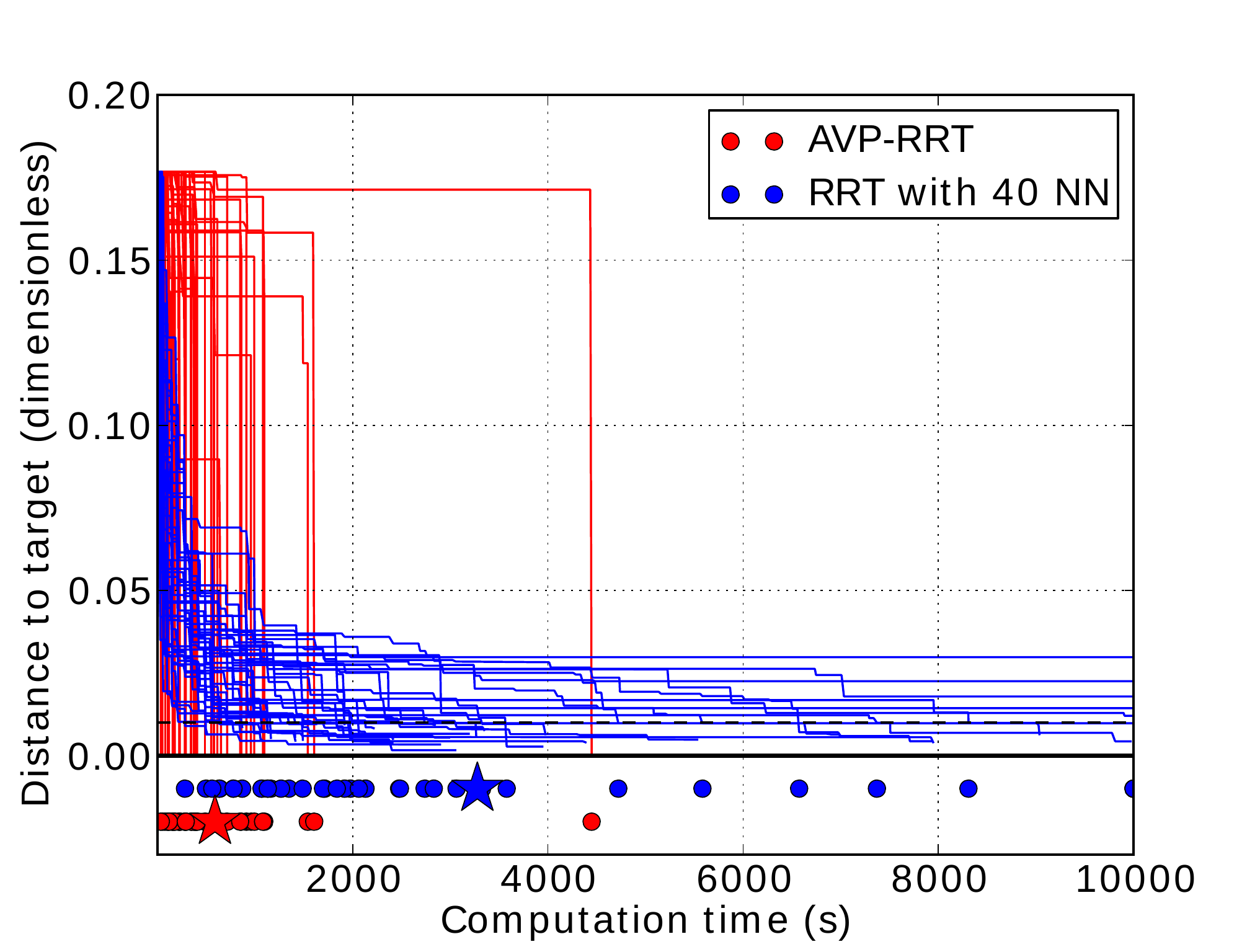}
  \caption{Comparison of AVP-RRT and $K$NN-RRT. \textbf{A}\,: Percentage
    of trials that have reached the goal area at given time instants
    for $\taumax = (11,7)$. \textbf{B}\,: Individual plots for each
    trial. Each curve shows the distance to the goal as a function of
    time for a given instance (red: AVP-RRT, blue: RRT-40).  Dots
    indicate the time instants when a trial successfully
    terminated. Stars show the mean values of termination
    times. \textbf{C} and \textbf{D}\,: same legends as A and B but
    for $\taumax = (11,5)$.}
  \label{fig:comp}
\end{figure}

We equipped the state-space RRT with generic heuristics that we tuned to the
problem at hand, see Appendix~A. In particular, we selected the best number of
neighbors $K$ for $K\in\{1,10,40,100\}$. Figure~\ref{fig:comp} and
Table~\ref{tab:comp} summarize the results.

\begin{table}[th]
  \centering
     \caption{Comparison of AVP-RRT and $K$NN-RRT} 
{\footnotesize
    \begin{tabular}{|c|c|c|c|c|}
      \hline
      & \multicolumn{2}{c|}{$\tau^{\max}=(11,7)$} &
      \multicolumn{2}{c|}{$\tau^{\max}=(11,5)$} \\
      \hline
      Planner&Success & Search time&Success &Search time\\
      &rate & (min)&rate &(min)\\
      \hline
      AVP-RRT&100\%&3.3$\pm$2.6&100\%&9.8$\pm$12.1\\
      \hline
      RRT-1&40\%&70.0$\pm$34.1&47.5\%&63.8$\pm$36.6\\
      \hline
      RRT-10&82.5\%&53.1$\pm$59.5&85\%&56.3$\pm$60.1\\
      \hline
      RRT-40&92.5\%&44.6$\pm$42.6&87.5\%&54.6$\pm$52.2\\
      \hline
      RRT-100&82.5\%&88.4$\pm$54.0&92.5\%&81.2$\pm$46.7\\
      \hline
     \end{tabular}}
 \label{tab:comp}  
\end{table}

In the two problem instances, AVP-RRT was respectively 13.4 and 5.6
times faster than the best $K$NN-RRT in terms of search time. We noted
however that the search time of AVP-RRT increased significantly from
instance $(\taumax_1,\taumax_2) = (11,5)$ to instance
$(\taumax_1,\taumax_2) = (11,7)$, while that of RRT only marginally
increased. This may be caused by the ``superposition'' phenomenon\,:
as torque constraints become tighter, more ``pumping'' swings are
necessary to reach the upright configuration. However, since our
metric was only on the configuration-space variables, configurations
with different speeds (corresponding to different pumping cycles) may
become indistinguishable. While this problem could be addressed by
including a measure of reachable velocity intervals into the metric, 
we chose not to do so in the present paper in order to avoid 
over-fitting our implementation of AVP-RRT to the problem at hand. 
Nevertheless, AVP-RRT still significantly over-performed 
the best $K$NN-RRT.

\subsection{Non-prehensile object transportation}
\label{sec:bottle}

Here we consider the non-prehensile (i.e. without grasping)
transportation of a bottle, or ``waiter motion''. Non-prehensile
transportation can be faster and more efficient than prehensile
transportation since the time-consuming grasping and un-grasping
stages are entirely skipped. Moreover, in many applications, the
objects to be carried are too soft, fragile or small to be adequately
grasped (e.g. food, electronic components, etc.)

\subsubsection{Obstruction to quasi-static planning} 

A plastic milk bottle partially filled with sand was placed (without
any fixation device) on a tray. The mass of the bottle was 2.5\,kg,
its height was 24\,cm (the sand was filled up to 16\,cm) and its base
was a square of size 8\,cm $\times$ 8\,cm. The tray was mounted as the
end-effector of a 6-DOF serial manipulator (Denso VS-060). The task
consisted in bringing the bottle from an initial configuration towards
a goal configuration, these two configurations being separated by a
small opening (see Fig.~\ref{fig:bottle}A).

For the bottle to remain stationary with respect to the tray, the
following three conditions must be satisfied\,:
\begin{itemize}
\item (Unilaterality) The normal component $f_n$ of the reaction force
  must be non-negative; 
\item (Non-slippage) The tangential component $\bff_t$ of the reaction
  force must satisfy $\|\bff_t\| \leq \mu f_n$, where $\mu$ is the
  static friction coefficient between the bottle and the tray. In our
  experimental set-up, the friction coefficient was set to a high
  value ($\mu=1.7$), such that the non-slippage condition was never
  violated before the ZMP condition;
\item (ZMP) The ZMP of the bottle must lie inside the bottle
  base~\citep{VukX01humanoids}.
\end{itemize}
The height of the opening was designed so that, for the bottle to go
through the opening, it must be tilted by at least an angle
$\theta^\mathrm{qs}$. However, when the bottle is tilted by that
angle, the center of mass (COM) of the bottle projects outside of the
bottle base. As the projection of the COM coincides with the ZMP in
the quasi-static condition, tilting the bottle by the angle
$\theta^\mathrm{qs}$ thus violates the ZMP condition and as a result,
the bottle will tip over. One can therefore conclude that \emph{no
  quasi-static motion} can bring the bottle through the opening
without tipping it over.

\subsubsection{Solution using AVP-RRT} 

We first reduced the three aforementioned conditions to the form
of~(\ref{eq:gen}). Details of this reduction can be found
in~\citet{LP14ijcai}.  We next used the bi-directional version of
AVP-RRT presented in Section~\ref{sec:implementation}. All vertices in
the tree were considered for possible connection from a new random
configuration, but they were sorted by increasing distance from the
new configuration (a simple Euclidean metric in the configuration
space was used for the distance computation). As the opening was very
small (narrow passage), we made use of the bridge
test~\citep{HsuX03icra} in order to automatically sample a sizable
number of configurations inside or close to the opening. Note that the
use of the bridge test was natural thanks to path-velocity
decomposition.

\begin{figure}[htp]
    \centering
    \textbf{A} \hspace{4.0cm} \textbf{B} \hspace{4.0cm} \textbf{C}\\
    \vspace{0.1cm}
    \includegraphics[height=3.25cm]{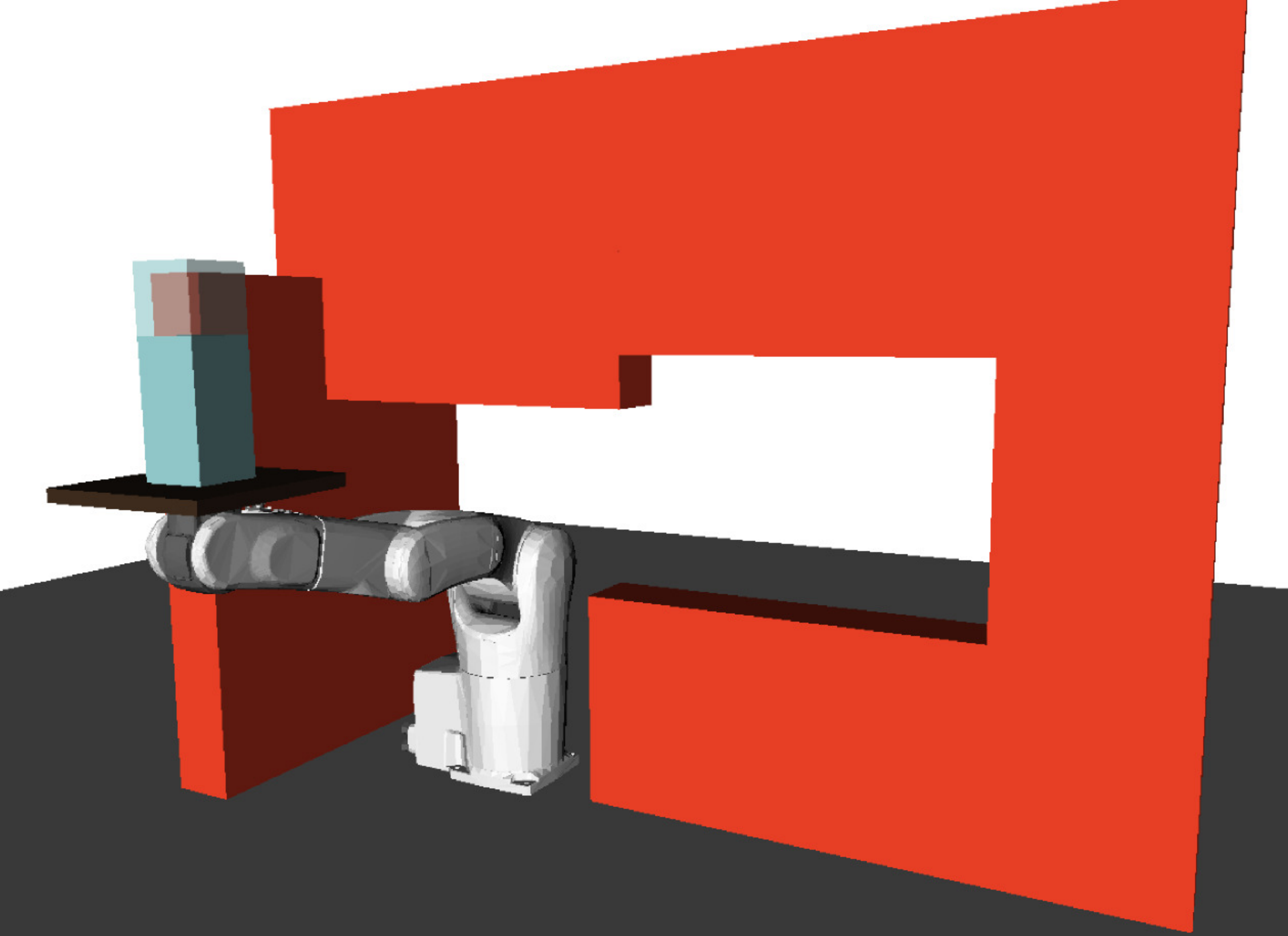}
    \includegraphics[height=3.25cm]{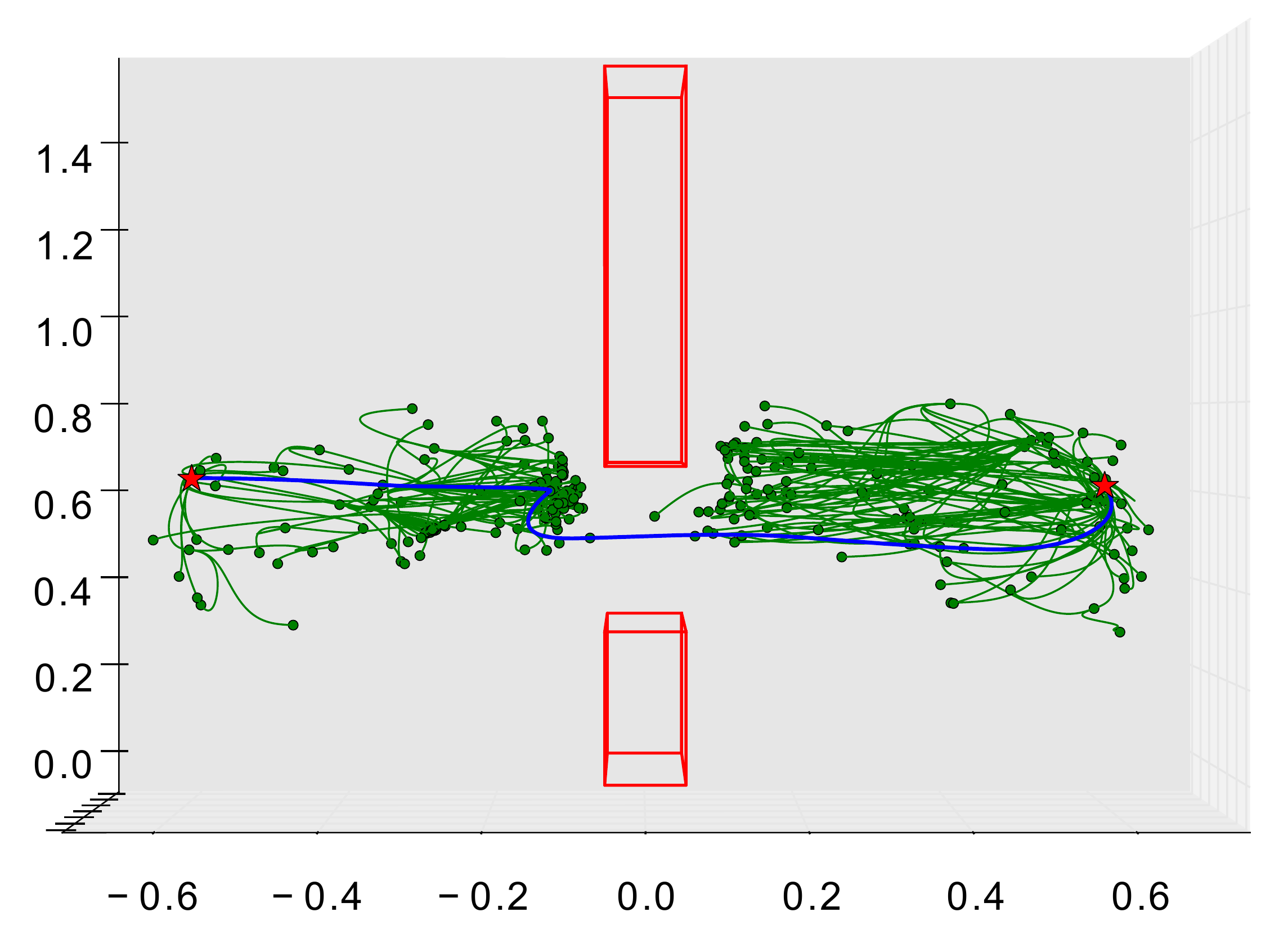}
    \includegraphics[height=3.25cm]{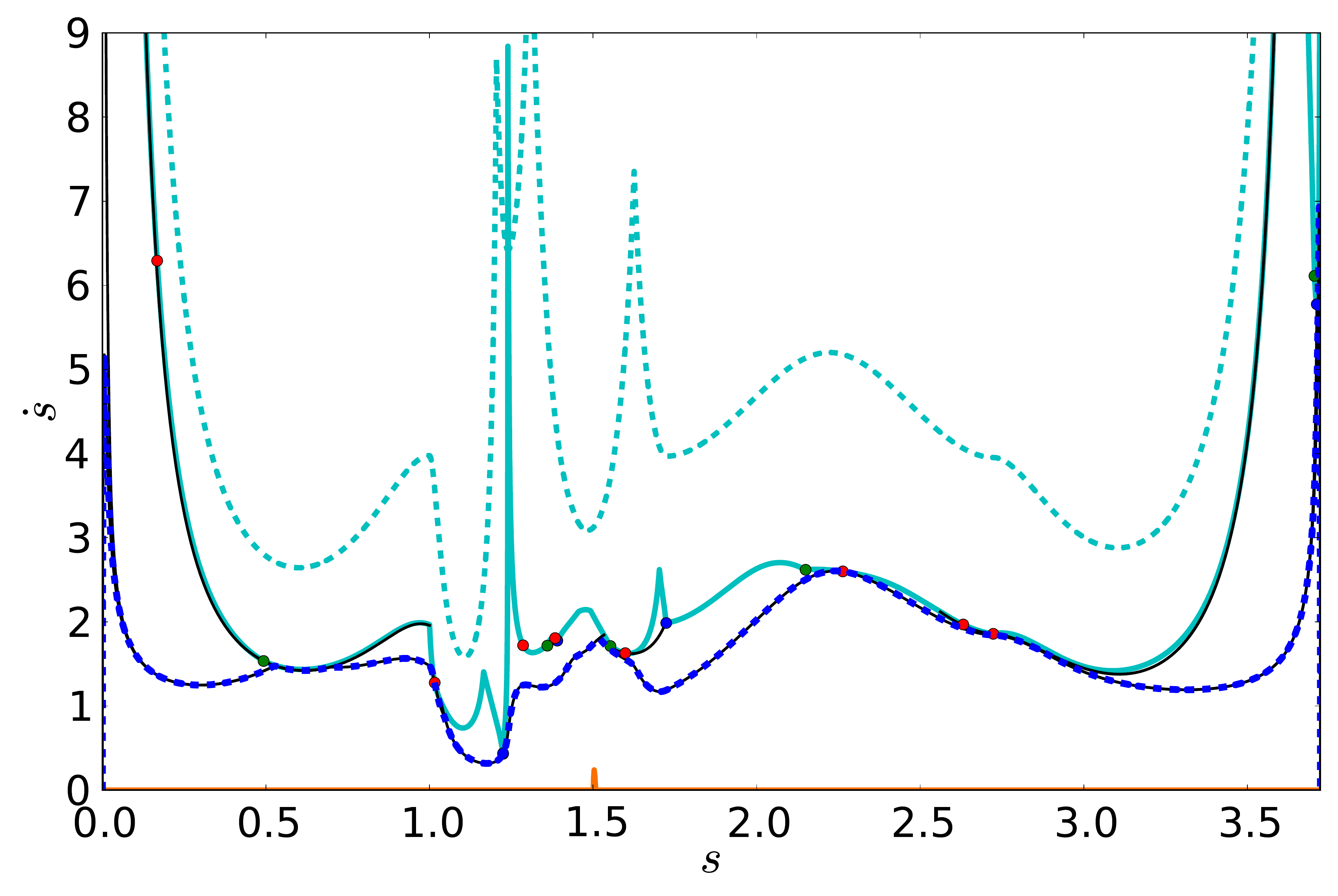}
    \vspace{0.2cm}\\
    \textbf{D}  \hspace{4.0cm} \textbf{E} \hspace{4.0cm}  \textbf{F} \\
    \vspace{0.1cm}
    \includegraphics[width=4.7cm]{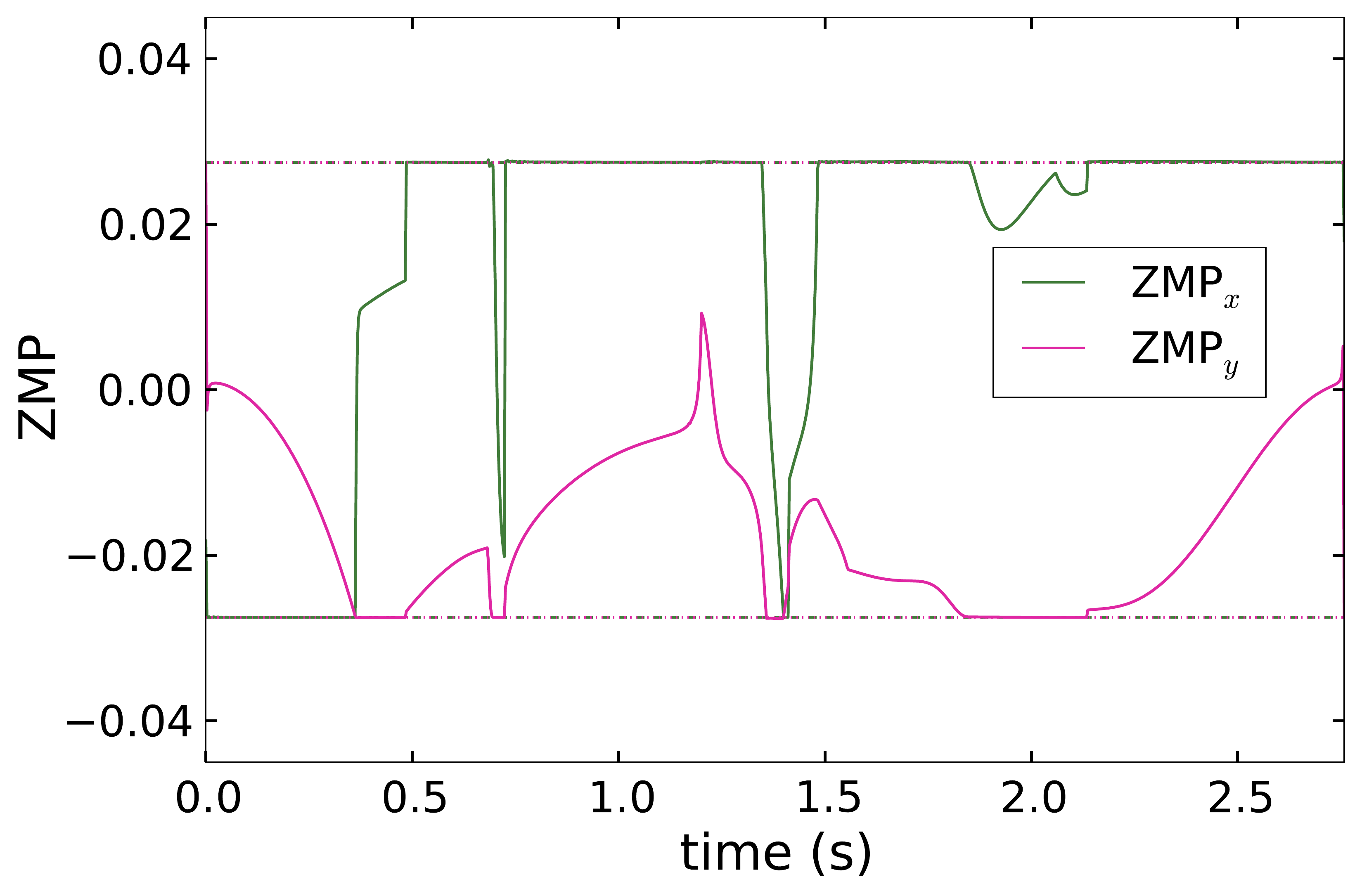}
    \includegraphics[width=4.7cm]{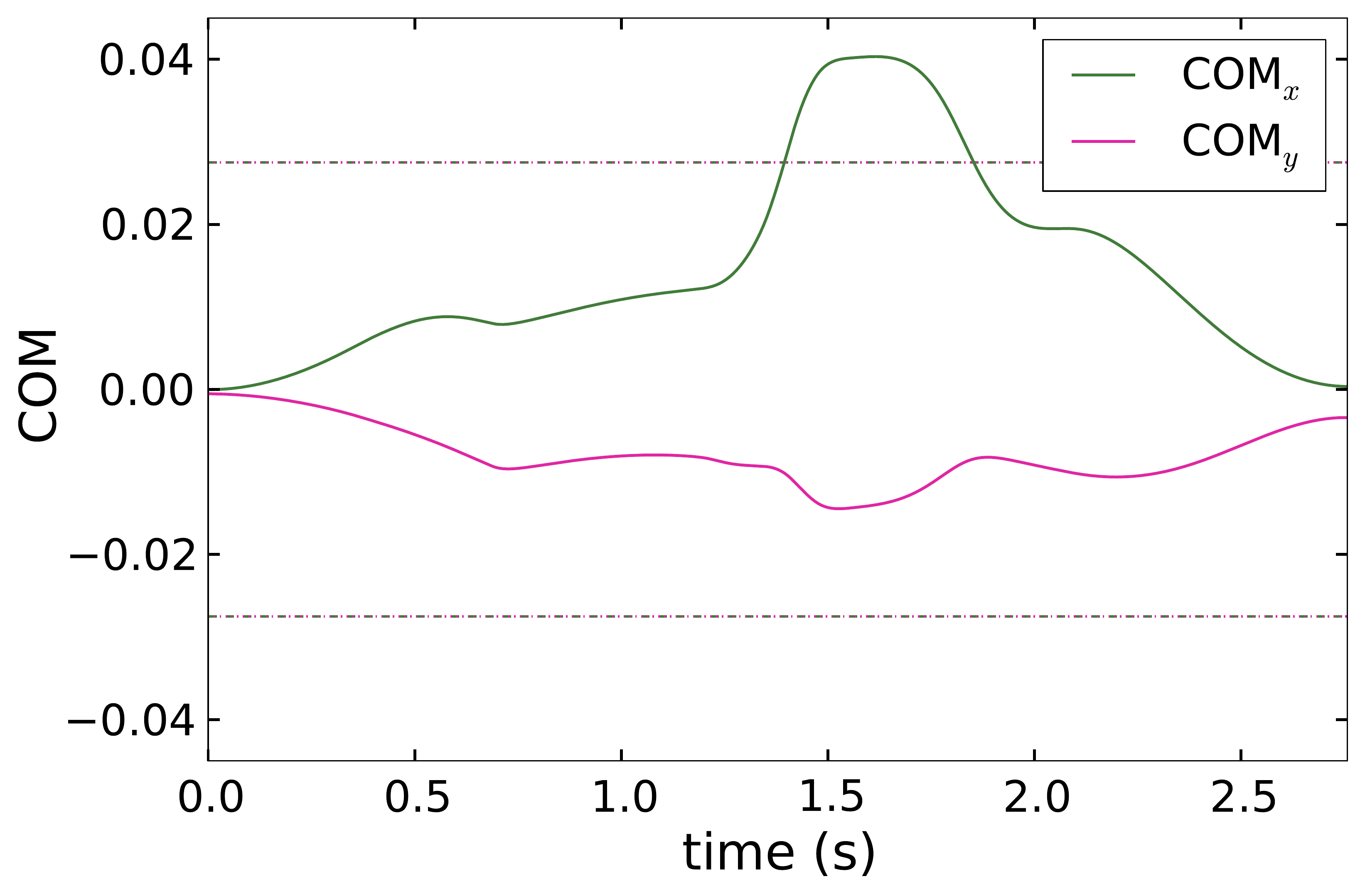}
    \includegraphics[width=4.7cm]{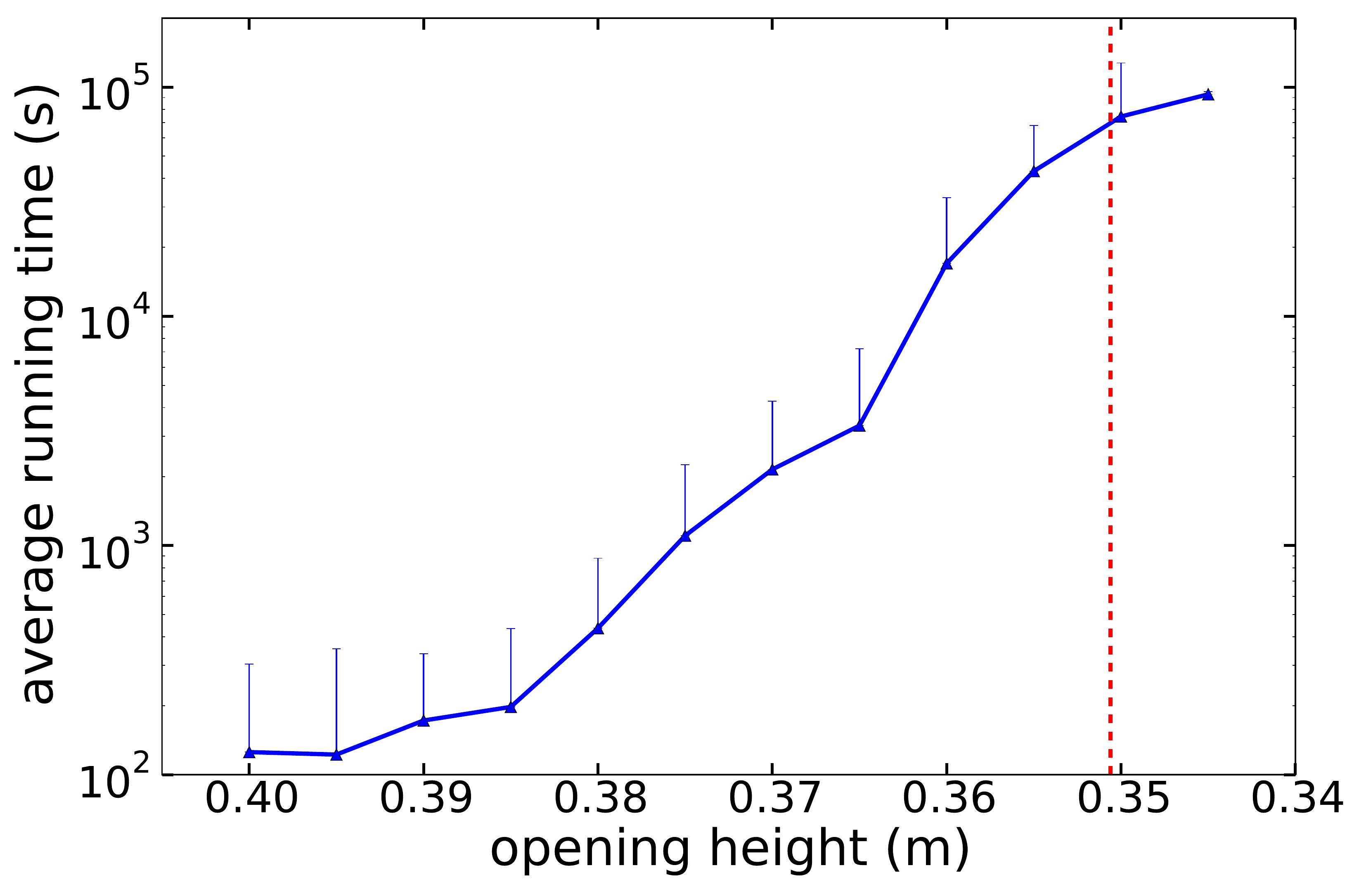}
    \caption{Non-prehensile transportation of a bottle. \textbf{A}\,:
      Simulation environment. The robot must bring the bottle to the
      other side of the opening while keeping it balanced on the
      tray. \textbf{B}\,: Bi-RRT tree in the workspace\,: the start
      tree had 125 vertices and the goal tree had 116 vertices.  Red
      boxes represent the obstacles. Red stars represent the initial
      and goal positions of the bottle COM.  Green lines represent the
      paths of the bottle COM in the tree. The successful path is
      highlighted in blue\,: it had 6 vertices. \textbf{C}\,: MVC and
      velocity profiles in the $(s,\dot s)$ space. Same legend as in
      Fig.~\ref{fig:pendulum}C. \textbf{D}\,: ZMP of the bottle in the
      tray reference frame (RF) for the successful trajectory. Note
      that the ZMP always stayed within the imposed safety borders
      $\pm 2.75$\,cm (the actual borders were $\pm
      4$\,cm). \textbf{E}\,: COM of the bottle in the tray RF for the
      successful trajectory. Note that the X-coordinate of the COM
      reached the maximum value of 4.03\,cm, around the moment when
      the bottle went through the opening, indicating that the
      successful trajectory would not be quasi-statically feasible.
      \textbf{F}\,: Here we varied the opening height (X-axis, from
      left to right\,: higher opening to lower opening) and determine
      the average and standard deviation of computation time (Y-axis,
      logarithmic scale) required to find a solution. We carried out
      30 runs for opening heights from 0.4m to 0.365m, 10 runs for
      0.36m, 3 runs for 0.355m and 0.35m and 2 runs for 0.345m. The
      red dashed vertical line indicates the critical height below
      which no quasi-static trajectory was possible. Here, we used
      $\pm 4$\,cm as boundaries for the ZMP, so that the computed
      motions, while theoretically feasible, might not be actually
      feasible.}
  \label{fig:bottle}
\end{figure}

\begin{figure}[htp]
  \centering
  \vspace{0.4cm}
  \includegraphics[width=3.5cm]{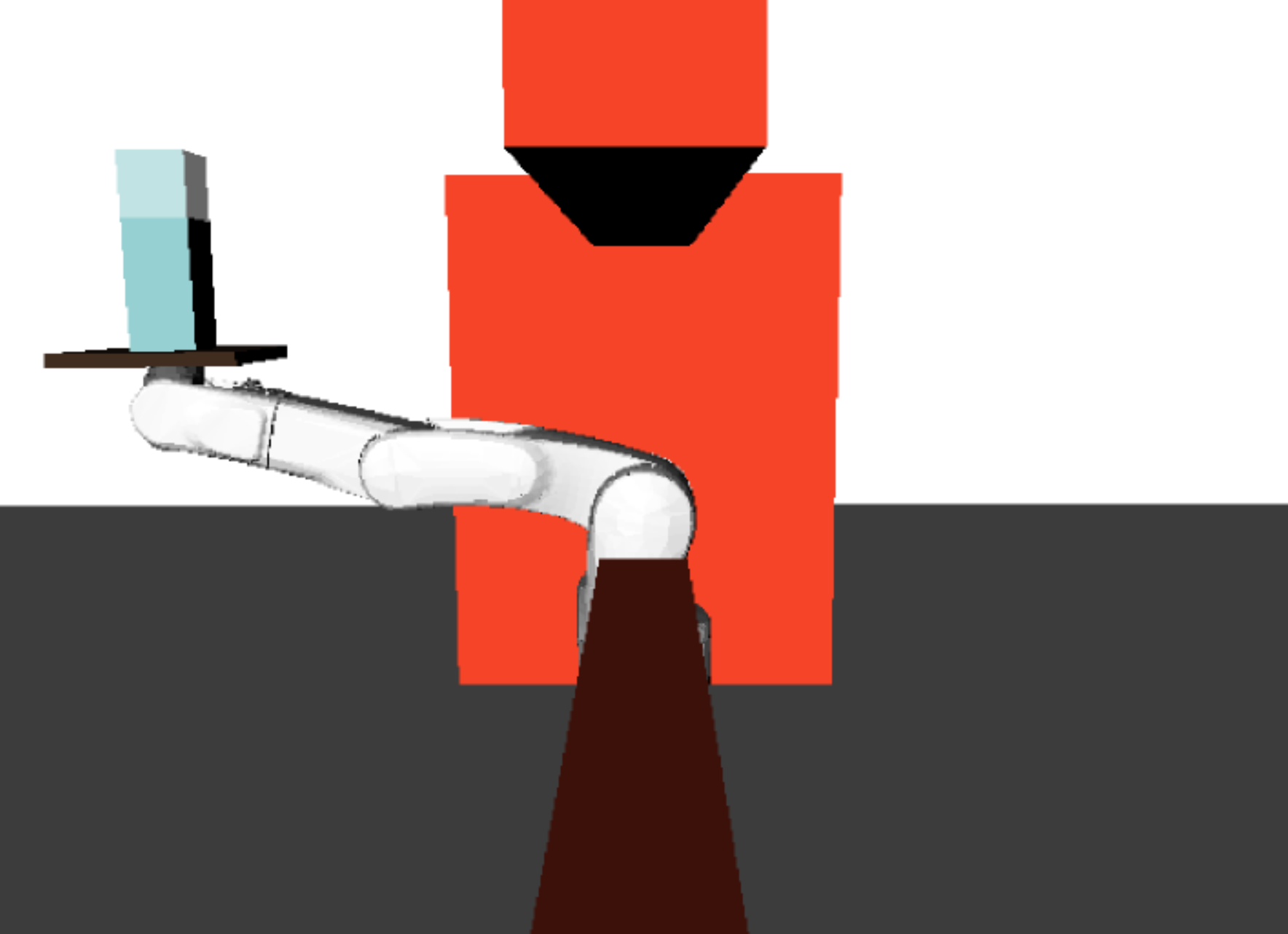}  
  \includegraphics[width=3.5cm]{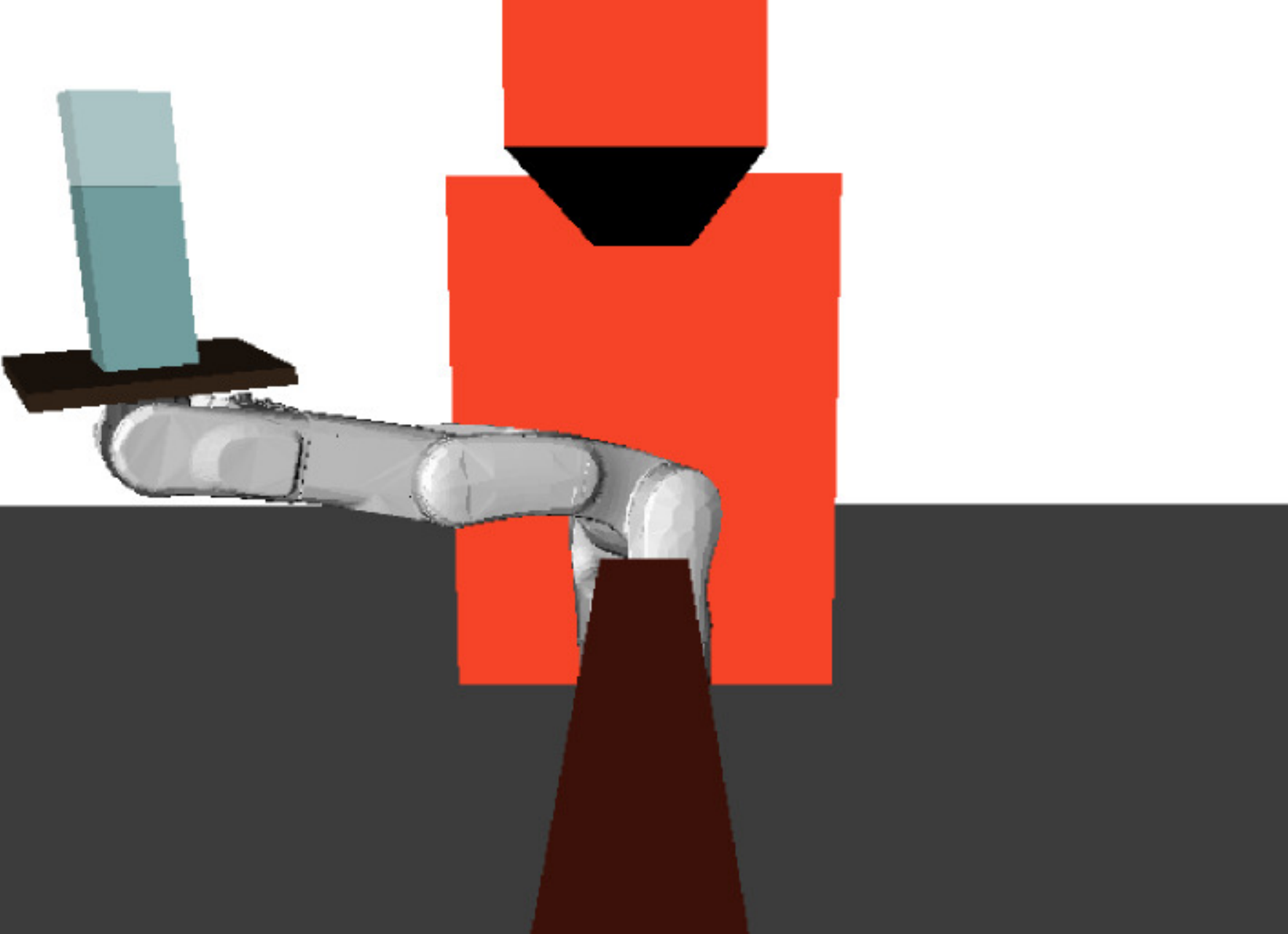}
  \includegraphics[width=3.5cm]{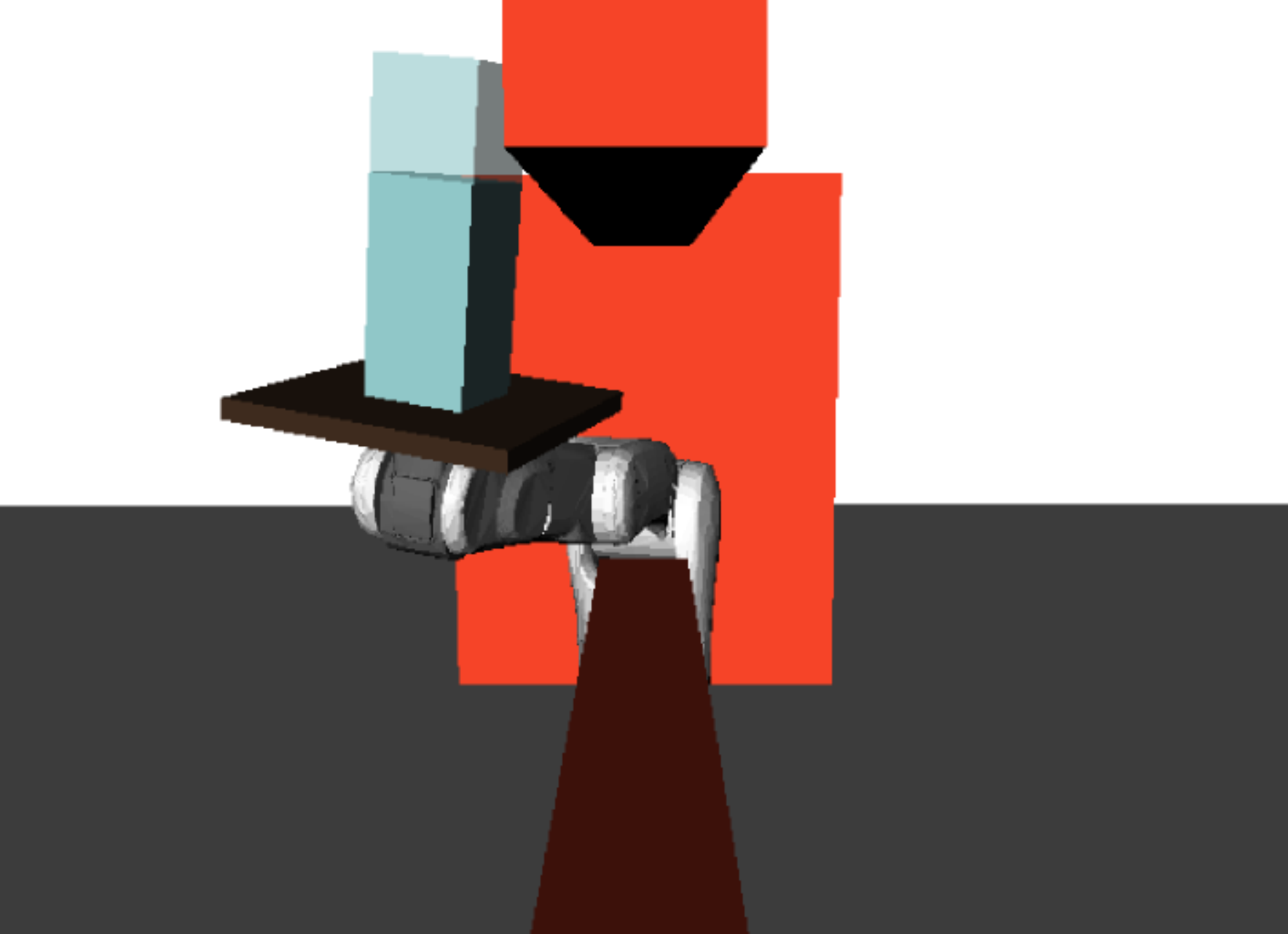}
  \includegraphics[width=3.5cm]{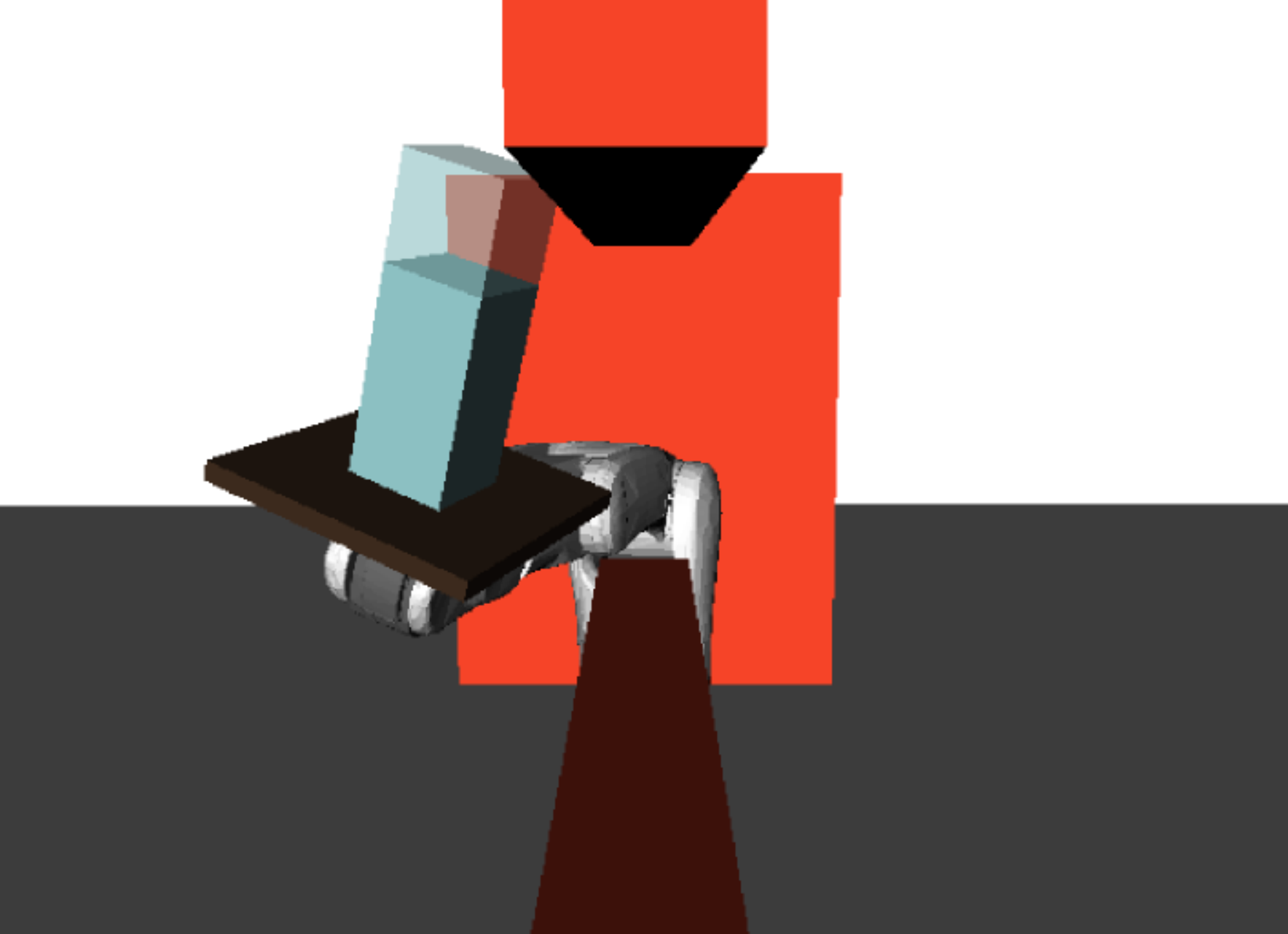}\\
  \vspace{0.2cm}
  \includegraphics[width=3.5cm]{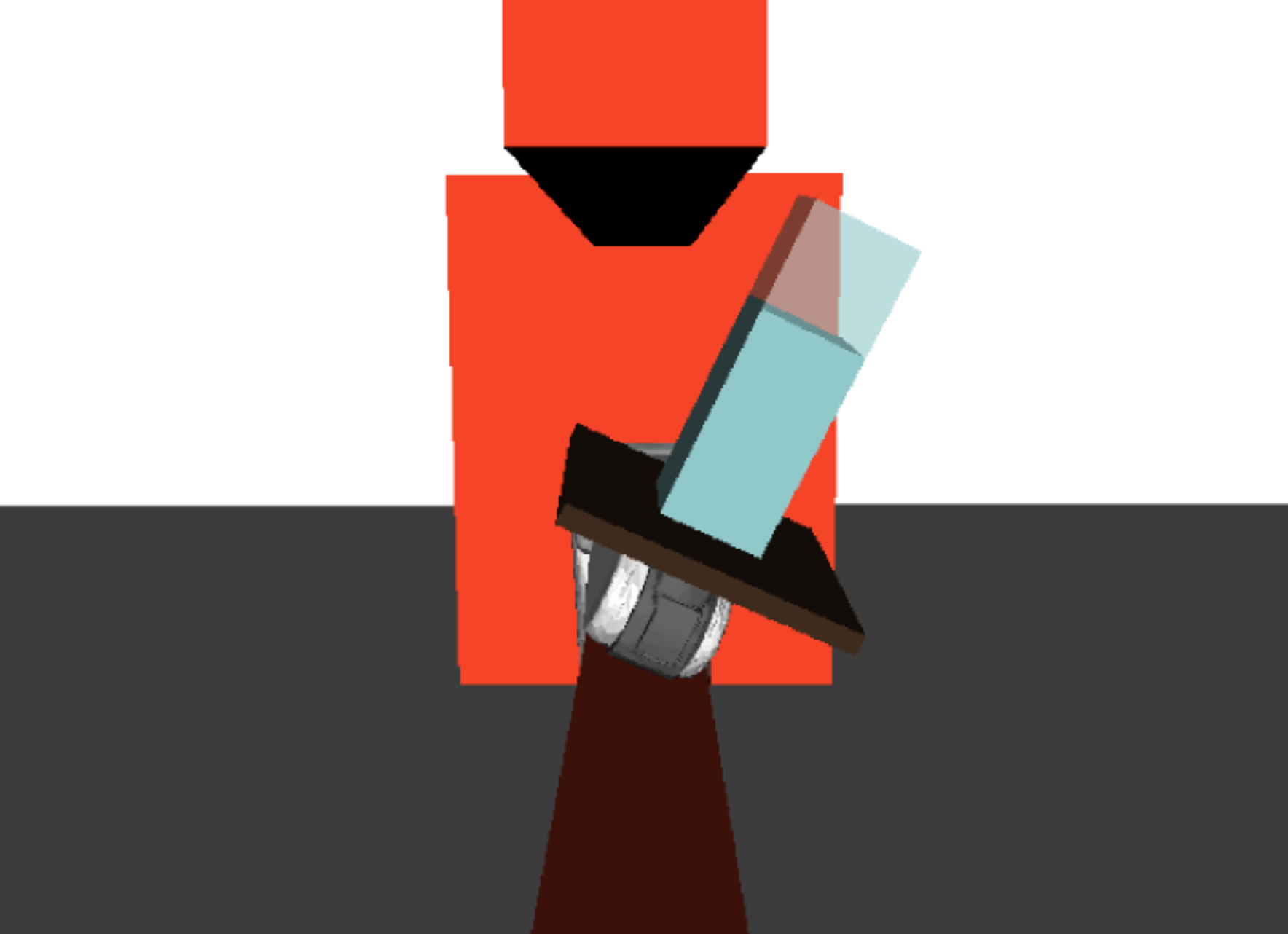}
  \includegraphics[width=3.5cm]{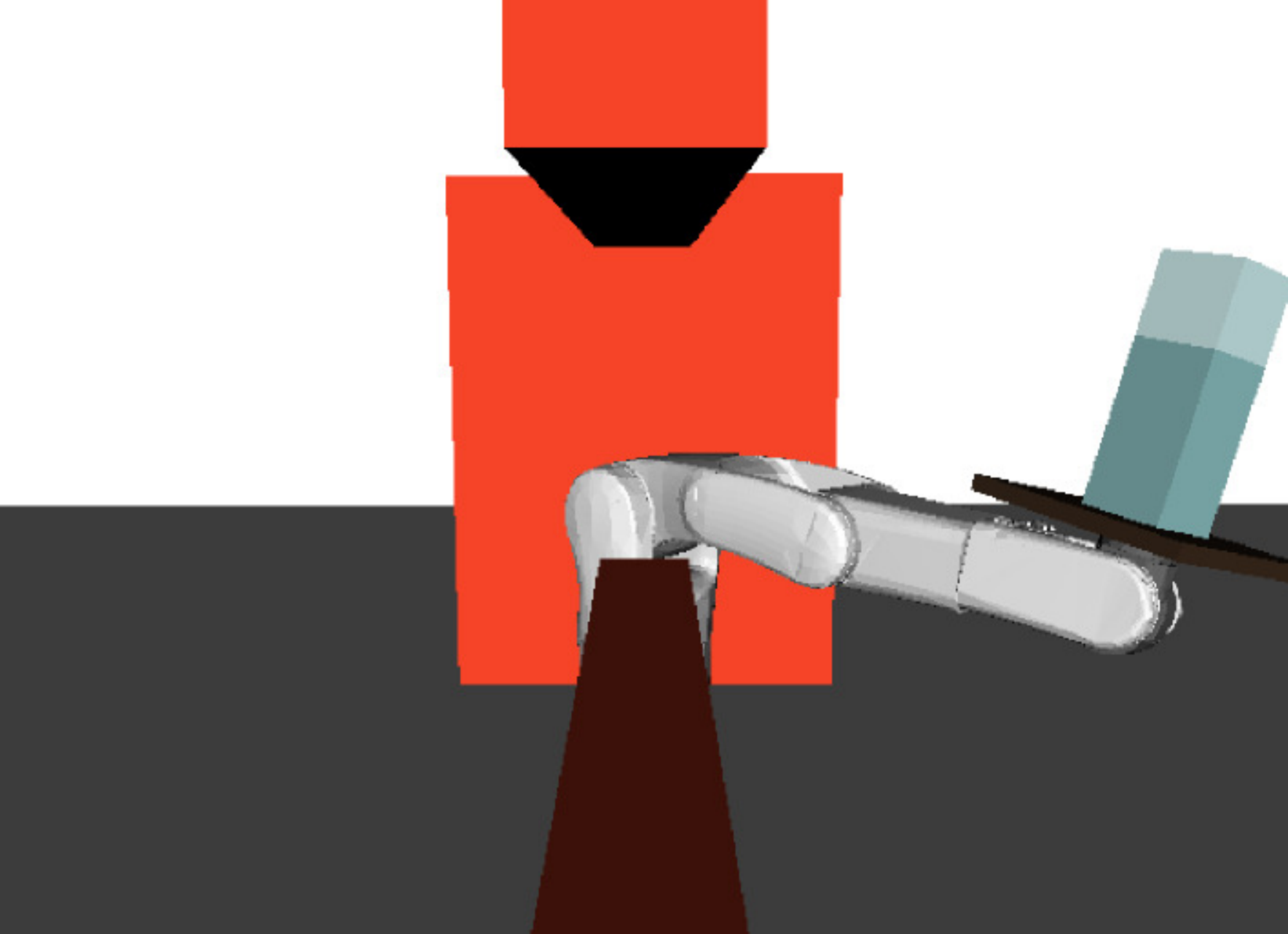}
  \includegraphics[width=3.5cm]{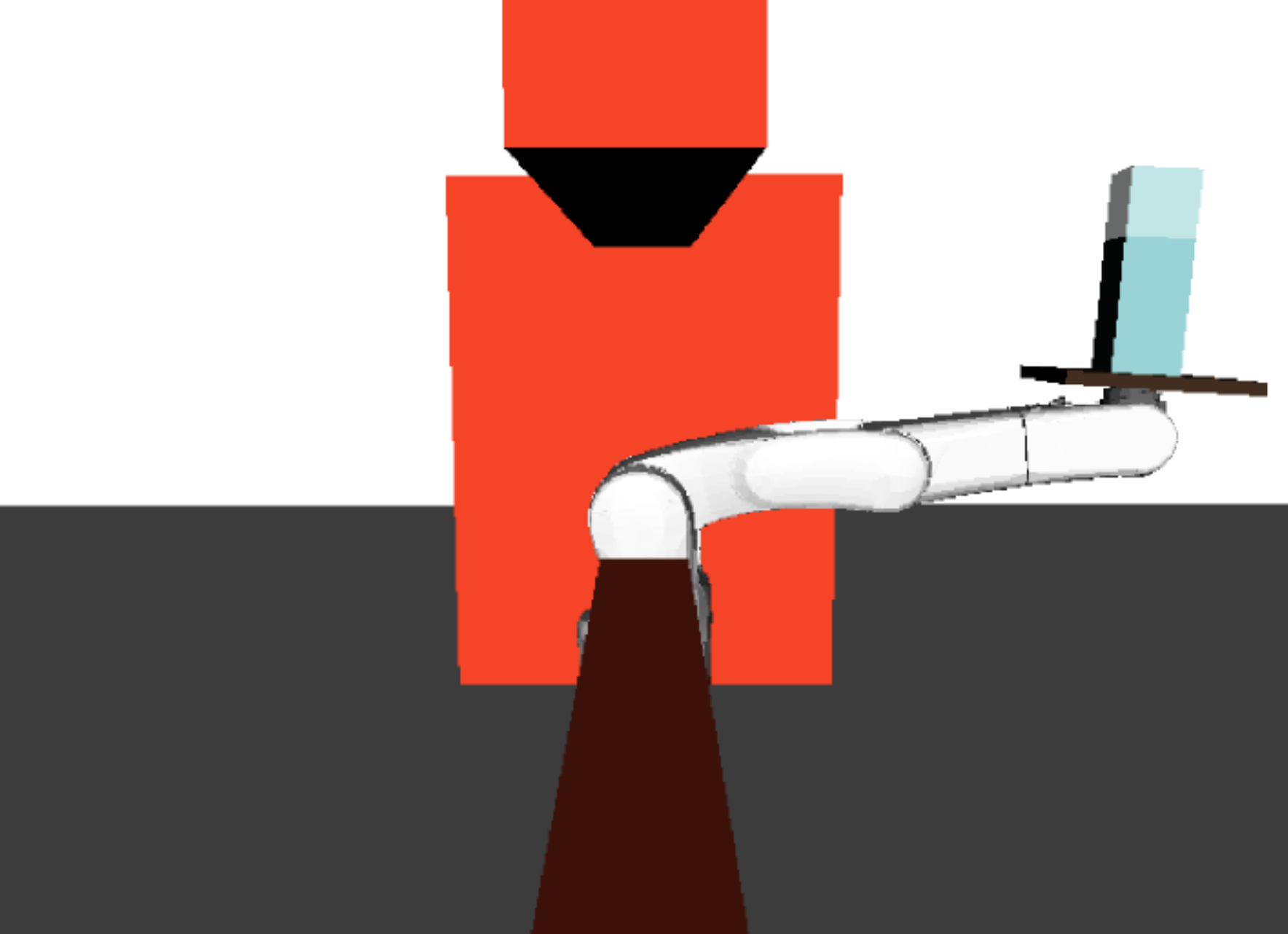}
  \includegraphics[width=3.5cm]{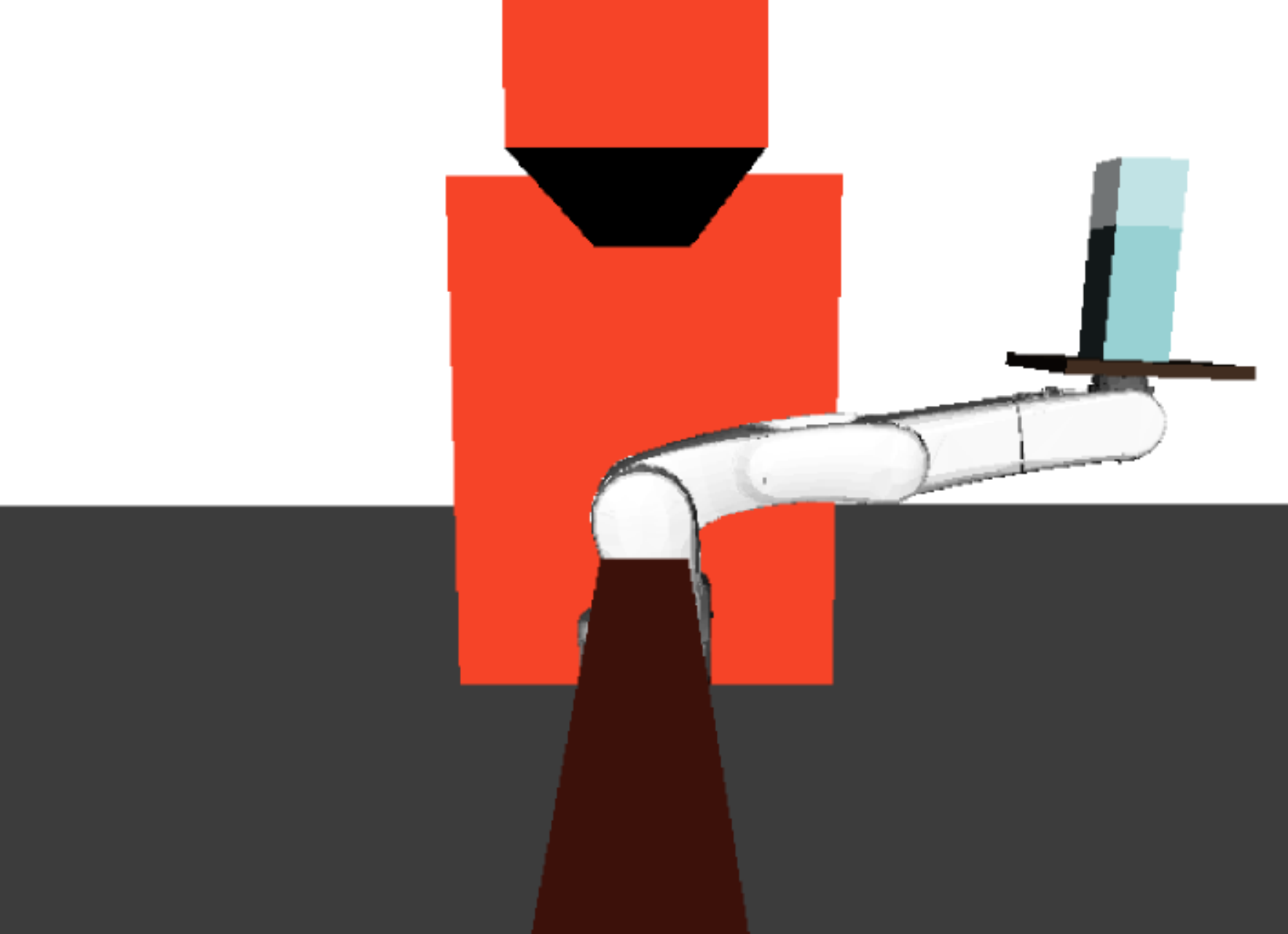}\\
  \vspace{0.2cm}
  \includegraphics[width=3.5cm]{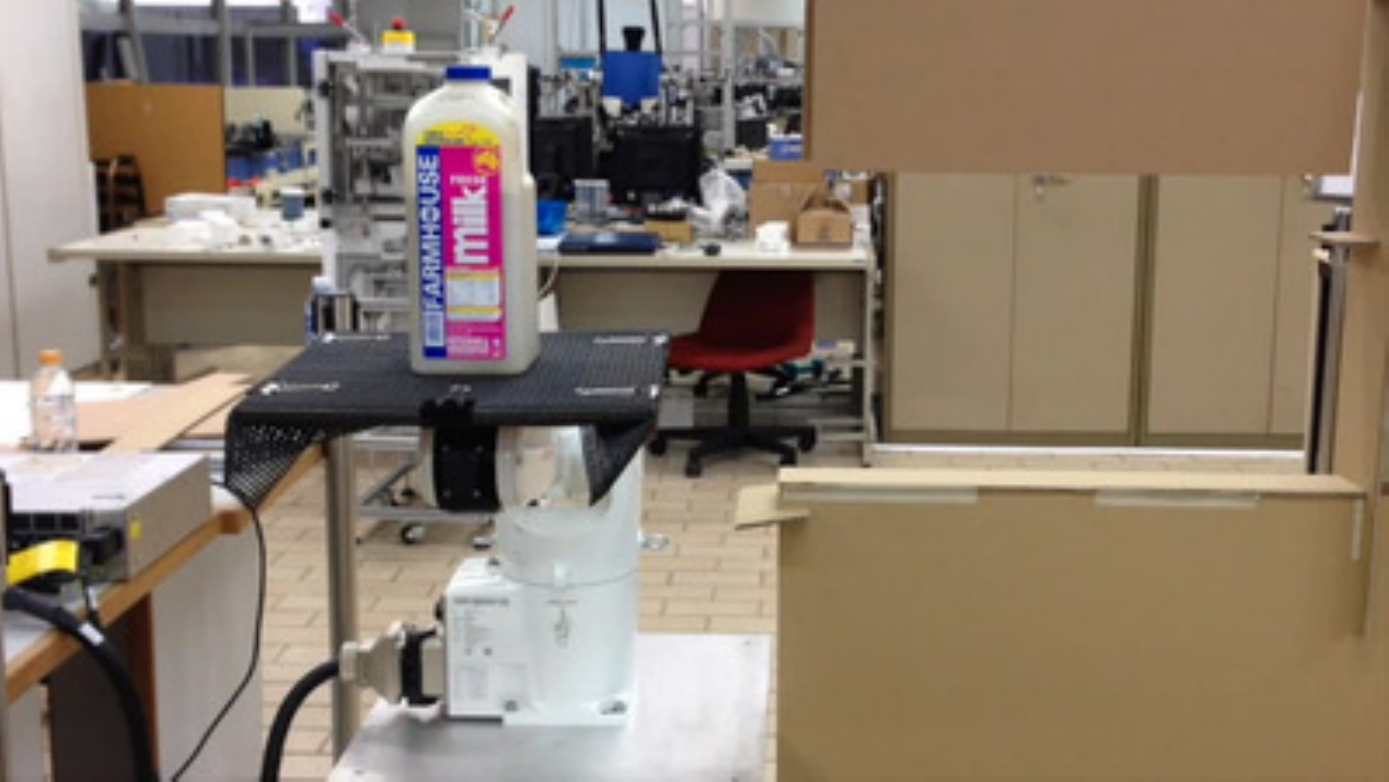}  
  \includegraphics[width=3.5cm]{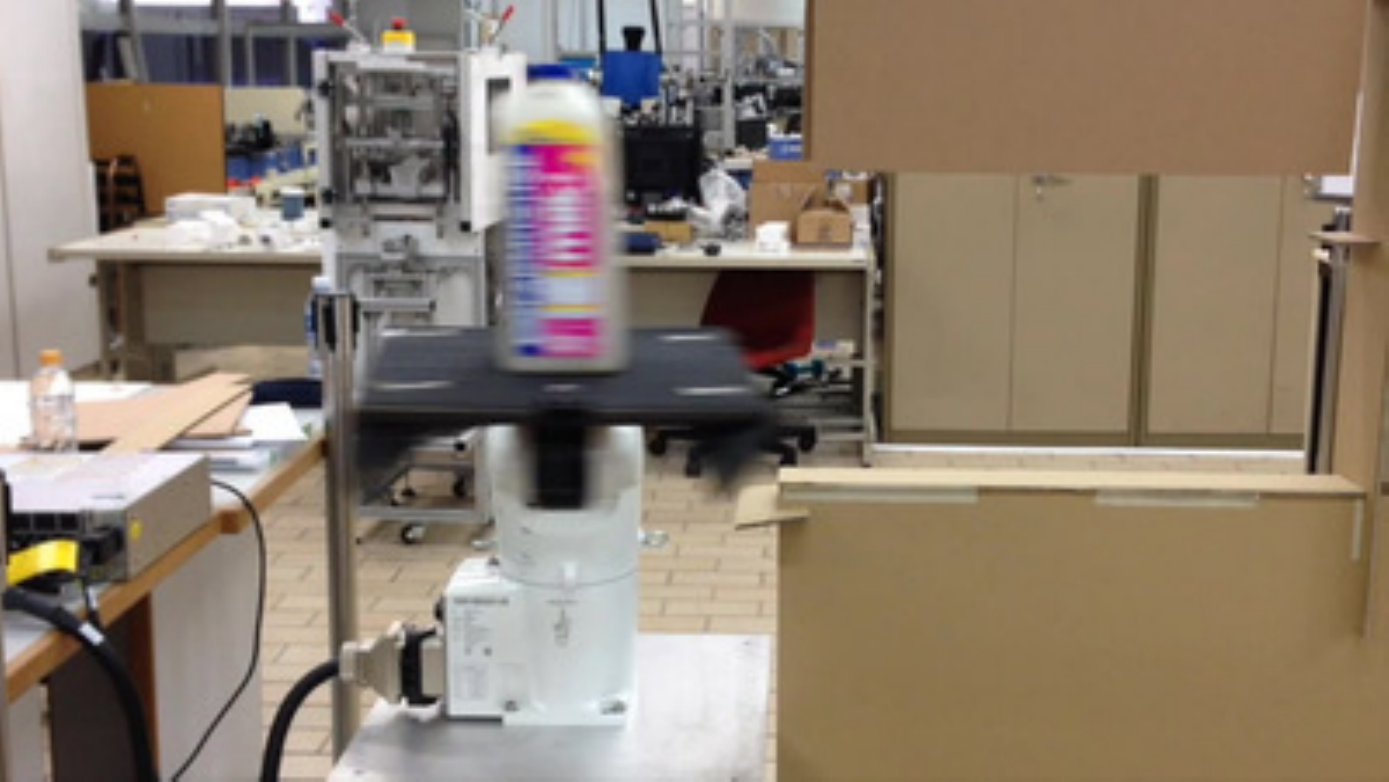}
  \includegraphics[width=3.5cm]{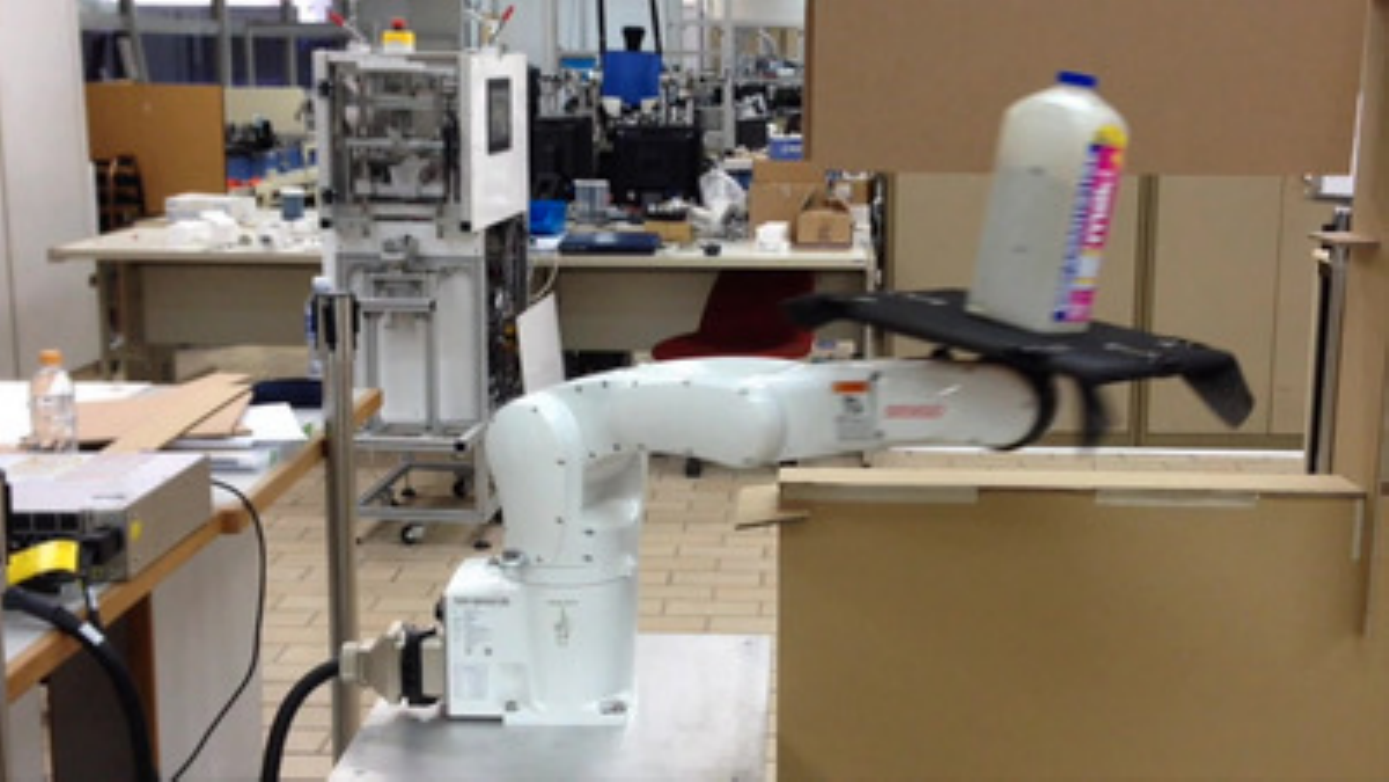}
  \includegraphics[width=3.5cm]{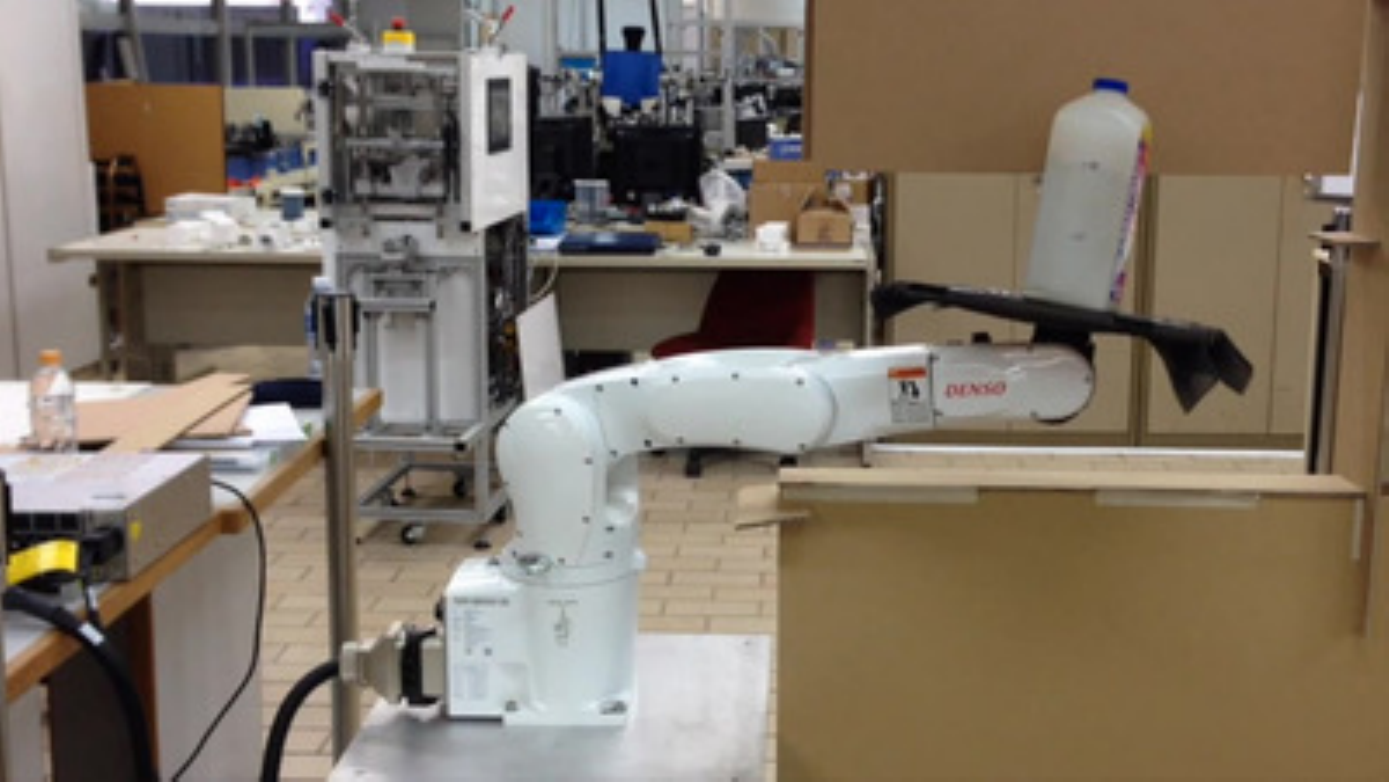}\\
  \vspace{0.2cm} 
  \includegraphics[width=3.5cm]{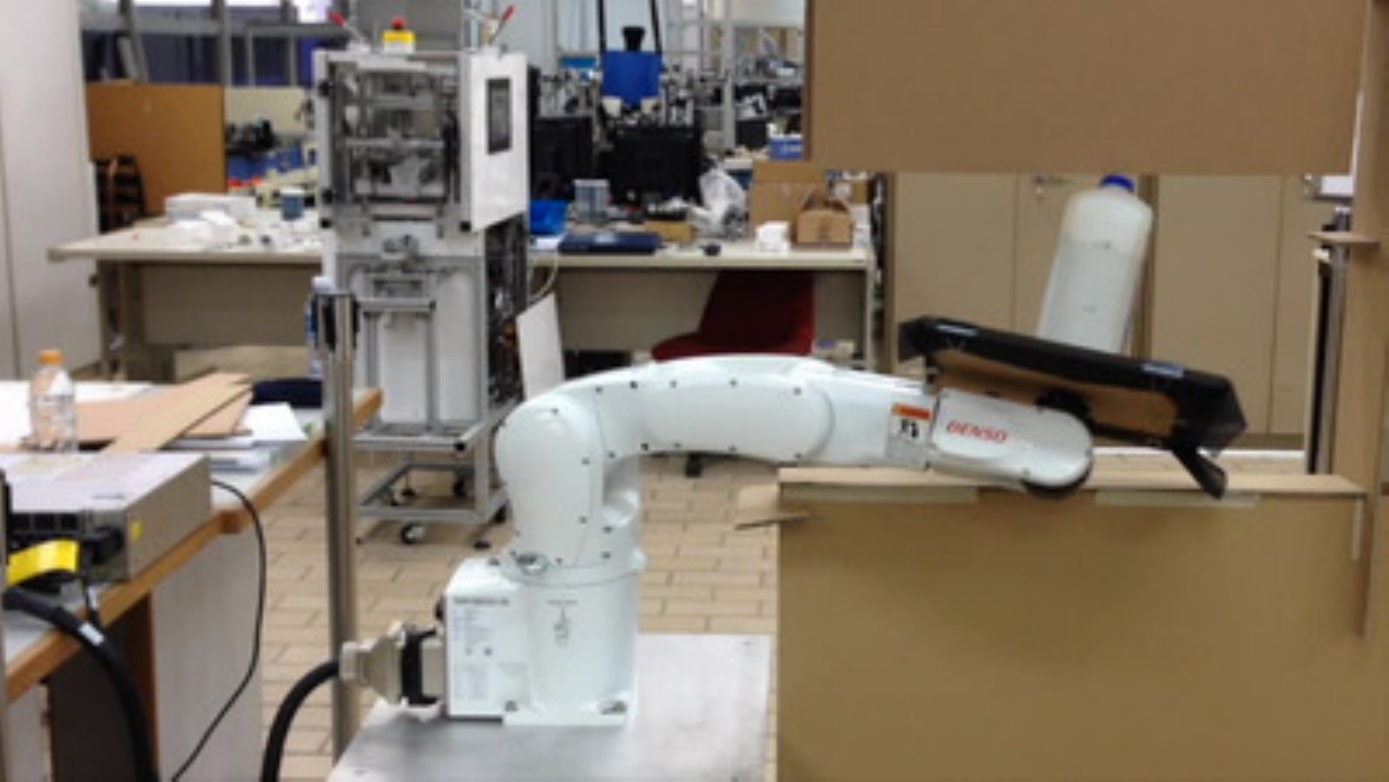}
  \includegraphics[width=3.5cm]{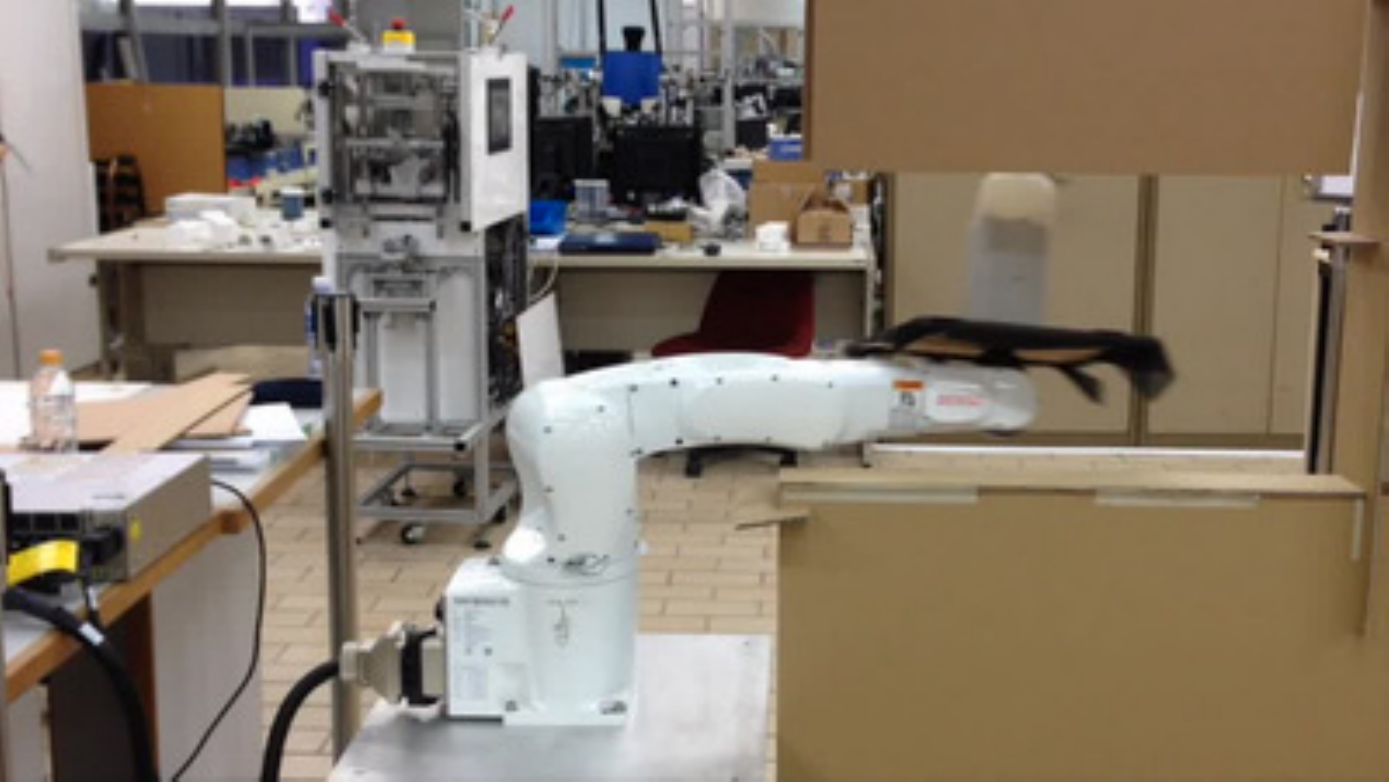}
  \includegraphics[width=3.5cm]{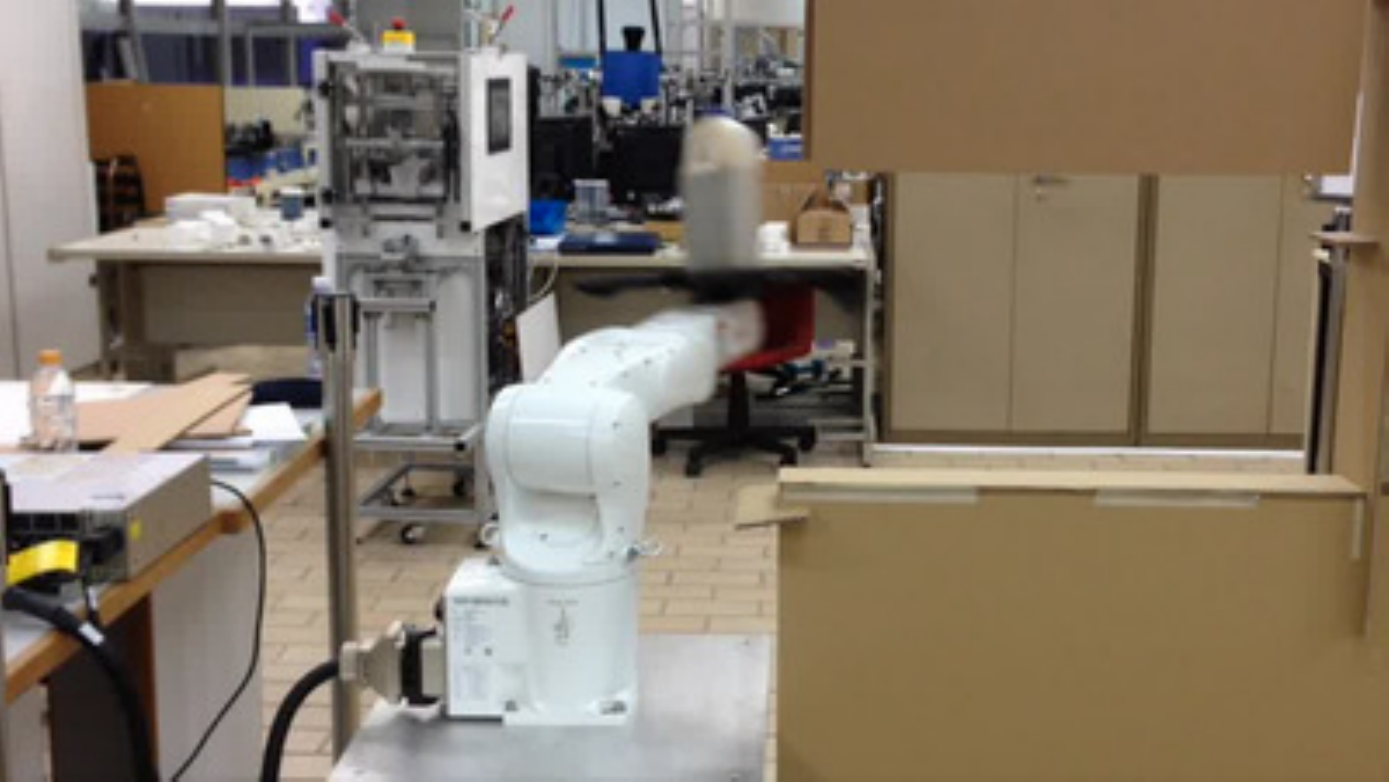}
  \includegraphics[width=3.5cm]{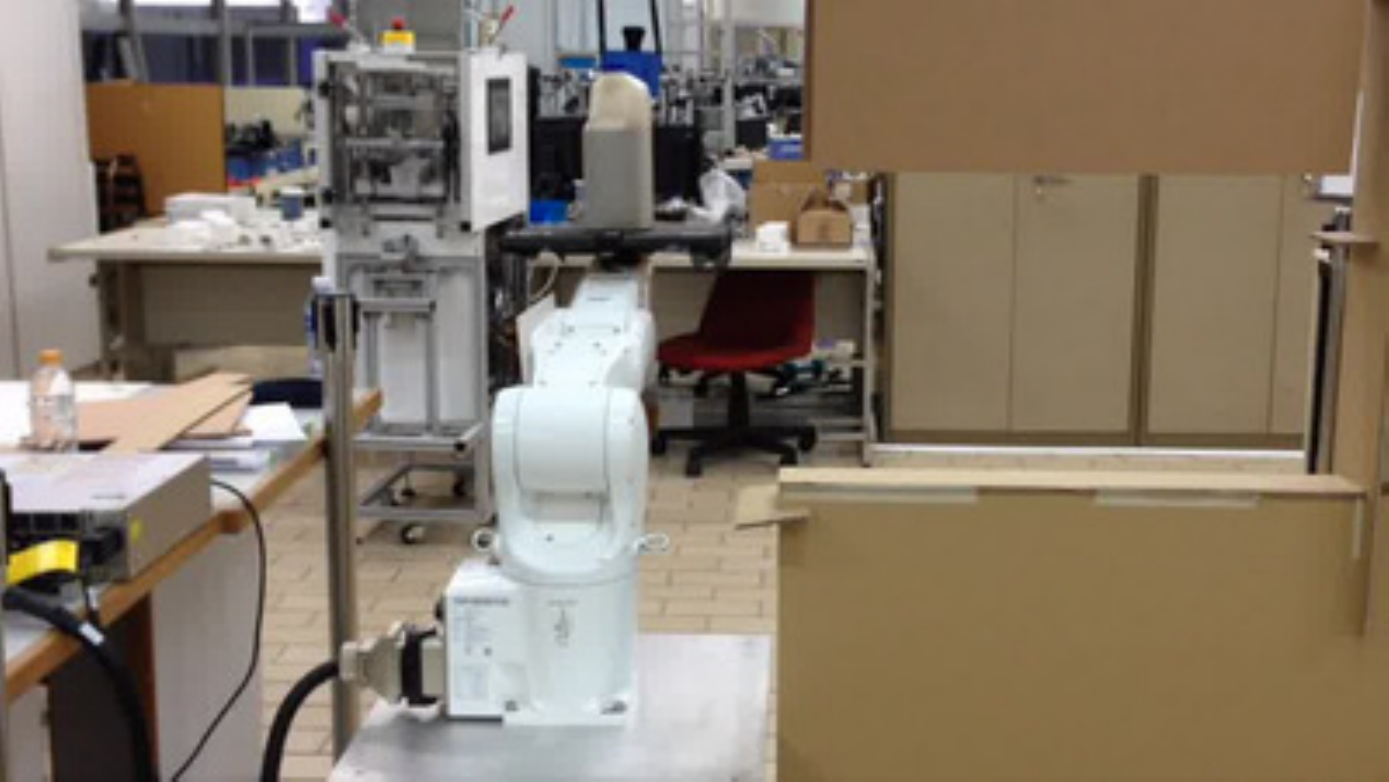}
  \caption{Snapshots of the motion of Fig.~\ref{fig:bottle}B--E taken
    every 0.5\,s. \textbf{Top two rows}\,: front view of the motion in
    the simulation environment. \textbf{Bottom two rows}\,: side view
    of the motion executed on the actual robot (see also the video
    at~\url{http://youtu.be/LdZSjNwpJs0}). }
  \label{fig:bottle2}
\end{figure}

Because of the discrepancy between the planned motion and the motion
actually executed on the robot (in particular, actual acceleration
switches cannot be infinitely fast), we set the safety boundaries to
be a square of size 5.5\,cm $\times$ 5.5\,cm (the actual base size was
8\,cm $\times$ 8\,cm), which makes the planning problem even
harder. Nevertheless, our algorithm was able to find a feasible
movement in about 3 hours on a 3.2\,GHz Intel Core computer with
3.8\,GB RAM (see Fig.~\ref{fig:bottle}B--E), and this movement could
be successfully executed on the actual robot, see
Fig.~\ref{fig:bottle2} and the video
at~\url{http://youtu.be/LdZSjNwpJs0}. Note that the computation time
of 3 hours was for a particularly difficult problem instance\,: if the
opening was only 5\,cm higher, computation time would be about 2
minutes, see Fig.~\ref{fig:bottle}F.

\subsubsection{Comparison with OMPL-KPIECE}

We were interested in comparing AVP-RRT with a state-of-the-art
planner on this bottle-and-tray problem. We chose KPIECE since it is
one of the most generic and powerful existing kinodynamic
planners~\citep{SK12tro}. Moreover, a robust open-source implementation
exists as a component of the widely-used Open Motion Planning Library
(OMPL)~\citep{SMK12ram}.

The methods and results of the comparison are reported in detail in
Appendix~\ref{sec:kpiece}. Briefly, we first fine-tuned OMPL-KPIECE on
the same 6-DOF manipulator model as above. At this stage, we
considered only bounds on velocity and accelerations, the bottle and
the tray were ignored for simplicity. Next, we compared AVP-RRT
(Python/C++) and OMPL-KPIECE (pure C++, with the best possible tunings
obtained previously) in an environment similar to that of
Fig.~\ref{fig:bottle}. Here, we considered bounds on velocity and
accelerations and collisions with the environment. We ran each planner
20 times with a time limit of 600 seconds. AVP-RRT had a success rate
of $100\%$ and an average running time of $68.67$\,s, while
OMPL-KPIECE failed to find any solution in any run. Based on this
decisive result, we decided not to try OMPL-KPIECE on the full
bottle-and-tray problem.

These comparison results thus further suggest that planning directly
in the state-space, while interesting from a theoretical viewpoint and
successful in simulations and/or on custom-built systems, is unlikely
to scale to practical high-DOF problems.

\section{Discussion}
\label{sec:discussion}

We have presented a new algorithm, Admissible Velocity Propagation
(AVP) which, given a path and an interval of reachable velocities at
the beginning of that path, computes exactly and efficiently the
interval of valid final velocities. We have shown how to combine AVP
with well-known sampling-based geometric planners to give rise to a
family of new efficient kinodynamic planners, which we have evaluated
on two difficult kinodynamic problems.

%%  In particular, we believe the
%% bottle transportation presented in Section~\ref{sec:bottle} is the
%% \emph{first successful demonstration of kinodynamic planning} in a
%% complex setting (DOF $\geq$ 6), \emph{on a commercial robot}, in an
%% environment where quasi-static motions are guaranteed to fail.

\paragraph{Comparison to existing approaches to kinodynamic planning}

Compared to traditional planners based on path-velocity decomposition,
our planners remove the limitation of quasi-static feasibility,
precisely by propagating admissible velocity intervals at each step of
the tree extension. This enables our planner to find solutions when
quasi-static trajectories are guaranteed to fail, as illustrated by
the two examples of Section~\ref{sec:applications}.

Compared to other approaches to kinodynamic planning, our approach
enjoys the advantages associated with path-velocity decomposition,
namely, the separation of the complex planning problem into two
simpler sub-problems\,: geometric and dynamic, for both of which
powerful methods and heuristics have been developed.

The bottle transportation example in Section~\ref{sec:bottle}
illustrates clearly this advantage. To address the problem of the
narrow passage constituted by the small opening, we made use of the
bridge test heuristics -- initially developed for geometric path
planners~\citep{HsuX03icra} -- which provides a large number of
samples inside the narrow passage. It is unclear how such a method
could be integrated into the ``trajectory optimization'' approach for
example. Next, to steer between two configurations, we simply
interpolated a geometric path -- and can check for collision at this
stage -- and then found possible \emph{trajectories} by running
AVP. By contrast, in a ``state-space planning'' approach, it would be
difficult -- if not impossible -- to steer exactly between two
\emph{states} of the system, which requires for instance solving a
two-point boundary value problem. To avoid solving such difficult
problems, \citet{LK01ijrr,HsuX02ijrr} propose to sample a large number
of time-series of random control inputs and to choose the time-series
that steers the system the closest to the target state. However, such
shooting methods are usually considerably slower than ``exact''
methods -- which is the case of AVP --, as also illustrated in our
simulation study (see Section~\ref{sec:pendulum} and
Appendices~\ref{sec:knnrrt},~\ref{sec:kpiece}).

% The proposed approach also appears to be scalable\,: we have
% indeed successfully tackled a difficult kinodynamic planning problem
% on a commercially-available 6-DOF manipulator, while state-space
% planners such as LQR-RRT~\citep{ShkX09icra}, LQR-Tree~\citep{Ted09rss}
% or LQR-RRT$^*$~\citep{PerX12icra}, KPIECE~\citep{SK12tro},
% SST~\citep{LiX15wafr}, etc. have only been demonstrated so far in
% simulations and/or on custom-built systems.

\paragraph{Class of systems where AVP is applicable}

Since AVP is adapted from TOPP, AVP can handle all systems and
constraints that TOPP can handle, and only those systems and
constraints. Essentially, TOPP can be applied to a path $\bfq(s)$ in
the configuration space if the system can track that path at any
velocities $\dot s$ and accelerations $\ddot s$, subject only to
\emph{inequality constraints} on $\dot s$ and $\ddot s$. This excludes
-- \emph{a priori} -- all under-actuated robots since, for these
robots, most of the paths in the configuration space cannot be
traversed at all~\citep{Lau98book}, or at only \emph{one} specific
velocity. \citet{BL01tac} identified a subclass of under-actuated
robots (including e.g. planar 3-DOF manipulators with one passive
joint or 3D underwater vehicles with three off-center thrusters) for
which one can compute a large subset of paths that can be TOPP-ed
(termed ``kinematic motions''). Investigating whether AVP-RRT can be
applied to such systems is the subject of ongoing research.

At the other end of the spectrum, redundantly-actuated robots can
track most of the paths in their configuration space (again, subject
to actuation bounds). The problem here is that, for a given admissible
velocity profile along a path, there exists in general an infinity of
combinations of torques that can achieve that velocity
profile. \citet{PS15tmech} showed how to optimally exploit actuation
redundancy in TOPP, which can be adapted straightforwardly to
AVP-RRT.

\paragraph{Further remarks on completeness and complexity}

The AVP-RRT planner as presented in Section~\ref{sec:planning} is
likely \emph{not probabilistically complete}. We address in more
detail in Appendix~\ref{sec:completeness} the completeness properties
of AVP-RRT, and more generally, of AVP-based planners.

We now discuss another feature of AVP-based planners that makes them
interesting from a complexity viewpoint. Consider a trajectory or a
trajectory segment that is ``explored'' by a state-space planning or a
trajectory optimization method -- either in one extension step for the
former, or in an iterative optimization step for the latter. If one
considers the \emph{underlying path} of this trajectory, one may argue
that these methods are exploring only \emph{one} time-parameterization
of that path, namely, that corresponding to the trajectory at hand. By
contrast, for a given path that is ``explored'' by AVP, AVP precisely
explores \emph{all} time-parameterizations of that path, or in other
words, the whole \emph{``fiber bundle''} of path velocities above the
path at hand -- at a computation cost only slightly higher than that
of checking \emph{one} time-parameterization (see
Section~\ref{sec:avpremarks}). Granted that path velocity encodes
important information about possible violations of the dynamics
constraints as argued in the Introduction, this full and free (as in
free beer) exploration enables significant performance gains.

\paragraph{Future works}

As just mentioned, we have recently extended TOPP to
redundantly-actuated systems, including humanoid
robots in multi-contact tasks~\citep{PS15tmech}. This enables
AVP-based planners to be applied to multi-contact planning for
humanoid robots. In this application, the existence of kinematic
closure constraints (the parts of the robot in contact with the
environment should remain fixed) makes path-velocity decomposition
highly appealing since these constraints can be handled by a kinematic
planner independently from dynamic constraints (torque limits,
balance, etc.)  In a preliminary experiment, we have planned a
non-quasi-statically-feasible but dynamically-feasible motion for a
humanoid robot (see \url{http://youtu.be/PkDSHodmvxY}).
Going further, we are currently investigating how AVP-based planners
can enable existing quasi-static multi-contact planning
methods~\citep{HauX08ijrr,EscX13ras} to discover truly dynamic motions
for humanoid robots with multiple contact changes.

\subsection*{Acknowledgments} We are grateful to Prof. Zvi Shiller for
inspiring discussions about the TOPP algorithm and kinodynamic
planning. This work was supported by a JSPS postdoctoral fellowship,
by a Start-Up Grant from NTU, Singapore, and by a Tier 1 grant from
MOE, Singapore.

\appendix

\section{Probabilistic completeness of AVP-based planners}
\label{sec:completeness}

Essentially, the probabilistic completeness of AVP-based planners
relies on two properties: the completeness of the path sampling
process (Property~1), and the completeness of velocity propagation
(Property~2)

\begin{description}
\item[Property 1] any smooth path $\cP$ in the configuration space
  will be approximated arbitrarily closely by the sampling process for
  a sufficiently large number of samples;
\item[Property 2] if a smooth path
  $\widehat\cP=\hat\bfq(s)_{s\in[0,1]}$ obtained by the sampling
  process can be time-parameterized into a solution trajectory
  according to a certain velocity profile $\hat v$, then $\hat v$ is
  contained within the velocity band propagated by AVP.
\end{description}

We first discuss the conditions under which these two Properties are
verified and then establish the completeness of AVP-based planners.

\begin{Def}
  Let $d$ designate a $\cL^\infty$-type distance
  between two trajectories or between two paths\,:
  \begin{eqnarray}
    \label{eq:Pi}
    d(\Pi,\widehat\Pi)&\defeq&\sup_{t\in[0,T]}\|\Pi(t)-\widehat\Pi(t)\|,\\
    \label{eq:cP}
    d(\cP,\widehat\cP)&\defeq&\sup_{s\in[0,1]}\|\bfq(s)-\widehat\bfq(s)\|+
    \sup_{s\in[0,1]}\|\bfu(s)-\widehat\bfu(s)\|,
  \end{eqnarray}
  where $\bfu(s)$ is the unit tangent vector to $\cP$ at $s$.
\end{Def}

\begin{Prop}
  Property 1 is true under the following hypotheses on the sampling
  process 
  
  \begin{description}
  \item[H1] each time a random configuration $\bfq_\rand$ is sampled,
    consider the set $\cS$ of existing vertices within a distance
    $\delta>0$ of $\bfq_\rand$ in the configuration space. Select $K$
    random vertices within $\cS$, where $K$ is proportional to the
    number of vertices currently existing in the tree, and attempt to
    connect these vertices to $\bfq_\rand$ through the usual
    interpolation and AVP procedures. For each successful connection,
    create a new vertex $V_\new$, which has the same configuration
    $\bfq_\rand$ but a different ``inpath'' and a different
    ``interval'', depending on the parent vertex in
    $\cS$\,\footnote{Note that enforcing this hypothesis on the
      AVP-RRT planner presented in Section~\ref{sec:avprrt} will turn
      it into an ``AVP-PRM''.};
  \item[H2] consider the path interpolation from $(\bfu_1,\bfq_1)$ to
    $\bfq_2$. The unit vector $\bfu_2$ at the end of the interpolated
    path $\cP_\mathrm{int}$ is set to be the unit vector pointing from
    $\bfq_1$ to $\bfq_2$, denoted
    $\bfu_{\bfq_1\to\bfq_2}$\,\footnote{Note that, if
      $\bfq_1=\bfq_\start$, then there is no associated unit tangent
      vector at $\bfq_1$. In such case, sample a random unit tangent
      vector $\bfu_\start$ for each interpolation call.};
  \item[H3] for every $\Delta>0$, there exists $\eta>0$ such that, if
    $\|\bfu_1-\bfu_{\bfq_1\to\bfq_2})\|<\eta$, then
    $d(\cP_\mathrm{int},\cP_\mathrm{str}(\bfq_1,\bfq_2))<\Delta/3$,
    where $\cP_\mathrm{str}(\bfq_1,\bfq_2)$ is the straight segment
    joining $\bfq_1$ to $\bfq_2$\,\footnote{This hypothesis basically
      says that, if the initial tangent vector ($\bfu_1$) is
      ``aligned'' with the displacement vector
      ($\bfu_{\bfq_1\to\bfq_2}$), then the interpolation path is close
      to a straight line, which is verified for any ``reasonable''
      interpolation method.}.
  \end{description}
\end{Prop}

\begin{figure}[htp]
  \centering
  \hspace{2cm}\textbf{A}\hspace{7cm}\textbf{B}\\
  \vspace{0.4cm}
  \includegraphics[height=3.5cm]{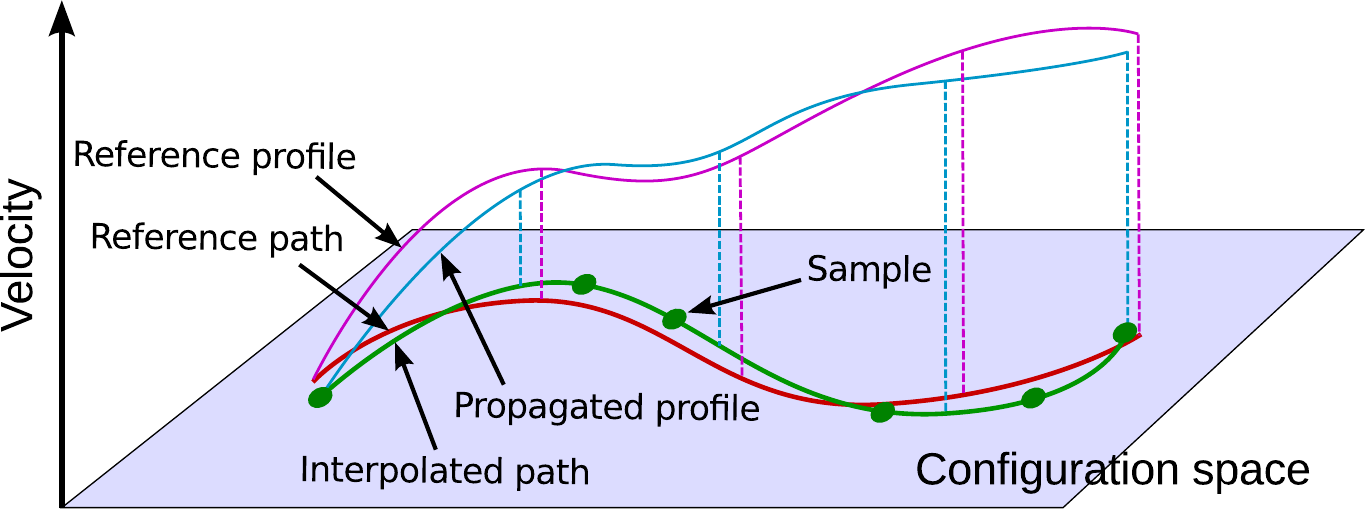}
  \hspace{0.2cm}
  \includegraphics[height=3.5cm]{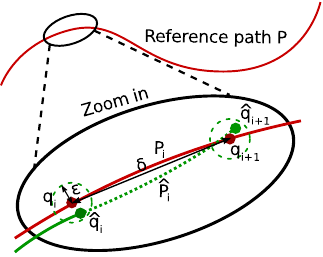}
  \caption{Completeness of AVP-RRT. \textbf{A}\,: Existence of an
    admissible velocity profile above an approximated
    path. \textbf{B}\,: Approximation of a given smooth path.}
  \label{fig:complete}
\end{figure}

\noindent\textbf{Proof}   Consider a smooth path $\cP=\bfq(s)_{s\in[0,1]}$ in the
configuration space such that $\bfq(0)=\bfq_\start$. Since $\cP$ is
smooth, for $s_1$ and $s_2$ close enough, the path segment between
$s_1$ and $s_2$ will look like a straight line, see
Fig.~\ref{fig:complete}B. This intuition can be more formally stated
as follows: consider an arbitrary $\Delta>0$,
\begin{itemize}
\item there exists a $\delta_1$ such that, if
  $\|\bfq(s_2)-\bfq(s_1)\|\leq \delta_1$, then
  \begin{equation}
    \label{eq:fact1}
    d(\bfq(s)_{s\in[s_1,s_2]},\cP_\mathrm{str}(\bfq(s_1),\bfq(s_2)))<\Delta/3;    
  \end{equation}
\item there exists $\delta_2$ such that,
  if $\|\bfq(s_2)-\bfq(s_1)\|\leq \delta_2$, then
  \begin{equation}
    \label{eq:fact2}   
    \|\bfu(s_1)-\bfu_{\bfq(s_1)\to\bfq(s_2)})\|<\eta/6
    \quad\textrm{and}\quad
    \|\bfu(s_2)-\bfu_{\bfq(s_1)\to\bfq(s_2)})\|<\eta/6,
  \end{equation}
  where $\eta$ is defined in (H3).
\end{itemize}

Divide now the path $\cP$ into $n$ subpaths $\cP_1$,\dots,$\cP_n$ of
lengths approximately $\delta\defeq\min(\delta_1,\delta_2)$. Let
$\bfq_i,\bfu_i$ denote the starting configuration and unit tangent
vector of subpath $\cP_i$. Consider the balls $B_i$ centered on the
$\bfq_i$ and having radius $\epsilon$, where
$\epsilon\defeq\frac{\eta\delta}{12}$. With probability 1, there will
exist a time in the sampling process when
\begin{description}
\item[(s1)] $n$ \emph{consecutive} random configurations
  $\hat\bfq_1,\dots,\hat\bfq_1$ are sampled in $B_1,\dots,B_n$
  respectively;
\item[(s2)] $\bfq_\start$ is selected for connection attempt towards
  $\hat\bfq_1$, and the random $\bfu_\start$ verifies
  $\|\bfu_\start-\bfu_{\bfq_\start\to\bfq_1}\|<2\eta/3$.  The
  interpolation results in a new vertex $V_1$ and a new subpath
  $\widehat\cP_1$ connecting $\bfq_\start$ to $V_1$;
\item[(s3)] for $i\in[1,n-1]$, $V_i$ is selected for connection attempt to
  $\hat\bfq_{i+1}$, resulting in a new vertex $V_{i+1}$ and a new  subpath
  $\widehat\cP_i$ connecting $V_i$ to $V_{i+1}$.
\end{description}
Note that (s2) and (s3) are possible since, by (H1), the number of
connection attempts $K$ grows linearly with the number of existing
vertices in the tree.

We first prove that, for all $i\in[0,n]$, we have
$\|\hat\bfu_i-\bfu_i\|<2\eta/3$. At rank $0$, the property is true
owing to (s2). For $i\geq 1$, we have
\begin{itemize}
\item $\|\hat\bfu_i-\bfu_{\hat\bfq_{i-1}\to\hat\bfq_i}\|=0$ by (H2);
\item
  $\|\bfu_{\hat\bfq_{i-1}\to\hat\bfq_i}-\bfu_{\bfq_{i-1}\to\bfq_i}\|<
  2\epsilon/\delta=\eta/6$ by the fact that each $\bfq_i$ is contained
  in the ball $B_i$;
\item $\|\bfu_{\bfq_{i-1}\to\bfq_i}-\bfu_i\|<\eta/6$
  by~(\ref{eq:fact2}).
\end{itemize}
Applying triangle inequality yields $\|\hat\bfu_i-\bfu_i\|<2\eta/3$.

Next, we prove for all $i\in[0,n-1]$ that
$d(\widehat{\cP}_i,\cP_i)<\Delta$. Note that
\begin{itemize}
\item $\|\hat\bfu_i-\bfu_i\|<2\eta/3$ by the above reasoning;
\item $\|\bfu_i-\bfu_{\bfq_{i}\to\bfq_{i+1}}\|<\eta/6$ by~(\ref{eq:fact2});
\item
  $\|\bfu_{\bfq_{i}\to\bfq_{i+1}}-\bfu_{\hat\bfq_{i}\to\hat\bfq_{i+1}}\|<
  2\epsilon/\delta=\eta/6$ by the fact that each $\bfq_i$ is contained
  in the ball $B_i$.
\end{itemize}
Thus, by triangle inequality, we have
$\|\hat\bfu_i-\bfu_{\hat\bfq_{i}\to\hat\bfq_{i+1}}\|<\eta$. By (H3),
we have
$d(\widehat\cP_i,\cP_\mathrm{str}(\hat\bfq_1,\hat\bfq_2))<\Delta/3$.
Next,
$d(\cP_\mathrm{str}(\hat\bfq_1,\hat\bfq_2),\cP_\mathrm{str}(\bfq_1,\bfq_2))$
can be made smaller than $\Delta/3$ for judicious choices of
$\epsilon$ and $\delta$. Finally, we have
$d(\cP_i,\cP_\mathrm{str}(\bfq(s_1),\bfq(s_2)))<\Delta/3$ by
(\ref{eq:fact1}). Applying the triangle inequality again, we obtain
$d(\widehat{\cP}_i,\cP_i)<\Delta$. $\Box$

\begin{Prop}
  Property 2 is true.
\end{Prop}

\noindent\textbf{Proof} Consider a path $\widehat\cP$ obtained by the
sampling process, i.e.  $\widehat\cP$ is composed of $n$ interpolated
path segments $\widehat\cP_1$,\dots,$\widehat\cP_n$. Let
$v_1$,\dots,$v_n$ be the corresponding subdivisions of the associated
velocity profile $v$. We prove by induction on $i\in[0,n]$ that the
concatenated profile $[v_1,\dots,v_i]$ is contained within the
velocity band propagated by AVP.

For $i=0$, i.e., at the start vertex, $v(0)=0$ is contained within the
initial velocity band, which is $[0,0]$. Assume that the statement
holds at $i$. This implies in particular that the final value of
$v_i$, which is also the initial value of $v_{i+1}$, belongs to
$[v_{\min},v_{\max}]$, where $(v_{\min},v_{\max})$ are the values
returned by AVP at step $i$. Next, consider the velocity band that AVP
propagates at step $i+1$ from $[v_{\min},v_{\max}]$. Since
$v_{i+1}(0)\in[v_{\min},v_{\max}]$ and that $v_{i+1}$ is continuous,
the whole profile $v_{i+1}$ will be contained, by construction, in the
velocity band propagated by AVP. $\Box$

We can now prove the probabilistic completeness for a class of
AVP-based planners.

\begin{Theo}
  \label{theo:complete}
  An AVP-based planner that verifies Properties~1 and~2 is
  probabilistically complete.
\end{Theo}

\noindent\textbf{Proof} Assume
that there exists a smooth state-space trajectory $\Pi$ that solves
the query, with $\Delta$-clearance in the state space, i.e., every
smooth trajectory $\widehat\Pi$ such that $d(\Pi,\widehat\Pi)\leq
\Delta$ also solves the query\,\footnote{Note that this property
  presupposes that the robot is fully-actuated, see also the paragraph
  ``Class of systems where AVP is applicable'' in
  Section~\ref{sec:discussion}.}. Let $\cP$ be the underlying path of
$\Pi$ in the configuration space. By Property~1, with
% actually it is useless to say "with probability 1"
probability~1, there exists a time when the sampling process will
generate a smooth path $\widehat\cP$ such that
$d(\cP,\widehat\cP)\leq\Delta/2$. One can then construct, by
continuity, a velocity profile $\hat v$ above $\widehat\cP$, such that
the time-parameterization of $\widehat\cP$ according to $\hat v$
yields a trajectory $\widehat\Pi$ within a radius $\Delta$ of $\Pi$
(see Fig.~\ref{fig:complete}A). As $\Pi$ has $\Delta$-clearance,
$\widehat\Pi$ also solves the query. Thus, by Property~2, the velocity
profile (or time-parameterization) $\hat v$ must be contained within
the velocity band propagated by AVP, which implies finally that
$\widehat\cP$ can be successfully time-parameterized in the last step
of the planner. $\Box$

\section{Comparison of AVP-RRT with $K$NN-RRT on a 2-DOF pendulum}
\label{sec:knnrrt}

Here, we detail the implementation of the standard state-space planner
$K$NN-RRT and the comparison of this planner with AVP-RRT. The full
source code for this comparison is available at
\url{https://github.com/stephane-caron/avp-rrt-rss-2013}.  Note that, for
fairness, all the algorithms considered here were implemented in
Python (including AVP). Thus, the presented computation times, in
particular those of AVP-RRT, should not be considered in absolute
terms.

\subsection{$K$NN-RRT}

\subsubsection{Overall algorithm}
\label{subsec:gen}

Our implementation of RRT in the state-space \citep{LK01ijrr} is
detailed in Boxes~\ref{algo:rrt-annex} and \ref{algo:extend-annex}.

\begin{algorithm}
    \caption{$K$NN\_RRT$(\bfx_\init,\bfx_\goal)$}
    \label{algo:rrt-annex}
    \begin{algorithmic}[1]
        \STATE $\cT$.INITIALIZE($\bfx_\init$)
        \FOR{$\mathrm{rep}=1$ to $N_\mathrm{max\_rep}$}
        \STATE $\bfx_\rand\leftarrow$ RANDOM\_STATE() \textbf{if} mod(rep,5) $\neq 0$ \textbf{else} $\bfx_\goal$
        \STATE $\bfx_\new \leftarrow$ EXTEND($\cT,\bfx_\rand$)
        \STATE $\cT$.ADD\_VERTEX($\bfx_\new$)
        \STATE $\bfx_{\new 2} \leftarrow$ EXTEND($\bfx_\new,\bfx_\goal$)
        \IF{$d(\bfx_\new,\bfx_\goal) \leq \epsilon$ or $d(\bfx_{\new2},\bfx_\goal) \leq \epsilon$}
        \RETURN Success
        \ENDIF
        \ENDFOR
        \RETURN Failure
    \end{algorithmic}
\end{algorithm}

\begin{algorithm}
    \caption{EXTEND($\cT,\bfx_\rand$)}
    \label{algo:extend-annex}
    \begin{algorithmic}[1]
        \FOR{$k=1$ to $K$}
        \STATE $\bfx_\near^k\leftarrow$
        KTH\_NEAREST\_NEIGHBOR($\cT,\bfx_\rand,k$)
        \STATE $\bfx_\new^k\leftarrow$ STEER($\bfx_\near^k,\bfx_\rand$)
        \ENDFOR
        \RETURN $\arg\min_k d(\bfx_\new^k,\bfx_\rand)$
    \end{algorithmic}
\end{algorithm}

\paragraph{Steer-to-goal frequency} We asserted the efficiency of the
following strategy: every five extension attempts, try to steer
directly to $\bfx_\goal$ (by setting $\bfx_\rand = \bfx_\goal$ on line
3 of Box~\ref{algo:rrt-annex}). See also the discussion in
\citet{LK01ijrr}, p.~387, about the use of uni-directional and
bi-directional RRTs. We observed that the choice of the steer-to-goal
frequency (every 5, 10, etc., extension attempts) did not
significantly alter the performance of the algorithm, except when it
is too large, \eg once every two extension attempts.

\paragraph{Metric} The metric for the neighbors search in EXTEND
(Box~\ref{algo:extend-annex}) and to assess whether the goal has
been reached (line 7 of Box~\ref{algo:rrt-annex}) was defined
as:
\begin{eqnarray}
  \label{eq:d}
  d(\bfx_a,\bfx_b) 
  &=& d\left((\bfq_a,\bfv_a),(\bfq_b,\bfv_b)\right)\nonumber\\
  &=& \frac{\sum_{j=1,2}{\sqrt{1-\cos({\bfq_a}_j-{\bfq_b}_j)}}}{4} + 
  \frac{\sum_{j=1,2}|{\bfv_a}_j-{\bfv_b}_j|} {4V_{\max}},
\end{eqnarray}
where $V_{\max}$ denotes the maximum velocity bound set in the random sampler
(function RANDOM\_STATE() in Box~\ref{algo:rrt-annex}). This simple metric is
similar to an Euclidean metric but takes into account the periodicity of the
joint values.

\paragraph{Termination condition} We defined the goal area as a ball of radius
$\epsilon = 10^{-2}$ for the metric \eqref{eq:d} around the goal state
$\bfx_\goal$. As an example, $d(\bfx_a, \bfx_a)=\epsilon$ corresponds to
a maximum angular difference of $\Delta q_1 \approx
0.057$~rad~$\approx~$3.24~degrees in the first joint.

This choice is connected to that of the integration time step (used \eg in
Forward Dynamics computations in section~\ref{sec:local}), which we set to
$\delta t = 0.01$~s. Indeed, the average angular velocities we observed in our
benchmark was around $\bar{V}~=~5$ rad.s$^{-1}$ for the first joint, which
corresponds to an average instantaneous displacement $\bar{V} \cdot \delta
t \approx 5 . 10^{-2}$~rad of the same order as $\Delta q_1$ above.

\paragraph{Nearest-neighbor heuristic} Instead of considering only
extensions from the nearest neighbor, as has commonly been done, we
considered the ``best'' extension from the $K$ nearest neighbors (line
5 in Box~\ref{algo:extend-annex}), i.e. the extension yielding the
state closest to $\bfx_\rand$ for the metric $d$ (\cf
Equation~\eqref{eq:d}).

\subsubsection{Local steering}
\label{sec:local}

Regarding the local steering scheme (STEER on line 3 of
Box~\ref{algo:extend-annex}), there are two main approaches,
corresponding to the two sides of the equation of motion\,:
state-based and control-based steering \citep{CarX14icra}.

\paragraph{Control-based steering} In this approach, a control input
$\tau(t)$ is computed first. It generates a given trajectory
computable by forward dynamics. Because $\tau(t)$ is computed
beforehand, there is no direct control on the end-state of the
trajectory. To palliate this, the function $\tau(t)$ is then updated,
with or without feedback on the end-state, until some satisfactory
result is obtained or a computation budget is exhausted. For example,
in works such as~\citet{LK01ijrr, HsuX02ijrr}, random functions $u$
are sampled from the set of piecewise-constant functions.  A number of
them are tried and only the one bringing the system closest to the
target is retained.  Linear-Quadratic
Regulation~\citep{PerX12icra,Ted09rss} is another example of
control-based steering where the function $u$ is computed as the
optimal policy for a linear approximation of the system dynamics
(given a quadratic cost function).

In the present work, we followed the control-based approach from
\citet{LK01ijrr, HsuX02ijrr}, as described by Box~\ref{algo:steer}.
The random control is a stationary $(\tau_1, \tau_2)$ sampled as:
\[
    (\tau_1, \tau_2)\ \sim\ {\cal U}([-\taumax_1, \taumax_1] \times
    [-\taumax_2, \taumax_2]).
\] 
where $\cal U$ denotes uniform sampling from a set. The random time duration
$\Delta t$ is sampled uniformly in $[\delta t, \Delta t_{\max}]$ where $\Delta
t_{\max}$ is the maximum duration of local trajectories (parameter to be
tuned), and $\delta t$ is the time step for the forward dynamics integration
(set to $\delta t = 0.01$~s as discussed in Section \ref{subsec:gen}). The
number of local trajectories to be tested, $N_\mathrm{local\_trajs}$, is also
a parameter to be tuned.

\begin{algorithm}
    \caption{STEER($\bfx_\near,\bfx_\rand$)}
    \label{algo:steer}
    \begin{algorithmic}[1]
        \FOR{$p=1$ to $N_\mathrm{local\_trajs}$} \STATE $\bfu \leftarrow$
        RANDOM\_CONTROL($\taumax_1,\taumax_2$) \STATE $\Delta t \leftarrow$
        RANDOM\_DURATION($\Delta t_{\max}$) \STATE $\bfx^p\leftarrow$
        FORWARD\_DYNAMICS($\bfx_\near,\bfu,\Delta t$)
        \ENDFOR
        \RETURN $\mathrm{argmin}_p d(\bfx^p,\bfx_\rand)$
    \end{algorithmic}
\end{algorithm}

\paragraph{State-based steering} In this approach, a trajectory $\wq(t)$ is
computed first. For instance, $\wq$ can be a Bezier curve matching the initial
and target configurations and velocities. The next step is then to compute
a control that makes the system track it. For fully- or over-actuated system,
this can done using inverse dynamics. If no suitable controls exist, the
trajectory is rejected. Note that both the space $\Im(\wq)$ and timing $t$
impact the dynamics of the system, and therefore the existence of admissible
controls. Bezier curves or B-splines will conveniently solve the spatial part
of the problem, but their timing is arbitrary, which tends to result in invalid
controls and needs to be properly cared for.

To enable meaningful comparisons with AVP-RRT, we considered the
simple state-based steering described in Box~\ref{algo:state-based}.
Trying to design the best possible nonlinear controller for the double
pendulum would be out of the scope of this work, as it would imply
either problem-specific tunings or substantial modifications to the
core RRT algorithm \citep[as done e.g. in][]{PerX12icra}.

\begin{algorithm}
    \caption{STEER($\bfx_\near,\bfx_\rand$)}
    \label{algo:state-based}
    \begin{algorithmic}[1]
        \FOR {10 trials of $T \sim {\cal U}([0.01, 2.])$}
            \STATE $\wq \leftarrow \INTERPOLATE(T, \bfx_\near, \bfx_\rand)$
            \STATE $\wtau := \textrm{INVERSE\_DYNAMICS}(\wq(t), \wqd(t), \wqdd(t))$
            \STATE $t^\dagger = \sup\{t | |\wtau(t)| \leq \taumax\}$
            \IF {$t^\dagger > 0$}
                \RETURN $\wq(t^\dagger)$
            \ENDIF
        \ENDFOR
        \RETURN failure
    \end{algorithmic}
\end{algorithm}

Here, $\INTERPOLATE(T, \bfx_a, \bfx_b)$ returns a third-order polynomial $P_i(t)$
such that $P_i(0)=\bfq_{ai},\ P'_i(0)=\bfv_{ai},\ P_i(T)=\bfq_{bi},\
P_i'(T)=\bfv_{bi}$, and our local planner tries 10 different values of $T$ between
0.01~s and 2~s. We use inverse dynamics at each time step of the trajectory to
check if a control $\wtau(t)$ is within torque limits. The trajectory is cut at the
first inadmissible control.

\begin{figure}[htp]
  \centering
  \includegraphics[width=7cm,height=4cm,clip=true,trim=0cm 8cm 0cm 8cm]{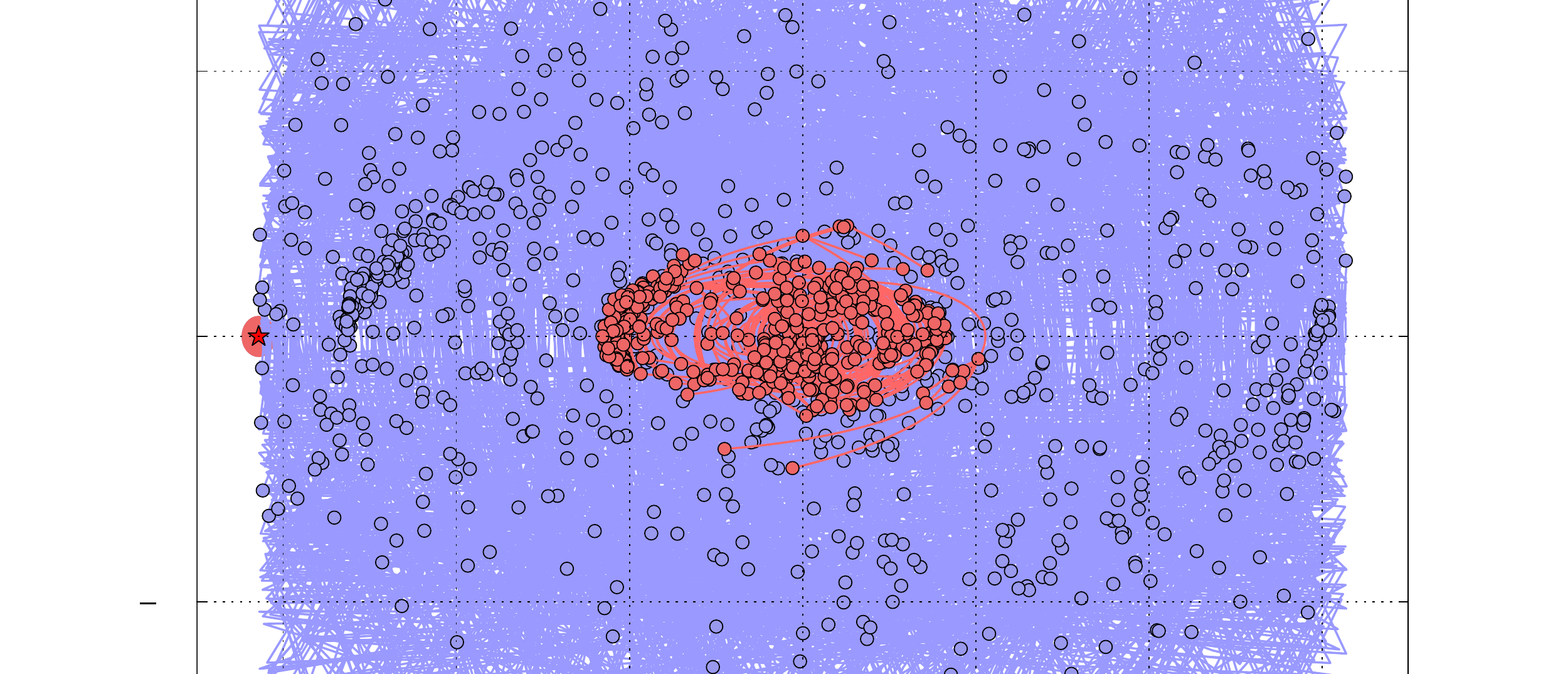}
  \hspace{.1cm}
  \includegraphics[width=7cm,height=4cm,clip=true,trim=0cm 8cm 0cm 8cm]{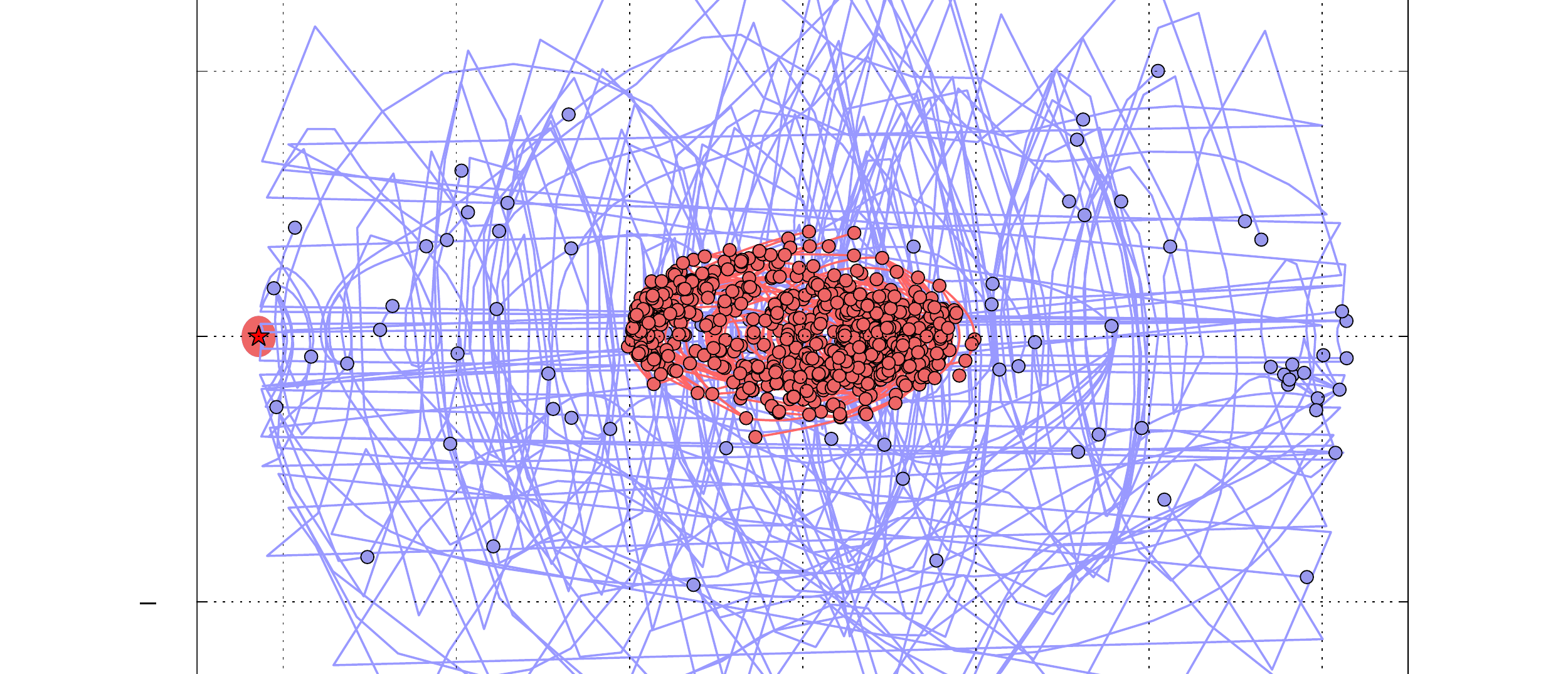}\\
  \vspace{.75cm}
  \includegraphics[width=7cm,height=4cm,clip=true,trim=0cm 8cm 0cm 8cm]{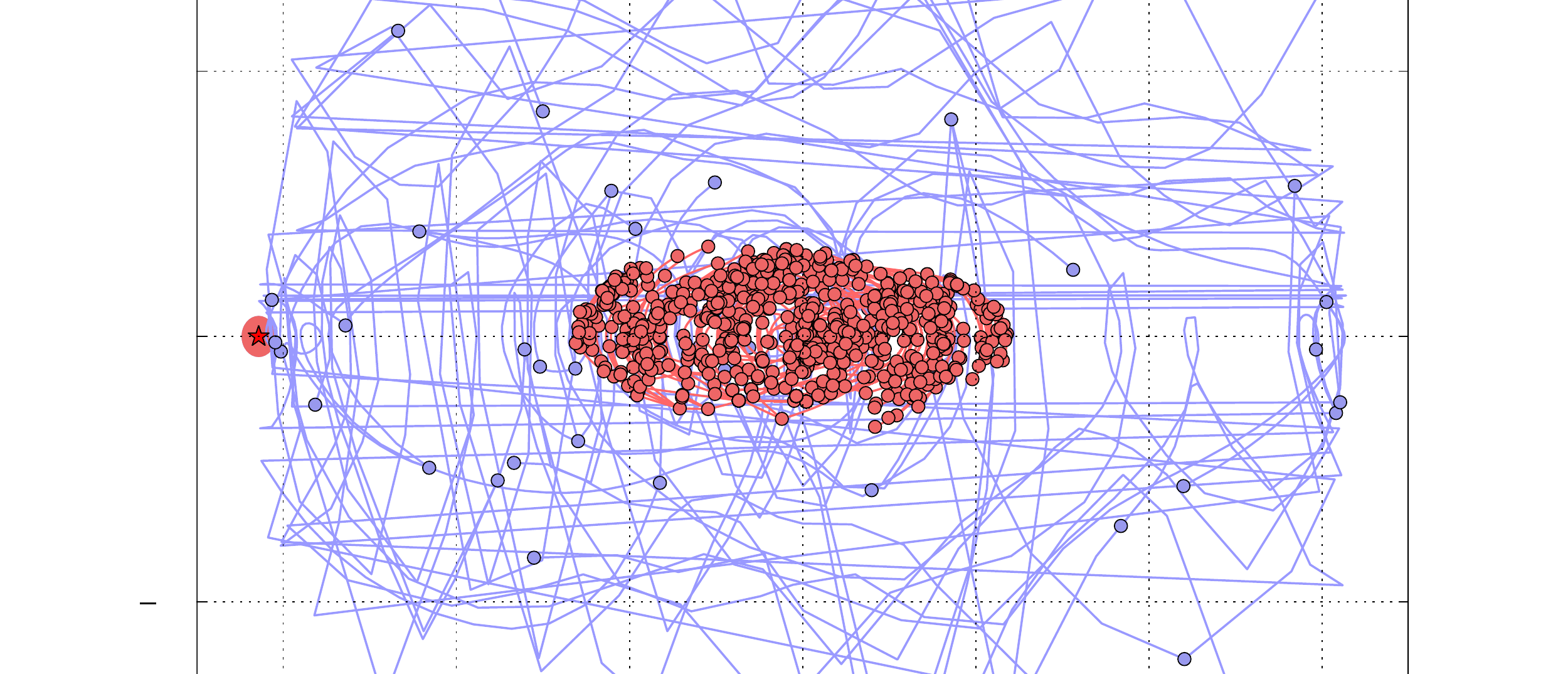}
  \hspace{.1cm}
  \includegraphics[width=7cm,height=4cm,clip=true,trim=0cm 8cm 0cm 8cm]{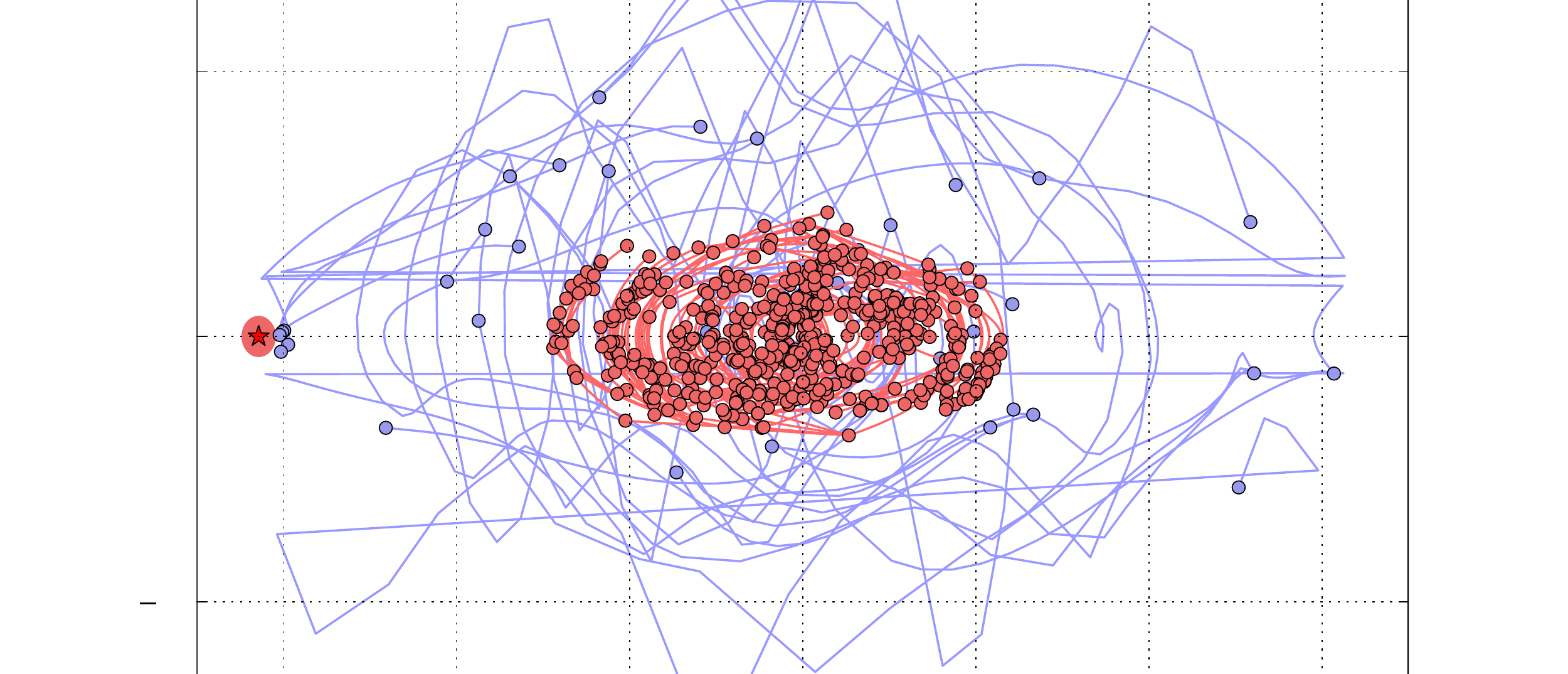}
  \caption{
      Comparison of control-based and state-based steering for $K=1$
      (left-top), $K=10$ (right-top), $K=40$ (left-bottom) and $K=100$
      (right-bottom). Computation time is fixed, which explains why there are more
      points for small values of $K$. The X-axis represents the angle of the
      first joint and the Y-axis its velocity. The trees grown by the
      state-based and control-based methods are in red and blue, respectively.
      The goal area is depicted by the red ellipse on the left side. Control-based
      steering yields better exploration of the state space, but fails to connect
      to the goal area.
  }
  \label{fig:steer}
\end{figure}

\paragraph{Comparing the two approaches} On the pendulum, state-based steering
yielded RRTs with slower exploration speeds compared to control-based
steering, as illustrated in Figure~\ref{fig:steer}. This slowness is
likely due to the uniform sampling in a wide velocity range
$[-V_{\max}, V_{\max}]$, which resulted in a large fraction of
trajectories exceeding torque limits. However, despite a better
exploration of the state space, trajectories from control-based
steering systematically ended outside of the goal area. To palliate
this, we added a subsequent step\,: from each state reached by
control-based steering, connect to the goal area using state-based
steering.  Thus, if a state is reached that is not in the goal area
but from which steering to goal is easy, this last step will take care
of the final connection. However, this patch improved only marginally
the success rate of the planner.  In practice, trajectories from
control-based steering tend to end at energetic states from which
steering to goal is difficult. As such, we found that this steering
approach was not performing well on the pendulum and turned to
state-based steering.

Let us remark here that, although AVP-RRT follows the state-based
paradigm (it indeed interpolates paths in configuration space and then
computes feasible velocities along the path using Bobrow-like
approach, which includes inverse dynamics computations), it is much
more successful. The reason for this lies in AVP\,: when the
interval of feasible velocities is small, a randomized approach will
have a high probability of sampling unreachable velocities. Therefore,
it will fail most of the time. Using AVP, the set of reachable
velocities is exactly computed and this failure factor
disappears. With AVP-RRT, failures only occur from ``unlucky''
sampling in the configuration space. Note however that the algorithm
only saves and propagates the \emph{norm} of the velocity vectors, not
their directions, which may make the algorithm probabilistically
incomplete (cf. discussion in Section~\ref{sec:discussion}).

\subsubsection{Fine-tuning of $K$NN-RRT}
\label{sec:id}

Based on the above results, we now focus on $K$NN-RRTs with state-based
steering for the remainder of this section. The parameters to be tuned are\,:
\begin{itemize}

    \item $N_\mathrm{local\_trajs}$: number of local trajectories tested in
        each call to STEER;

    \item $\Delta t_{\max}$: maximum duration of each local trajectory.

\end{itemize}
The values we tested for these two parameters are summed up in table
\ref{table:tunings}.
\begin{table}[htp]
\begin{center}
    \begin{tabular}{|c|c|c|}
    \hline
    Number of trials & $N_\mathrm{local\_trajs}$ & $\Delta t_{\max}$ \\
    \hline
    \hline
    10 & \color{OrangeRed} 1  & \color{PineGreen} 0.2 \\
    10 & \color{OrangeRed} 30 & \color{PineGreen} 0.2 \\
    10 & \color{OrangeRed} 80 & \color{PineGreen} 0.2 \\
    \hline
    20 & \color{PineGreen} 20 & \color{OrangeRed} 0.5 \\
    20 & \color{PineGreen} 20 & \color{OrangeRed} 1.0 \\
    20 & \color{PineGreen} 20 & \color{OrangeRed} 2.0 \\
    \hline
    \end{tabular}
    \caption{Parameter sets for each test.}
    \label{table:tunings}
\end{center}
\end{table}
The parameters we do not tune are\,:
\begin{itemize}

    \item Maximum velocity $V_{\max}$ for sampling velocities. We set $V_{\max}
        = 50$ rad.s$^{-1}$, which is about twice the maximum velocity observed
        in the successful trials of AVP-RRT;

    \item Number of neighbors $K$. In this tuning phase, we set $K=
      10$. Other values of $K$ will be tested in the final comparison with
      AVP in section~\ref{sec:compare};

    \item Space-time precision $(\epsilon, \delta t)$: as discussed in Section
        \ref{subsec:gen}, we chose $\epsilon = 0.01$ and $\delta t = 0.01$~s.

    %\item Maximum number of calls to EXTEND: $N_\mathrm{max\_rep} = 4000$
    % not mentioning this one: we set a maximum execution time anyway
\end{itemize}
Finally, in this tuning phase, we set the torque limit as
$(\taumax_1,\taumax_2) = (13,7)$ N.m, which are relatively ``slack'' values, in
order to obtain faster termination times for RRT. Tighter values such as
$(\taumax_1, \taumax_2) = (11, 5)$ N.m will be tested in our final comparison
with AVP-RRT in section~\ref{sec:compare}.

\begin{figure}[htp]
  \centering
  \textbf{A}\hspace{6cm}\textbf{B}

  \includegraphics[width=.49\textwidth]{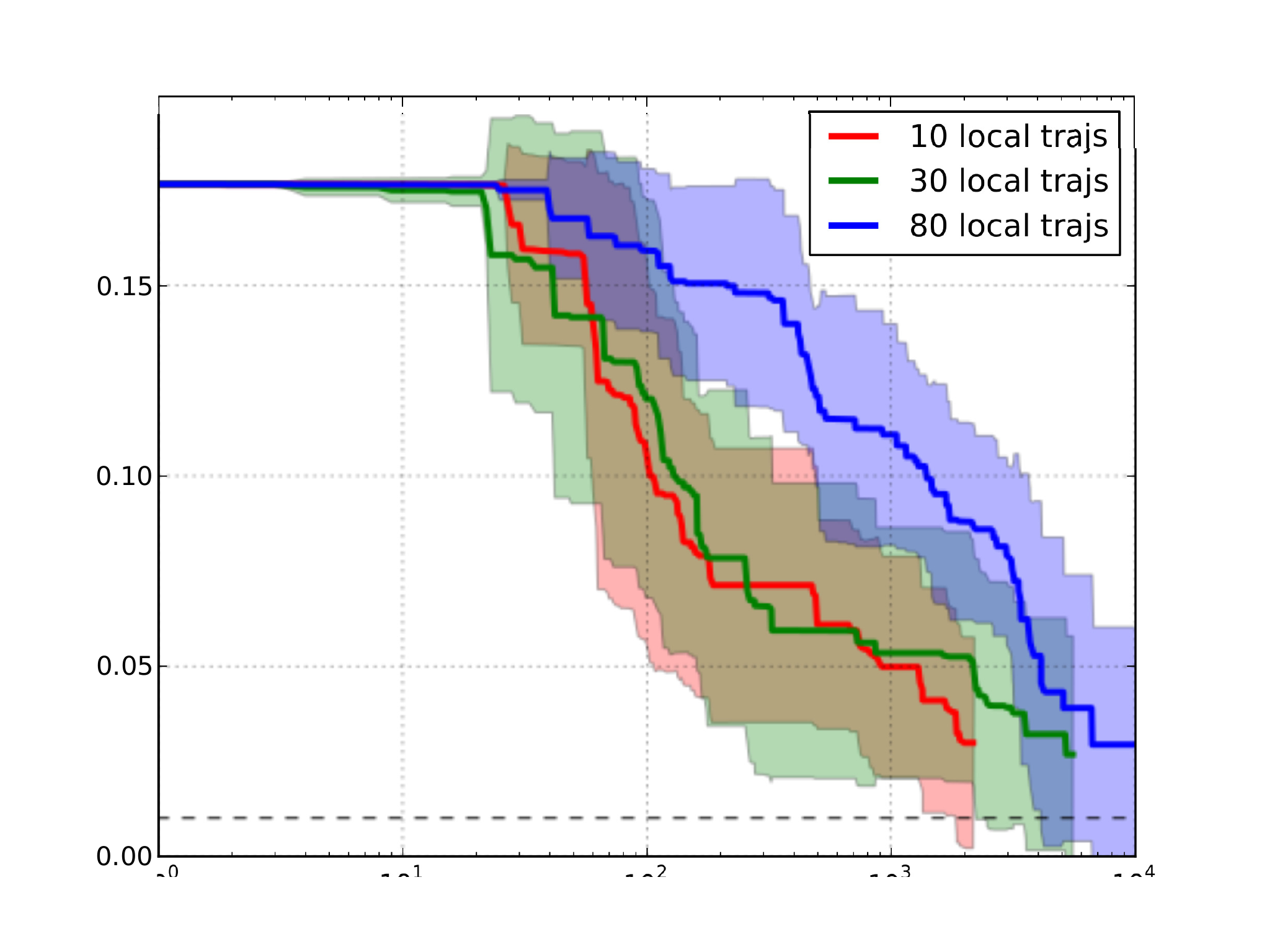}
  \includegraphics[width=.49\textwidth]{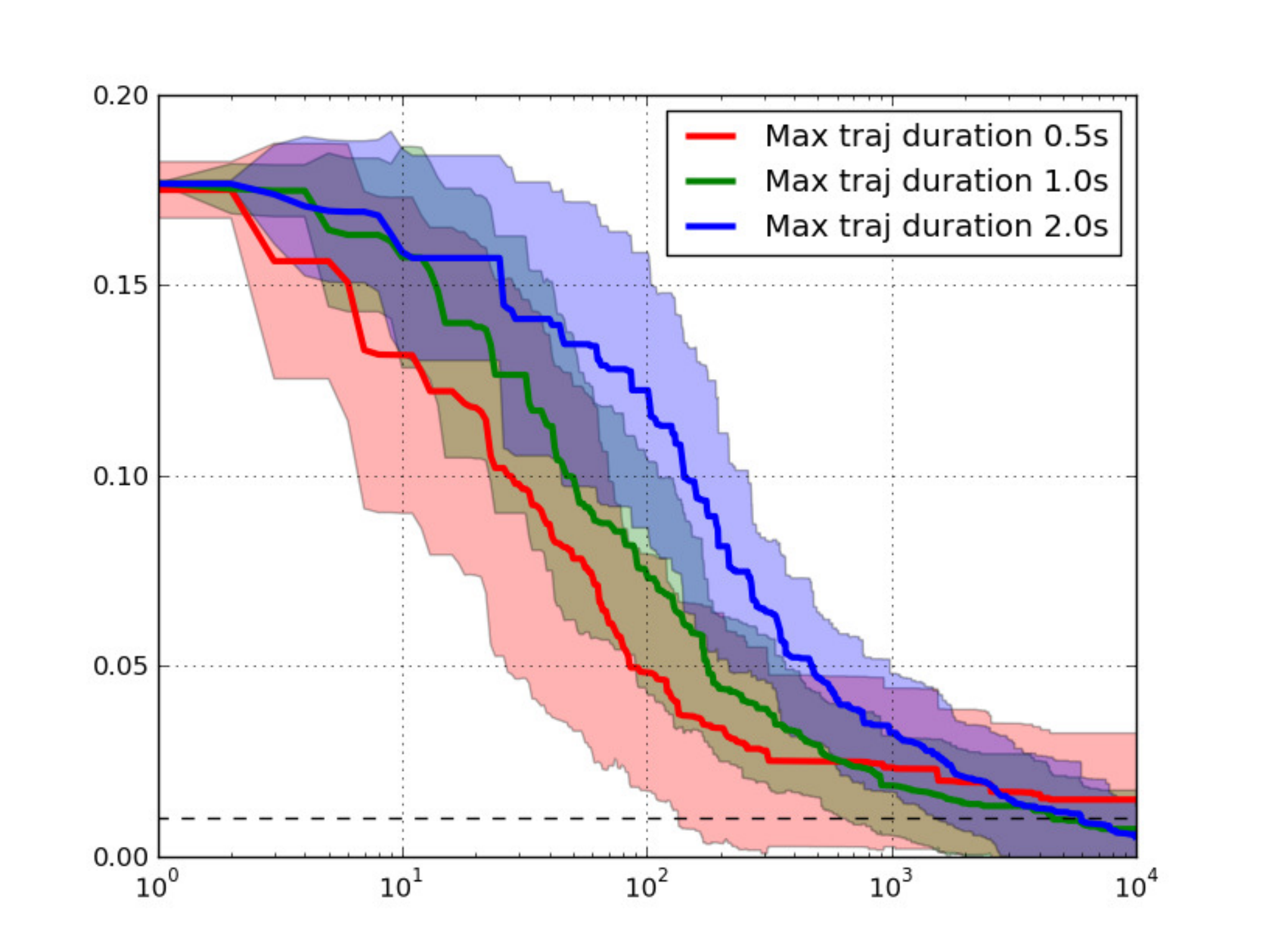}
  \caption{Minimum distance to the goal as a function of time for different
  values of $N_\mathrm{local\_trajs}$ and $\Delta t_{\max}$. At each instant,
  the minimum distance of the tree to the goal is computed. The average of this
  value across the 10 trials of each set is drawn in bold, while shaded areas
  indicate standard deviations. \textbf{A}: tuning of
  $N_\mathrm{local\_trajs}$. \textbf{B}: tuning of $\Delta t_{\max}$.}
  \label{fig:id}
\end{figure}

Fig.~\ref{fig:id}A shows the result of simulations for different values of
$N_\mathrm{local\_trajs}$. One can note that the performance of RRT is similar
for values $10$ and $30$, but gets worse for $80$. Based on this observation,
we chose $N_\mathrm{local\_trajs}=20$ for the final comparison in
section~\ref{sec:compare}.

Fig.~\ref{fig:id}B shows the simulation results for various values of $\Delta
t_{\max}$. Observe that the performance of RRT is similar for the three tested
values, with smaller values (\eg 0.5~s) performing better earlier in the trial
and larger values (\eg 2.0~s) performing better later on. We also noted that
smaller values of $\Delta t_{\max}$ such as 0.1~s or 0.2~s tended to yield
poorer results (not shown here). Our choice for the final comparison was thus
$\Delta t_{\max} = 1.0$~s.

\subsection{Comparing $K$NN-RRT and AVP-RRT}
\label{sec:compare}

In this section, we compare the performance of $K$NN-RRT (for
$K\in\{1, 10, 40, 100\}$, the other parameters being set to the values
discussed in the previous section) against AVP-RRT with 10
neighbors. For practical reasons, we further limited the execution
time of every trial to $10^4$~s, which had no impact in most cases or
otherwise induced a slight bias in favor of RRT (since we took
$10^4$~s as our estimate of the ``search time'' when RRT does not
terminate within this time limit).

We ran the simulations for two instances of the problem, namely
\begin{itemize}
\item $(\taumax_1, \taumax_2) = (11, 7)$ N.m;
\item $(\taumax_1, \taumax_2) = (11, 5)$ N.m.
\end{itemize}
For each problem instance, we ran 40 trials for each planner AVP-RRT,
state-space RRT with 1 nearest neighbor (RRT-1), RRT-10, RRT-40 and
RRT-100.  Note that for each trial $i$, all the planners received the
same sequence of random states
\begin{equation*}
  \mathbf{X}_i = \left\{\bfx_{\rand}^{(i)}(t) \in \mathbf{R}^4\ \middle|\ t \in \mathbf{N}\right\}
  \sim {\cal U}\left((]-\pi, \pi]^2 \times [-V_{\max}, +V_{\max}]^2)^{\mathbf{N}}\right),
\end{equation*}
although AVP-RRT only used the first two coordinates of each sample since it
plans in the configuration space. The results of this benchmark were already
illustrated in Fig.~\ref{fig:comp}. Additional details are provided in
Tables~\ref{tab:1107} and~\ref{tab:1105}. All trials of AVP successfully
terminated within the time limit.

For $(\taumax_1, \taumax_2) = (11, 7)$, the average search time was $3.3$~min.
Among the $K$NN-RRT, RRT-40 performed best with a success rate of 92.5\% and an
average computation time ca. 45 min, which is however $13.4$ times slower than
AVP-RRT.

For $(\taumax_1, \taumax_2) = (11, 7)$, the average search time was $9.8$~min.
Among the $K$NN-RRT, again RRT-40 performed best in terms of search time (54.6
min on average, which was $5.6$~times slower than AVP-RRT), but RRT-100
performed best in terms of success rate within the $10^4$s time limit (92.5\%).

\begin{table}[th]
  \centering
    \begin{tabular}{|c|c|c|c|c|}
      \hline
      Planner&Success rate&Search time (min)\\
      \hline
      AVP-RRT&100\%&3.3$\pm$2.6\\
      \hline
      RRT-1&40\%&70.0$\pm$34.1\\
      \hline
      RRT-10&82.5\%&53.1$\pm$59.5\\
      \hline
      RRT-40&92.5\%&44.6$\pm$42.6\\
      \hline
      RRT-100&82.5\%&88.4$\pm$54.0\\
      \hline
     \end{tabular}
     \caption{Comparison of AVP-RRT and $K$NN-RRT for 
       $(\taumax_1,\taumax_2) = (11,7)$. 
       %Note that in the calculation of
       %average search times, we set the search times of
       %unsuccessful trials to the time-out limit $10^4$~s.
       %
       % -> we say that the time limit was not reached anyway
       } 
 \label{tab:1107}  
\end{table}

% already in main article (fig:comp)
%
% \begin{figure}[htp]
%   \centering
%   \textbf{A}\hspace{6cm}\textbf{B}
% 
%   \includegraphics[width=6cm]{fig/comp-all-11-07}
%   \includegraphics[width=6cm]{fig/comp-vip-vs-rrt40-11-07}
%   \caption{Results for torque limits (11,7). \textbf{A}: Percentage of trials
%     that have reached the goal area at given time instants. \textbf{B}:
%     Individual plots for each trial. Each curve shows the distance to the goal as
%     a function of time for a given instance (red: AVP-RRT, blue: RRT-40).  Dots
%     indicate the time instants when a trial successfully terminated. Stars show
%     the mean values of termination times.}
%   \label{fig:1107}
% \end{figure}

\begin{table}[th]
  \centering
    \begin{tabular}{|c|c|c|c|c|}
      \hline
      Planner&Success rate&Search time (min)\\
      \hline
      AVP-RRT&100\%&9.8$\pm$12.1\\
      \hline
      RRT-1&47.5\%&63.8$\pm$36.6\\
      \hline
      RRT-10&85\%&56.3$\pm$60.1\\
      \hline
      RRT-40&87.5\%&54.6$\pm$52.2\\
      \hline
      RRT-100&92.5\%&81.2$\pm$46.7\\
      \hline
     \end{tabular}
  \caption{Comparison of AVP-RRT and $K$NN-RRT for
    $(\taumax_1,\taumax_2) = (11,5)$.}    
 \label{tab:1105}  
\end{table}

\section{Comparison of AVP-RRT with KPIECE on a 6-DOF and a 12-DOF
  manipulators}
\label{sec:kpiece}

Here, we detail the comparison between AVP-RRT and the OMPL
implementation of KPIECE~\citep{SK12tro,SMK12ram} on a kinodynamic
problem involving a $n$-DOF manipulators, for $2\leq n\leq 12$. The
full source code for this comparison is available at
\url{https://github.com/quangounet/kpiece-comparison}.

\subsection{KPIECE}

We used the implementation of KPIECE available in the Open Motion
Planning Library (OMPL)~\citep{SMK12ram}. The library provides
utilities such as function templates, data structures, and generic
implementations of various planners, written in C++ with Python
interfaces. It, however, does not provide modules such as collision
checker and modules for visualization purposes. Therefore, we used
OpenRAVE~\citep{Dia10these} for collision checking and visualization.

\subsubsection{Overall algorithm}

KPIECE grows a tree of motions in the state-space. A motion $\nu$ is a
tuple $(s, u, t)$, where $s$ is the initial state of the motion, $u$
is the control being applied to $s$, and $t$ is the control
duration. Initially the tree contains only one motion
$\nu_{\start}$. Then in each iteration the algorithm proceeds by first
selecting a motion on the tree to expand from. A control input is then
selected and applied to the state for a time duration. Finally, the
algorithm will evaluate the progress that has been made so far.

To select an existing motion from the tree, KPIECE utilizes
information obtained from projecting states in the state-space
$\mathcal{Q}$ into some low-dimensional Euclidean space
$\mathbf{R}^{k}$. Low-dimensionality of the space allows the planner
to discretize the space into cells. KPIECE will then score each cell
based on several criteria (see~\citep{SK12tro} for more detail). Based
on an assumption that the coverage of the low-dimensional Euclidean
space can reflect the true coverage of the state-space, KPIECE uses
its cell scoring system to help bias the exploration towards
unexplored area.

For the following simulations, we used KPIECE planner implementation
in C++ provided via the OMPL library. Since the library only provides
generic implementation of the planner, we also needed to implement
some problem specific functions for the planner such as state
projection and state propagation. Those functions were also
implemented in C++. We will give details on state projection and state
propagation rules we used in our simulations.

\paragraph{State projection} Since the state-space exploration is
mainly guided by the projection (as well as their cell scoring), more
meaningful projections which better reflect the progress of the
planner will help improve its performance. For planning problems for a
robot manipulator, we used a projection that projects a state to an
end-effector position in $3$D space. \citet{SK12tro} suggested that
when planning for a manipulator motion, the tool-tip position in $3$D
space is representative. However, by simply discarding all the
velocity components we may lose information which can essentially help
solve the problem. Thus, we decided to include also the norm of
velocity into the projection. This inclusion of the norm of velocity
was also used in~\citep{SK12tro} when planning for a modular
robot. Therefore, the projection projects a state into a space of
dimension $4$.

\paragraph{State propagation} KPIECE uses a control-based steering
method. It applies a selected control to a state over a number of
propagation steps to reach a new state. In our cases, since the robot
we were using was position-controlled, our control input were joint
accelerations. Let the state be $(\bf{q}, \dot{\bf{q}})$, where $\bfq$
and $\dot{\bfq}$ are the joint values and velocities,
respectively. The new state $(\bfq^+, \dot{\bfq}^+)$ resulting from
applying a control $\ddot{\bfq}$ to $(\bf{q}, \dot{\bf{q}})$ over a
short time interval $\Delta t$ can be computed from
\begin{align}
  \label{eq:KPIECE_control_update}
  \bfq^{+} &= \bfq + \Delta t \dot{\bfq} +
  0.5(\Delta t)^{2}\ddot{\bfq}\\
  \dot{\bfq}^{+} &= \dot{\bfq} + \Delta t \ddot{\bfq}.
\end{align}

\subsubsection{Fine-tuning of KPIECE}
We employed $L_{2}$ norm as a distance metric in order not to bias the
planning towards any heuristics. Next, in order for the planner not to
spend too much running time into simulations for fine-tuning, we
selected the threshold value to be $0.1$. The threshold is used to
decide whether a state has reached the goal or not. If the distance
from a state to the goal, according to the given distance metric, is
less than the selected threshold, the problem is considered as
solved. Then we tested the algorithm with a number of sets of
parameters to find the best set of parameters.

At this stage, the testing environment consisted only of the models of
the Denso VS-$060$ manipulator and its base. There was no other object
in the environment. Here, to check validity of a state, we need to
check for only robot self-collision. In the following runs, we planned
motions for only the first two joints of the robot (the ones closest
to the robot base). The robot had to move from $(0, 0, 0, 0)$ to $(1,
1, 0, 0)$, where the first two components of the tuples are joint
values and the others are joint velocities. We set the goal bias to
$0.2$. With chosen parameters and projection, we ran simulations with
difference combinations of cell size, $c$, and propagation step size,
$p$. Note that here we assigned cell size, which defines the
resolution of discretization of the projecting space, to be equal in
every dimension. Both cell size and propagation step size were chosen
from a set $\{0.01, 0.05, 0.1, 0.5, 1.0\}$. We tested for all
different $25$ combinations of the parameters and recorded the running
time of the planner. We ran $50$ simulations for each pair $(c,
p)$. For any value of cell size, we noticed that the propagation step
size of $0.05$ performed best. For $p = 0.05$, the values $c$ being
$0.05, 0.1, 1.0$ performed better than the rest. The resulting running
times using those values of cell size did not significantly differ
from each other. The differences were in order of $1$ ms. Therefore,
in the following section, we repeated all the simulations with three
different pairs $(c, p) \in \{ (0.05, 0.05), (0.1, 0.05), (1.0, 0.05)
\}$.

\subsection{KPIECE simulation results and comparison with AVP-RRT}
With the previously selected parameters, we conducted simulations as
follows. First of all, to show how running time of KPIECE and AVP-RRT
scale when the dimensionality of the problem increases, we used both
planners to plan motions for a $n-DOF$ robot, with $2\leq n\leq
12$. For this, we concatenated two Denso VS-$060$ manipulator
manipulators together into a composite 12-DOF manipulator and used the
first $n$ joints of that robot.

The robot was to move from all-zeros configuration to all-ones
configuration. Initial and final velocities were set to zero. There
was no other obstacle in the scene. Since the implementation of KPIECE
is unidirectional, we also used a unidirectional version of
AVP-RRT. The AVP-RRT implementation was written in Python. Only the
time-parameterization algorithm was implemented in C++.

\begin{figure}[htp]
  \centering 
  \hspace{-0.5cm}\textbf{A}\hspace{5cm}\textbf{B}\\    
  \includegraphics[width = 0.49\textwidth]{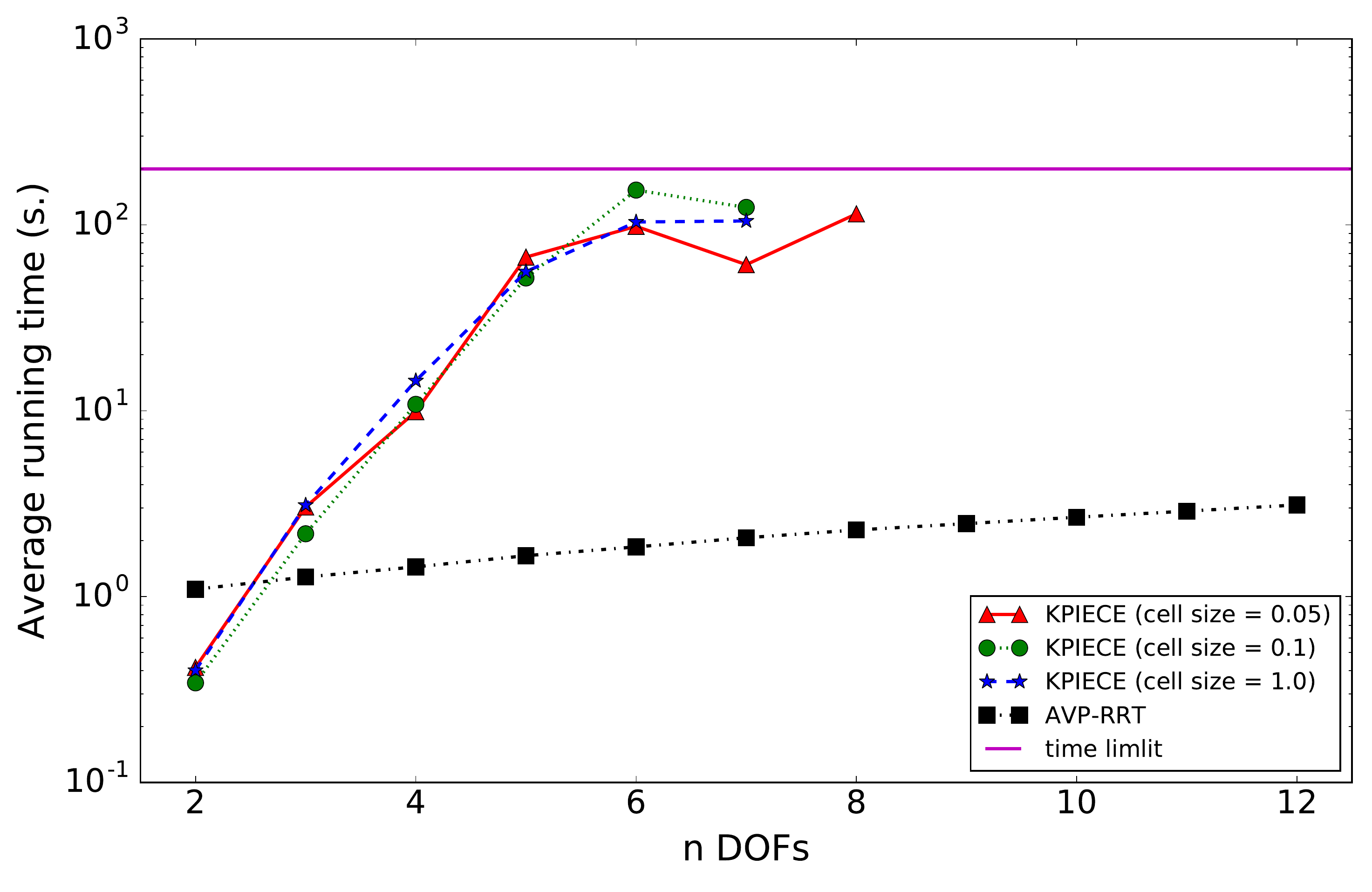}
  \includegraphics[width = 0.46\textwidth]{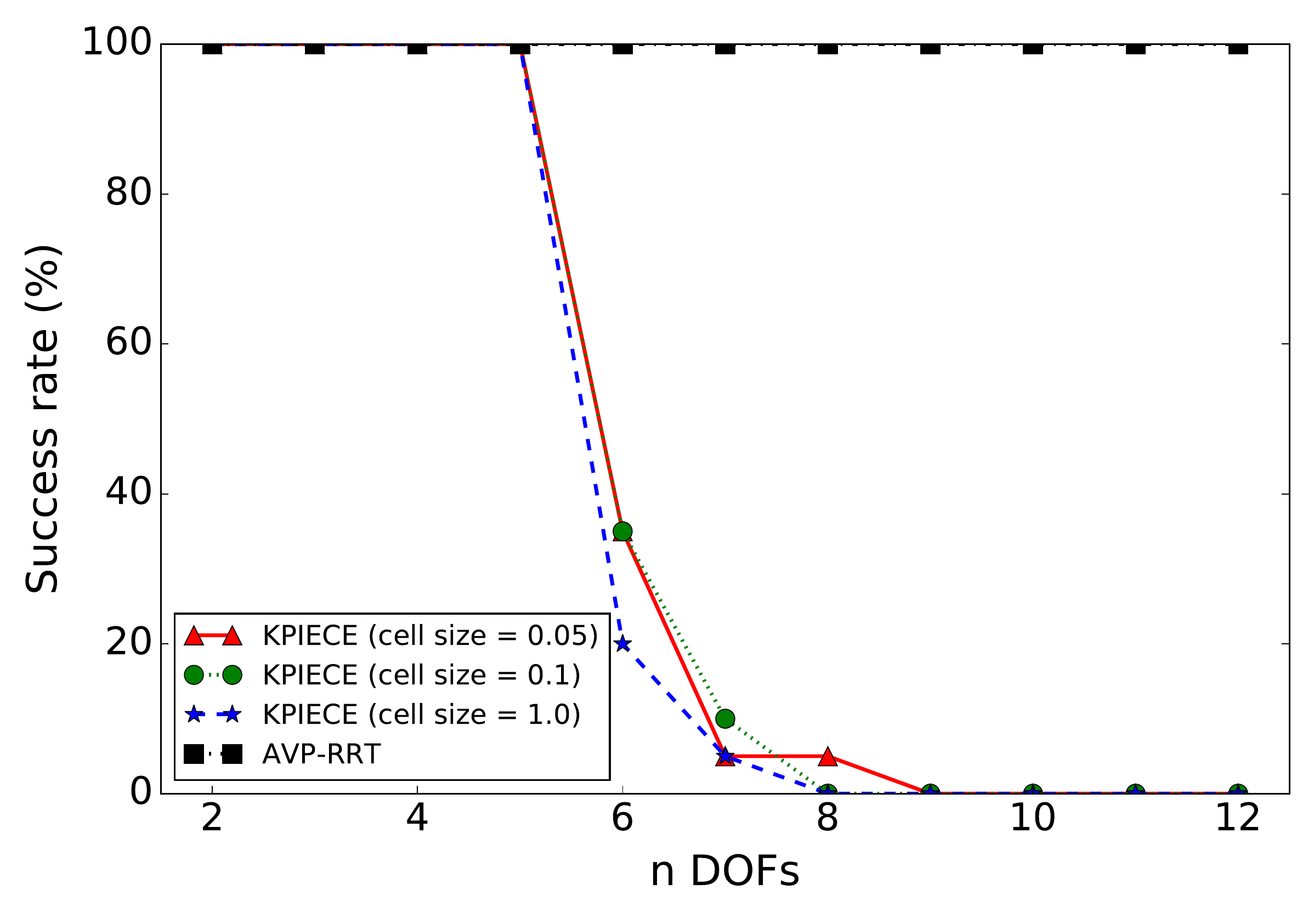}
  \caption{\textbf{A}\,: Average running time of KPIECE and AVP-RRT
    (black squares) taken over $20$ runs as a function of the number
    of DOF. For KPIECE, simulations were run with three different cell
    sizes: $0.05$ (red triangles), $0.1$ (green circles), and $1.0$
    (blue stars).  \textbf{B}\,: Succcess rates of KPIECE and AVP-RRT
    over $20$ runs. When the number of DOFs was higher than 8, KPIECE
    failed to find any solution within the given time limit of
    200\,s.}
  \label{fig:kpiece_sim}  
\end{figure}

We gave each planner 200\,s. for each run, and simulated $20$ runs for
each condition. For KPIECE we tested different values of cell sizes
(0.05, 0.1, and 1.0). Fig.~\ref{fig:kpiece_sim}A shows average running
times over 20 runs. From the figure, the three values of the cell size
produced similar results. Although KPIECE performed well when planning
for low numbers of DOFs, the running time increased very quickly
(exponentially) with increasing numbers of DOFs. Correlatively, the
success rate when planning using KPIECE dropped rapidly when the
number of DOFs increased, as can be seen from
Fig.~\ref{fig:kpiece_sim}A. When the number of DOFs was higher than 8,
KPIECE failed to find any solution within 200\,s. The computation time
for AVP-RRT also increased exponentially with the number of DOFs but
the rate was much lower as compared to KPIECE.

Finally, we considered an environment similar to that of our
experiment on non-prehensile object transportation of
Section~\ref{sec:bottle}. The tray and the bottle were, however,
removed from the (6-DOF) robot model. The problem was therefore less
constrained. We considered here only bounds on joint values, joint
velocities, and joint accelerations. We shifted the lower edge of the
opening upward for $13$\,cm. and set the opening height to be lower
($25$\,cm. in this case) to make the problem more interesting. Then
for each run, both planners had a time limit of $600$\,s to find a
motion for the robot to move from one side of the wall to the
other. We repeated simulations $20$ times for both planners. For
KPIECE, since the performance when using different cell size from
$\{0.05, 0.1, 1.0\}$ did not differ much from each other, we chose to
ran simulations with cell size $c = 0.05$.

\begin{figure}[htp]
  \centering
  \textbf{A}\hspace{7cm}\textbf{B}

  \includegraphics[width=0.45\textwidth]{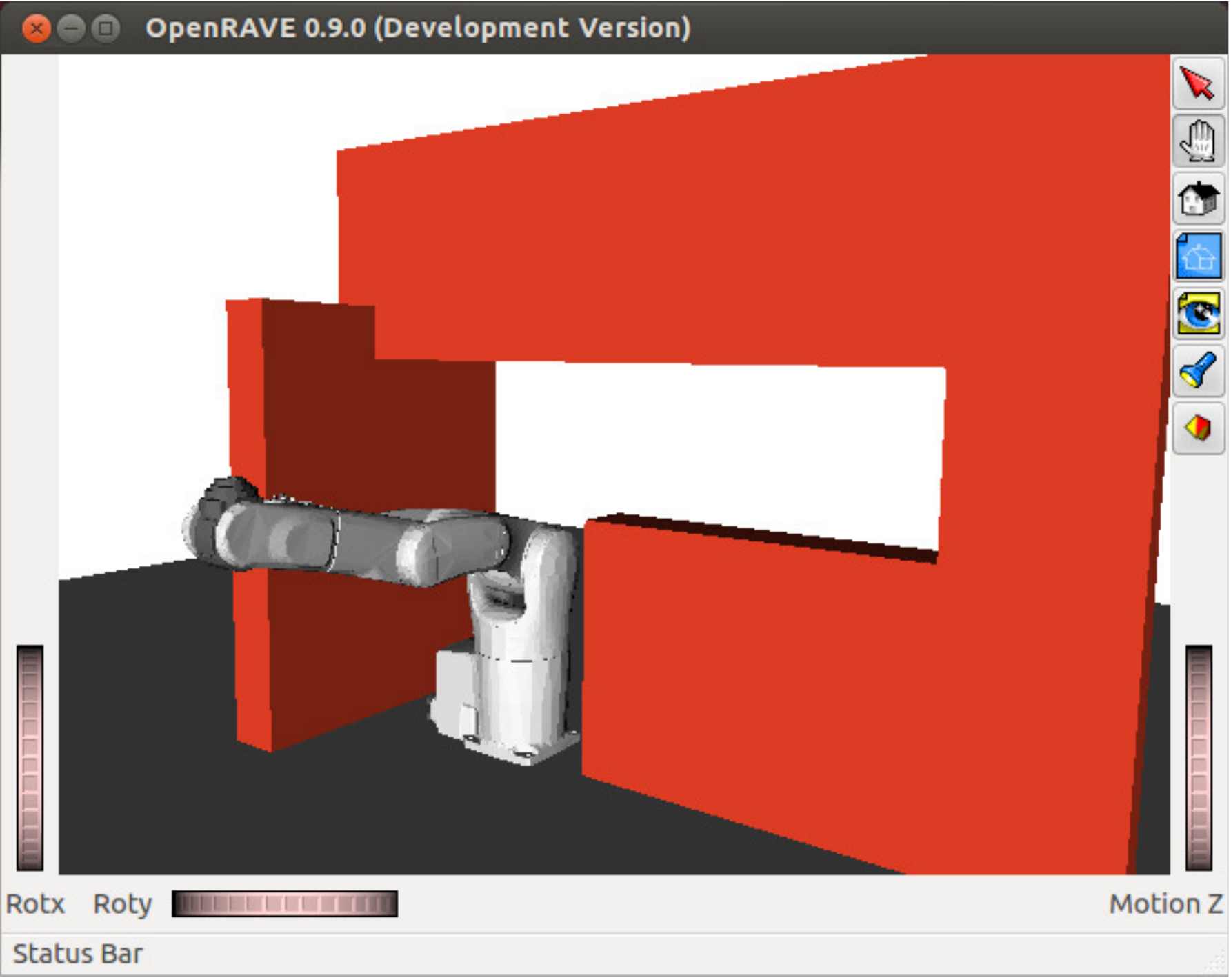}
  \quad
  \includegraphics[width=0.45\textwidth]{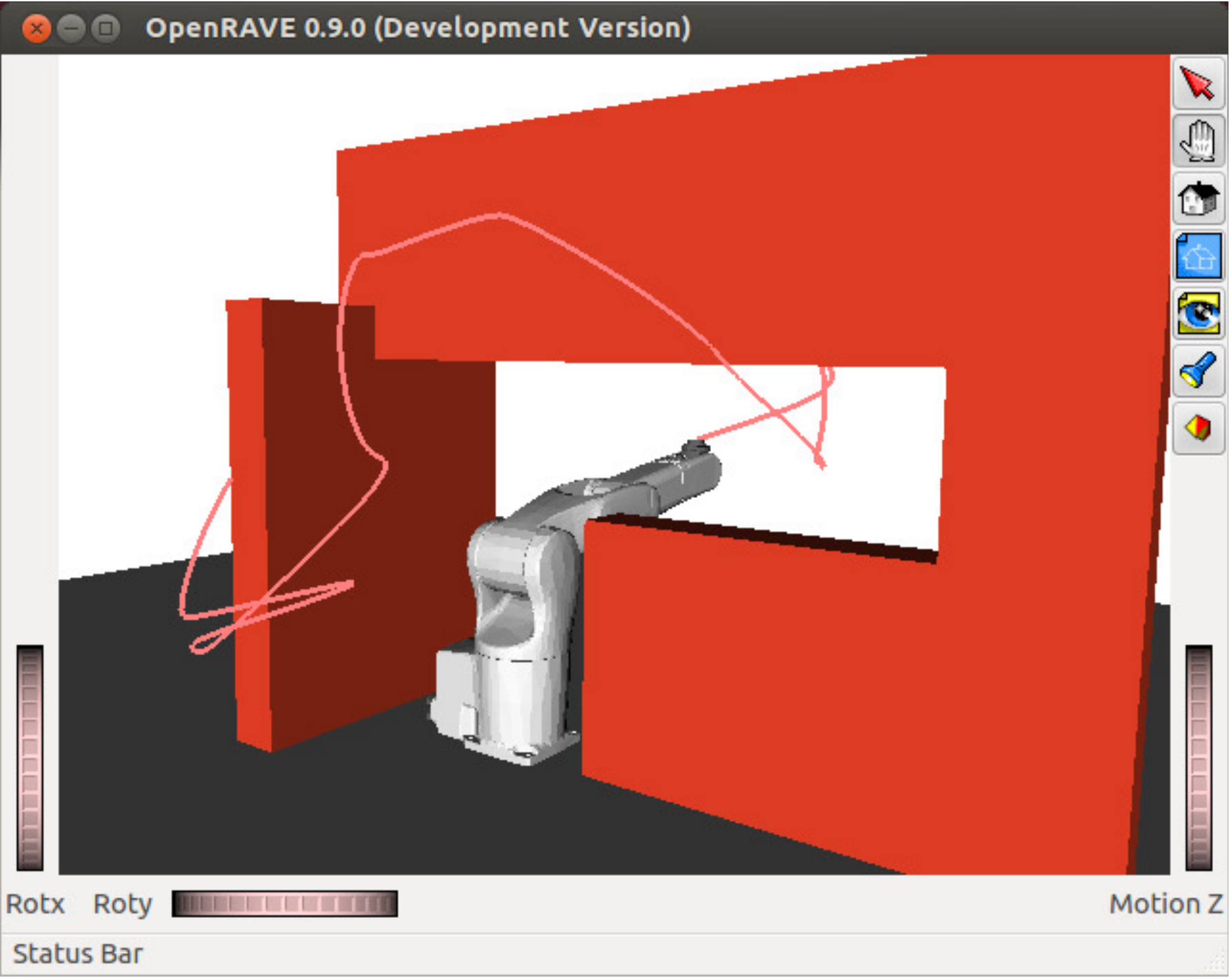}
  \caption{The scene used in the last experiment. Both KPIECE and
    AVP-RRT were to find a motion for the robot to move from one side
    of the wall to the other. \textbf{A}: the start configuration of
    the robot. \textbf{B}: the goal configuration of the robot. The
    pink line in the figure indicates an end-effector path from a
    solution found by AVP-RRT. KPIECE could not find any solution, in
    any run, within the given time limit of 600\,s.}
  \label{fig:avp-rrt-example-solution}
\end{figure}

The average running time for AVP-RRT in this case was $68.67$\,s with
a success rate of $100\%$. The average number of nodes in the tree
when the planner terminated was $60.15$ and the average number of
trajectory segments of the solutions was
$8.60$. Fig.~\ref{fig:avp-rrt-example-solution} shows the scene used
in simulations as well as an example of a solution trajectory found by
AVP-RRT. On the other hand, KPIECE could not find any solution, in any
run, within the given time limit.

\bibliographystyle{plainnat}
\bibliography{cri}

\end{document}